\documentclass[10pt,preprint]{article}
\PassOptionsToPackage{table,xcdraw}{xcolor}  %
\let\origaddcontentsline\addcontentsline

\usepackage{rlc}

\let\addcontentsline\origaddcontentsline

\usepackage{hyperref}
\hypersetup{
    colorlinks, linkcolor={dark-blue},
    citecolor={dark-blue}, urlcolor={dark-blue}
}

\usepackage{etoolbox}
\pretocmd{\section}{\phantomsection}{}{}
\pretocmd{\subsection}{\phantomsection}{}{}
\usepackage{tikz}
\usepackage{tcolorbox}
\usepackage{xcolor}
\usetikzlibrary{shapes,arrows,positioning,decorations.pathmorphing}

\usepackage{authblk} 
\usepackage{amssymb}            
\usepackage{mathtools}          
\usepackage{mathrsfs}           
\usepackage{amsthm}

\mathtoolsset{showonlyrefs}     
\usepackage{graphicx}           
\usepackage{subcaption}         
\usepackage[space]{grffile}     
\usepackage{url}                
\usepackage{booktabs}       
\usepackage{amsfonts}       
\usepackage{nicefrac}       
\usepackage{enumitem}  %
\usepackage{graphicx}  
\usepackage{longtable}  
\usepackage{array}  
\usepackage{tabularx}
\usepackage{float}
\usepackage{tikz}
\usepackage{tikz-qtree}
\usetikzlibrary{trees,positioning,shapes,arrows}
\usepackage{natbib}
\usepackage{doi}
\usepackage{svg}
\usepackage{tikz}
\usepackage[table]{xcolor}
\usetikzlibrary{trees, shapes, arrows, positioning, calc}
\usepackage{array, makecell, booktabs}
\newcolumntype{P}[1]{>{\raggedright\arraybackslash}p{#1}}
\definecolor{greyboxbg}{RGB}{221,221,221}
\definecolor{blueboxbg}{RGB}{240,240,240}
\definecolor{orangeboxbg}{RGB}{200,255,200}
\definecolor{greenboxbg}{RGB}{142,207,201}
\usepackage{multirow} 

\usepackage{tikz}
\usetikzlibrary{shadings,arrows.meta}

\usepackage{tocloft}

\usepackage[edges]{forest}
\usepackage{tikz}
\usetikzlibrary{arrows.meta}

\tcbset{
  examplebox/.style={
    colback=gray!5,
    colframe=gray!40,
    fonttitle=\bfseries,
    title=Example,
    coltitle=black,
    boxrule=0.4pt,
    arc=2mm,
    top=1mm,
    bottom=1mm,
    left=1mm,
    right=1mm,
  }
}
\newtcolorbox{example}[1][]{examplebox,#1}

\tcbset{Conditionbox/.style={
  colback=blue!3!white,
  colframe=blue!60!black,
  coltitle=black,
  title= Core Conditions for the Agentic Web.
}}

\tcbset{definitionbox/.style={
  colback=blue!5!white,
  colframe=blue!75!black,
  fonttitle=\bfseries,
  coltitle=white,
  title=Definition: Agentic Web
}}

\tcbset{
  myboxstyle/.style={
    fonttitle=\bfseries,
    fontupper=\normalsize,
    boxrule=0.6pt,
    arc=1.5mm,
    top=2mm,
    bottom=2mm,
    left=2mm,
    right=2mm,
  }
}

\tcbset{keyterminology/.style={
  colback=blue!3!white,
  colframe=blue!60!black,
  coltitle=black,
  title=Conceptual Layers,
  colbacktitle=blue!20!white,
  myboxstyle
}}

\tcbset{conceptrelation/.style={
  colback=blue!3!white,
  colframe=blue!60!black,
  coltitle=black,
  title=Core Conditions for the Agentic Web,
  colbacktitle=blue!20!white,
  myboxstyle
}}

\tcbset{frameworkoverview/.style={
  colback=blue!3!white,
  colframe=blue!60!black,
  coltitle=black,
  title=Framework Overview,
  colbacktitle=blue!25!white,
  myboxstyle
}}

\tcbset{example/.style={
  colback=blue!3!white,
  colframe=blue!60!black,
  coltitle=black,
  title=Example (Transactional),
  colbacktitle=blue!20!white,
  myboxstyle
}}

\tcbset{example2/.style={
  colback=blue!3!white,
  colframe=blue!60!black,
  coltitle=black,
  title=Example (Informational),
  colbacktitle=blue!20!white,
  myboxstyle
}}

\tcbset{layerinteraction/.style={
  colback=blue!3!white,
  colframe=blue!60!black,
  coltitle=black,
  title=Layer Interaction,
  colbacktitle=blue!22!white,
  myboxstyle
}}

\newtoggle{final}
\togglefalse{final} 

\title{\vspace{+1cm}Agentic Web:\\ Weaving the Next Web with AI Agents\vspace{-1cm}}

\author{
}

\author{\\
\name{Yingxuan Yang}$^{1}$,
\name{Mulei Ma}$^{2}$,
\name{Yuxuan Huang}$^{3}$,
\name{Huacan Chai}$^{1}$,
\name{Chenyu Gong}$^{2}$, \\
\name{Haoran Geng}$^{4}$,
\name{Yuanjian Zhou}$^{5}$,
\name{Ying Wen}$^{1}$,
\name{Meng Fang}$^{3}$, 
\name{Muhao Chen}$^{6}$,\\
\name{Shangding Gu}$^{4}$\textsuperscript{*}, \name{Ming Jin}$^{7}$,
\name{Costas Spanos}$^{4}$,
\name{Yang Yang}$^{2}$, \name{Pieter Abbeel}$^{4}$,\\
\name{Dawn Song}$^{4}$,
\name{Weinan Zhang}$^{1,5}$\textsuperscript{*},
\name{Jun Wang}$^{8}$\thanks{S. Gu, W. Zhang and J. Wang are the corresponding authors.} \vspace{+0.15cm} \\
$^1$Shanghai Jiao Tong University \quad\\
$^2$The Hong Kong University of Science and Technology, Guangzhou \quad\\
$^3$University of Liverpool \quad
$^4$University of California, Berkeley \quad
$^5$Shanghai Innovation Institute\quad
$^6$University of California, Davis \quad
$^7$Virginia Tech \quad
$^8$University College London \\

\texttt{zoeyyx@sjtu.edu.cn}\quad \texttt{shangding.gu@berkeley.edu}\\
\texttt{wnzhang@sjtu.edu.cn}\quad \texttt{jun.wang@cs.ucl.ac.uk}
}

\begin{document}
\maketitle

\vspace{-0.6cm}
\begin{abstract}\vspace{-0.4cm}
Traditionally, the Web has served as a platform for connecting information, resources, and people, enabling human–machine interaction through activities such as searching, browsing, and performing tasks that are informational, transactional, or communicational. This original Web was fundamentally about connection, linking users to content, services, and one another.

The emergence of AI agents powered by large language models (LLMs)  marks a pivotal shift toward the \emph{Agentic Web}, a new phase of the internet defined by autonomous, goal-driven interactions. In this paradigm, agents interact directly with one another to plan, coordinate, and execute complex tasks on behalf of users. This transition from human-driven to machine-to-machine interaction allows intent to be delegated, relieving users from routine digital operations and enabling a more interactive, automated web experience.

In this paper, we present a structured framework for understanding and building the Agentic Web. We trace its evolution from the PC and Mobile Web eras and identify the core technological foundations that support this shift. Central to our framework is a conceptual model consisting of three key dimensions: intelligence, interaction, and economics. These dimensions collectively enable the capabilities of AI agents, such as retrieval, recommendation, planning, and collaboration.

We analyze the architectural and infrastructural challenges involved in creating scalable agentic systems, including communication protocols, orchestration strategies, and emerging paradigms such as the Agent Attention Economy. We conclude by discussing the potential applications, societal risks, and governance issues posed by agentic systems, and outline research directions for developing open, secure, and intelligent ecosystems shaped by both human intent and autonomous agent behavior. A continuously updated collection of relevant studies for agentic web is available at: \url{https://github.com/SafeRL-Lab/agentic-web}.\vspace{-0.2cm}
\end{abstract}

\hspace{32pt} \textbf{Keywords:} Agentic Web, LLM Agents, Web Architecture, Safety \& Security


\newpage
\tableofcontents

\newpage
\section{Introduction}
The Web has long served as a platform for \textit{connectivity} \citep{berners1999weaving,castells2002internet}, linking people to information, services, and one another. In its early phases, the Web enabled human–machine interaction for tasks that were informational (e.g., reading news), transactional (e.g., online shopping), or communicational (e.g., messaging and email). Intelligence in this era resided in the tools that helped users access, filter, and interact with content: search engines \citep{brin1998anatomy}, recommender systems \citep{10.1145/1148170.1148257, koren2009matrix, 10.1145/2505515.2505690, 10.1145/2484028.2484126}, and user interfaces \citep{deaton2003elements}. However, the user was always the active party, manually navigating between pages, initiating actions, and making decisions at every step.

For the last few years, a shift has been taking place: the emergence of AI agents powered by large language models (LLMs) \citep{yang2023auto, kapoor2024ai}. These \emph{AI agents} are software entities capable of perceiving their environment, reasoning, and taking actions autonomously to achieve goals set by the user. With the integration of perception and execution components, LLMs are no longer limited to responding to prompts: they can act through agents that plan, remember, and interact across digital systems \citep{wang2023describe}. Importantly, these agents are not constrained to single-turn interactions but can carry out complex, long-horizon tasks. Moreover, multiple agents can be orchestrated to work collaboratively on sophisticated objectives \citep{qian2024chatdev, yang2025unlocking, gottweis2025towards,sapkota2025ai}.

The transformation toward agent-based systems is driven by two powerful forces. First, AI assistants are becoming increasingly capable of handling complex, multi-step tasks across domains such as research~\citep{ren2025scientificintelligencesurveyllmbased, huang2025deepresearchagentssystematic,schmidgall2025agentlaboratoryusingllm}, software development~\citep{hong2023metagpt,xia2024agentlessdemystifyingllmbasedsoftware}, customer support~\citep{10.1145/3626772.3661345}, and personal productivity~\citep{li2024personalllmagentsinsights}. These agents are no longer reactive tools responding to isolated prompts, but proactive collaborators that plan, reason, and execute actions over time. Second, users are becoming more comfortable delegating not just individual queries but entire workflows (sometimes spanning minutes, hours, or even days) to such agents~\citep{guo2024dsagentautomateddatascience,hong2024datainterpreterllmagent}. This growing trust in agent autonomy introduces new expectations and necessitates new interfaces, leading to a fundamental shift in how the Web is used and experienced.

This evolution lays the foundation for what we formally define as the \emph{Agentic Web}. In this emerging paradigm, the Web is no longer merely a platform for human interaction with content and services, but a dynamic environment in which autonomous agents act, communicate, and collaborate across services and domains on behalf of their users~\citep{petrova2025semanticwebmasagentic, lù2025buildwebagentsagents,chaffer2025know}. For instance, the ChatGPT Agent released in July 2025 enables AI agents to act on behalf of users by performing tasks such as planning and purchasing ingredients for a Japanese breakfast or booking reservations \citep{chatgpt2025agent}.

\vspace{-0.1cm}
\begin{tcolorbox}[definitionbox]
\textbf{The \emph{Agentic Web}} is a distributed, interactive internet ecosystem in which autonomous software agents, often powered by large language models, act as autonomous intermediaries that persistently plan, coordinate, and execute goal-directed tasks. In this paradigm, web resources and services are agent-accessible, enabling continuous agent-to-agent interaction, dynamic information exchange, and value creation alongside traditional human-web interactions.
\end{tcolorbox}
\vspace{-0.1cm}
Unlike the traditional Web, which serves primarily to connect documents, services, and users for informational, transactional, and communicational purposes, the Agentic Web enables intelligent, goal-directed interaction.
While the core functions of accessing information, completing transactions, and facilitating communication remain, they are now mediated by autonomous agents capable of reasoning, planning, and acting on behalf of users.

The defining shift is from short-term, one-off interactions between users and static content, to sustained, long-term interactions involving sequences of coordinated actions across multiple services, webpages, and domains. In the Agentic Web, the \emph{end users} remain human, while the \emph{mid users} (those who actively navigate, process, generate content, and interact with the environment) are AI agents. These agents interpret and carry out user intent by interacting with a distributed network of other agents and services.

\begin{figure}[t]
    \centering
    \includegraphics[width=0.85\linewidth]{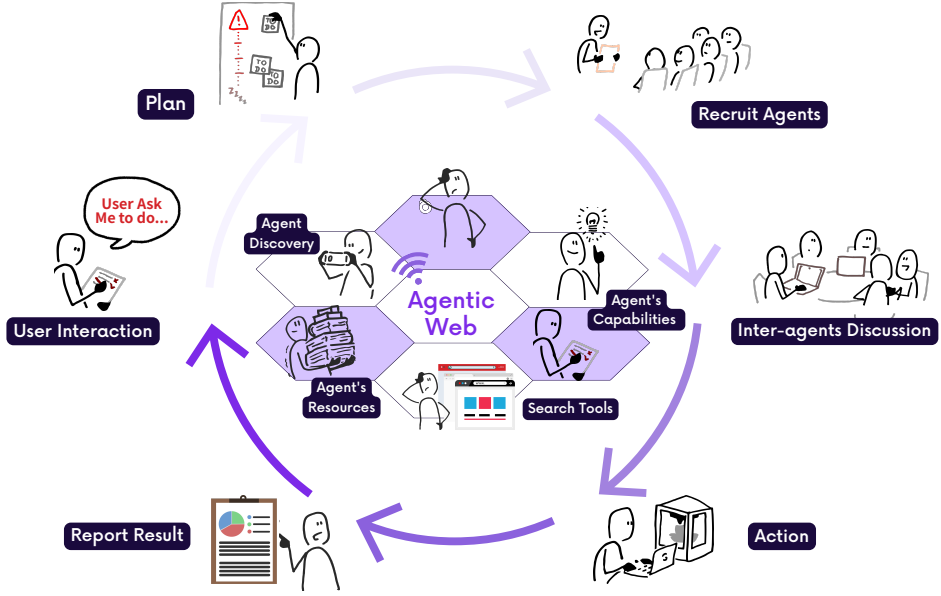}
    \vspace{-0.2cm}
    \caption{Illustration of the Agentic Web process cycle. The cycle begins with a user submitting a task request. The system then plans the task and identifies appropriate agents and tools. Recruited agents engage in inter-agent discussions, collaborate using their unique capabilities and resources, and execute the task. The results are reported back to the user, completing the cycle. The Agentic Web facilitates discovery, coordination, and cooperation among agents to fulfill user goals.}
    \label{fig:Agentic_web_overview}
    \vspace{-0.6cm}
\end{figure}

A user query is no longer a simple request for isolated information, but a delegation of a complex task, which may involve negotiation, planning, and adaptation over multiple steps. With the support of structured or open-ended communication protocols \citep{yang2025survey}, these agents collaborate across domains to complete workflows and deliver results that reflect high-level user goals \citep{lin2024mao,yang2025agentnet}. This agent-mediated process is illustrated in Figure~\ref{fig:Agentic_web_overview}, which depicts a typical task lifecycle from user intent to multi-agent execution and result delivery.

In this new paradigm, webpages evolve into active software agents, characterised not just by their static content but by their capabilities, interfaces, and roles within broader task structures. Hyperlinks, which once represented passive navigational paths, now act as coordination channels that facilitate inter-agent communication, dynamic task decomposition, and cooperative execution. The Agentic Web, therefore, transforms the Web from a network of linked documents into an ecosystem of interactive, intelligent agents.

Beyond changes in interaction models, the Agentic Web also redefines how information is stored, linked, and transmitted. In the early Personal Computer (PC) era, web content was mostly institutionally produced, with relatively small data volumes that users accessed primarily through keyword search. As the mobile internet expanded, User-Generated Content (UGC) exploded, increasing the scale and diversity of information. This shift raised the cost of search and gave rise to recommendation systems as the dominant paradigm for matching information supply and demand.

With the emergence of LLMs and agentic systems, the underlying logic of information flows undergoes another major transformation. Now, much of the world’s knowledge is not only stored on static web pages but also embedded in the parameters of LLMs themselves. Agents can access this learned knowledge directly, link it with real-time retrieval, and autonomously interact with other agents or online resources.

This enables agents to proactively recommend relevant content, going beyond traditional search engines, and to perform deeper and more personalized information retrieval. Moreover, agents can execute transactions and complete consumption processes on behalf of users, introducing a new production–consumption dynamic in which information and services may be created primarily for agents rather than humans. In some cases, web content may not be authored directly by humans at all but generated by agents in real time, leading to an ecosystem where agents both produce and consume knowledge.

\begin{tcolorbox}[example]
In the \emph{Traditional Web}, a transactional task such as booking a flight is manually performed by the user. The process typically involves visiting travel websites, entering search queries, adjusting filters, comparing ticket options across multiple tabs or platforms, and finalising the booking decision. While the Web may offer assistance through features such as recommendation engines, user interfaces, and search algorithms, the task execution remains user-driven and requires active, step-by-step involvement.

In the \emph{Agentic Web}, the same task is initiated through high-level intent delegation. The user provides a goal-oriented instruction (e.g., ``Book a flight to New York next weekend within my budget''), and an autonomous agent carries out the task on their behalf. The agent autonomously interacts with services and APIs, queries and parses webpages, refines options based on user preferences, and completes the booking. It may perform multiple iterations and coordinate with other agents, requiring no further user intervention. 
\end{tcolorbox}

\begin{figure}[t]
    \centering
    \includegraphics[width=0.93\linewidth]{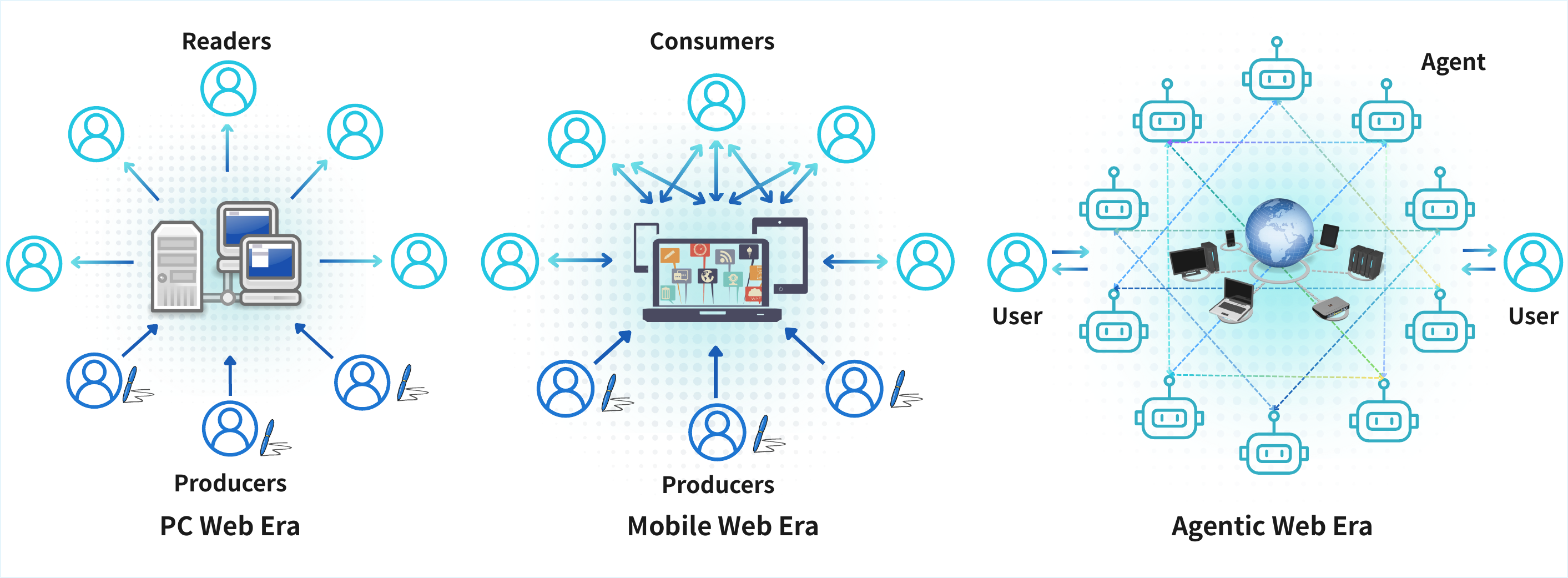}
    \vspace{-0.4cm}
    \caption{Evolution of user-system interaction across three internet eras. In the PC Web Era, users acted primarily as content consumers with limited interaction. The Mobile Web Era introduced a bidirectional flow, enabling users to both consume and produce content. In the emerging Agentic Web Era, tasks are delegated to ai agents, who interact with information networks on their behalf. The expanding and darkening circles reflect the increasing complexity and volume of information.}
    \label{fig:interaction}
    \vspace{-0.3cm}
\end{figure}

The above example illustrates the core distinction: the Traditional Web is defined by human-led interaction over static services, while the Agentic Web enables persistent, intelligent, machine-led workflows that extend across multiple services and interactions. Figure~\ref{fig:interaction} complements this distinction by visualizing how user-system interactions have evolved from passive consumption to active agent delegation across three Web eras.

\begin{tcolorbox}[example2]
In the \emph{Traditional Web}, an informational task such as understanding how different large language models process multimodal inputs requires the user to manually locate whitepapers, extract architecture diagrams, search for benchmark results, and assemble the findings into a report. This involves switching between academic search engines, blog posts, PDF viewers, and spreadsheet tools.

In the \emph{Agentic Web}, the same task is delegated to a Deep Research agent (e.g., ``Produce a report comparing how GPT-4o, Gemini, and Claude handle text and image inputs, including tables and flowcharts''). The agent interprets the query and plans a multi-stage workflow. It retrieves content from online sources and technical repositories via API calls, browser access, and the Model Context Protocol (MCP)~\citep{anthropic2025}, which enables standardized access to external tools and structured resources. The agent then parses PDF and HTML documents, invokes specialized modules for table extraction, diagram generation, and result visualization, and integrates the outputs into a structured report through multi-step reasoning. 
\end{tcolorbox}

This example illustrates how the Agentic Web extends beyond static content retrieval to complex, adaptive information processing.

As a result, foundational Web concepts such as PageRank \citep{page1999pagerank}, along with broader systems including web search~\citep{10.1145/792550.792552}, recommender systems~\citep{10.1145/245108.245121}, and computational advertising models~\citep{nelson1974advertising}, must be reinterpreted. Rather than focusing solely on static link popularity or historical user interactions, they may increasingly reflect the dynamic utility, responsiveness, and cooperation potential of agents operating within the network.
Similarly, traditional techniques like web crawlers, once designed to index static content, could evolve into agent crawlers, autonomous explorers that discover and negotiate with other agents, indexing not just data but service capabilities, interface affordances, and cooperation histories.
The metadata of webpages becomes richer and more actionable: beyond simple tags or descriptors, agent metadata may include standardized schemas describing APIs, trust levels, performance benchmarks, or negotiation protocols. 
The old idea of web directories or yellow pages, once manually curated lists of websites categorized by topic, can be reimagined as dynamic agent registries or marketplaces that index available agents by domain expertise, reputation, and inter-agent compatibility.
In such an agentic environment, search engines could transform into sophisticated orchestrators, not only retrieving relevant agents but also composing, coordinating, and managing workflows among them to fulfill complex delegated tasks. Just as PageRank once signaled page authority, future agent ranking algorithms may factor in cooperation success rates, responsiveness, and the agent’s contribution to multi-agent workflows. Together, these reinterpretations and shifts pave the way for a new generation of algorithms and protocols for agent discovery, trust calibration, incentive alignment, and orchestration \citep{lin2024mao,wang2025internet}, enabling the Web to operate as an open, distributed, and continuously evolving collective of collaborative intelligences.

Therefore, it becomes essential to revisit the foundational technologies and modules of the Web and reinterpret them in the context of the Agentic Web. Core components such as HTTP protocols, HTML semantics, indexing, search, and recommender systems must be reconsidered through the lens of agent autonomy and collaboration. Despite the rapid emergence of Agentic AI, there is a noticeable gap in the current literature in systematically analyzing and redefining these web fundamentals for an agent-driven future. Bridging this gap is crucial for understanding and shaping the next evolution of the Internet, which is the goal of this article.

In summary, the key contributions and the structure of this article are outlined as follows. In Section \ref{sec:history}, we review the historical evolution of the Web and offer a forecasting-style analysis to project the development trajectory of the Agentic Web in the near future. Section \ref{sec:agentic-web} introduces and conceptualizes the Agentic Web as a fundamentally new form of the Web, presenting a three-dimension model along with a set of research propositions that frame its emerging dynamics. In Section \ref{sec:techniques}, we delve into the core tasks and enabling techniques of the Agentic Web, covering areas such as information retrieval, Recommender Systems, agent planning, and multi-agent learning and coordination. Section \ref{sec:system} explores the evolving system landscape and proposes key design principles to guide the development of Agentic Web infrastructure. In Section \ref{sec:agentic-web-applications}, we examine representative applications of the Agentic Web, including use cases like e-commerce ordering, travel planning, and enterprise knowledge assistants. Section \ref{sec:risk} addresses the associated technical risks, information security concerns, regulatory challenges, and potential mitigation strategies. Finally, in Section \ref{sec:Challenges} and Section \ref{sec:conclusion}, we conclude by summarizing the major themes of the paper and discussing the future outlook for the continued evolution of the Agentic Web.

\section{Historical Evolution of the Web}\label{sec:history}

\begin{figure}[t]
\centering
\begin{tikzpicture}[scale=0.8]

\draw[thick, -stealth] (0,0) -- (18,0) node[right] {Time};

\shade[left color=blue!20, right color=green!20] (5.5,0) rectangle (6,3);     
\shade[left color=green!20, right color=red!20] (11.5,0) rectangle (12,3);    

\fill[blue!20] (0.5,0) rectangle (5.5,3);     
\fill[green!20] (6,0) rectangle (11.5,3);     
\fill[red!20] (12,0) rectangle (17.5,3);      

\draw[dashed] (5.5, 0) -- (5.5, 3);
\draw[dashed] (6, 0) -- (6, 3);
\draw[dashed] (11.5, 0) -- (11.5, 3);
\draw[dashed] (12, 0) -- (12, 3);

\node[font=\tiny] at (5.75,3.2) {Transition};
\node[font=\tiny] at (11.75,3.2) {Transition};

\node[font=\bfseries] at (3.0,2.6) {PC Web Era};
\node[font=\bfseries] at (8.75,2.6) {Mobile Web Era};
\node[font=\bfseries] at (14.75,2.6) {Agentic Web Era};

\node at (3.0,-0.5) {Start: 1990s};
\node at (8.75,-0.5) {Rise: Late 2000s};
\node at (14.75,-0.5) {Start: 2025};

\node[font=\small] at (3.0,2) {Search Paradigm};
\node[font=\small] at (8.75,2) {Recommendation Paradigm};
\node[font=\small] at (14.75,2) {Action Paradigm};

\node[font=\tiny, text width=3cm, align=center] at (3.0,1.2) {
    PageRank\\
    Keyword Matching\\
    Static Pages
};
\node[font=\tiny, text width=3.3cm, align=center] at (8.75,1.2) {
    Recommender Systems\\
    \hspace{2em}\textit{(emerging since 1990s)}\\
    Behavioral Analysis\\
    Personalization
};
\node[font=\tiny, text width=3.3cm, align=center] at (14.75,1.2) {
    Multi-Agent Systems\\
    AI Orchestration\\
    Agent Protocols
};

\node[font=\tiny, text width=3cm, align=center] at (3.0,0.4) {
    PPC Advertising\\
    Search Marketing
};
\node[font=\tiny, text width=3.3cm, align=center] at (8.75,0.4) {
    Feed Advertising\\
    Attention Economy
};
\node[font=\tiny, text width=3.3cm, align=center] at (14.75,0.4) {
    Agent Attention\\
    Service Competition
};

\draw[-stealth, thick, blue] (5.8,1.5) -- (6.2,1.5);
\draw[-stealth, thick, green] (11.3,1.5) -- (11.7,1.5);

\end{tikzpicture}
\caption{Timeline of Web Evolution. The three eras of web evolution are not strictly distinct. Their transitions happened gradually, with technologies, features, and business models often overlapping and coexisting across different periods.}
\label{fig:web_timeline}
\end{figure}
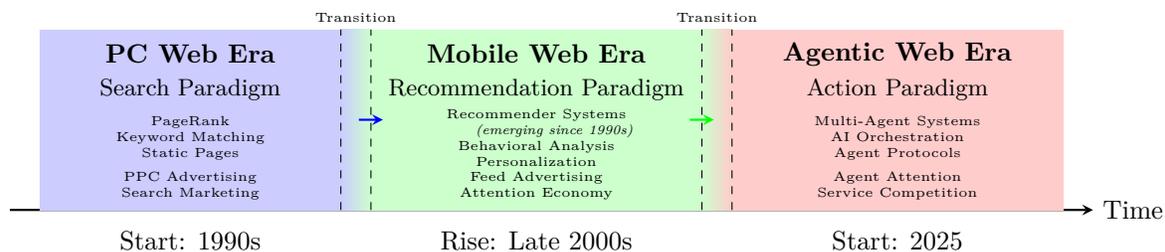

In this section, a chronological review of three milestone phases in the evolution of the Web is conducted: the \textit{PC Web Era}, the \textit{Mobile Web Era}, and the \textit{Agentic Web Era}. This progression is visualized in Figure~\ref{fig:web_timeline}, which presents a high-level timeline of the Web’s evolution across technological paradigms and business models.

\begin{figure}[t]
\centering
\begin{tikzpicture}

\begin{scope}[yshift=4cm]
\node[font=\bfseries] at (-1,0) {PC Web:};
\node[circle, draw, fill=blue!20] (user1) at (1,0) {User};
\node[rectangle, draw] (search1) at (3,0) {Search};
\node[rectangle, draw] (ads1) at (5,0) {Ads};
\node[rectangle, draw] (auction1) at (7,0) {Ad Auction};

\draw[-stealth, thick] (user1) -- (search1) node[midway, above, yshift=5pt] {\tiny Query};
\draw[-stealth, thick] (search1) -- (ads1) node[midway, above, yshift=5pt] {\tiny Result};
\draw[-stealth, thick] (ads1) -- (auction1) node[midway, above, yshift=5pt] {\tiny Auction};
\end{scope}

\begin{scope}[yshift=2cm]
\node[font=\bfseries] at (-1,0) {Mobile Web:};
\node[circle, draw, fill=green!20] (user2) at (1,0) {User};
\node[rectangle, draw] (algo2) at (3,0) {Algorithm};
\node[rectangle, draw] (feed2) at (5,0) {Feed};
\node[rectangle, draw] (engage2) at (7,0) {Engage};

\draw[<-stealth, thick] (user2) -- (algo2) node[midway, above, yshift=5pt] {\tiny Data};
\draw[-stealth, thick] (algo2) -- (feed2) node[midway, above, yshift=5pt] {\tiny Curate};
\draw[-stealth, thick] (feed2) -- (engage2) node[midway, above, yshift=5pt] {\tiny Consume};
\end{scope}

\begin{scope}[yshift=0cm]
\node[font=\bfseries] at (-1,0) {Agentic Web:};
\node[circle, draw, fill=red!20] (user3) at (1,0) {User};
\node[rectangle, draw] (agent3) at (3,0) {Agent};

\node[rectangle, draw, fill=purple!20] (service1) at (5,1) {Service A};
\node[rectangle, draw, fill=purple!20] (service2) at (5,0) {Service B};
\node[rectangle, draw, fill=purple!20] (service3) at (5,-1) {Service C};

\node[rectangle, draw] (execute3) at (7,0) {Execute};

\draw[-stealth, thick] (user3) -- (agent3) node[midway, above, yshift=5pt] {\tiny Task};
\draw[-stealth, thick] (agent3) -- (service1) node[midway, above left] {\tiny Query};
\draw[-stealth, thick] (agent3) -- (service2);
\draw[-stealth, thick] (agent3) -- (service3) node[midway, below left] {\tiny \textcolor{red}{Select}};
\draw[-stealth, thick] (service2) -- (execute3) node[midway, above, yshift=7pt] {\tiny Action};

\draw[<->, dashed, red] (service1) -- (service2) node[midway] {\tiny Compare};
\draw[<->, dashed, red] (service2) -- (service3) node[midway] {\tiny Compose};
\end{scope}

\end{tikzpicture}
\caption{Attention Flow Evolution Across Web Eras. This diagram illustrates the transition from the PC Web, where attention follows a linear search-query-ad model, to the Mobile Web, where algorithmic systems curate feeds based on user data, and finally to the Agentic Web, where autonomous agents interpret user intent and select among competing services to execute tasks.  Dashed arrows in the agentic stage indicate competitive or compositional relationships between services.}
\label{fig:attention_flow}
\end{figure}
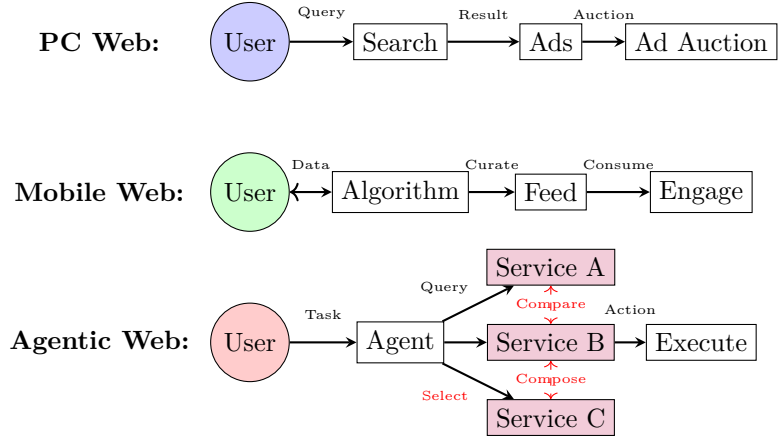

Each era is characterised by significant shifts in technological paradigms, commercial models, and user behavior patterns. 
The \textit{PC Web Era} was centred around information directories and search paradigms, with content organized through static web pages that users manually browsed to locate desired information. Search engines emerged to support efficient retrieval, and keyword-based advertising systems marked the beginning of the commercial Web. The subsequent \textit{Mobile Web Era} introduced a fundamental shift toward recommendation-driven content consumption, where algorithmic curation became essential due to the explosion of user-generated content and mobile platform constraints. Today, the Web is entering the \textit{Agentic Web Era}, propelled by breakthroughs in foundation models and agent-based paradigms, where intelligent agents coordinate complex tasks and reshape both technical architecture and commercial logic. Figure~\ref{fig:attention_flow} illustrates how user attention flows have evolved across these Web eras, from linear search and ad delivery models, to algorithmic feed curation, and finally to agent-mediated task execution involving multiple competing services.

\subsection{PC Web Era}
The PC Web Era represents the foundational stage of the Internet’s evolution, marked by static content delivery and goal-oriented information retrieval. During this period, the user experience was shaped by limited interactivity, minimal personalization, and the early commercialization of the web through keyword-based search and advertising systems.

\subsubsection{Static Pages and Search-based Commercial Marketing}

The PC Web was dominated by a \textit{retrieval paradigm} characterized by users relying on active queries and manual browsing to access information in an era of rapidly expanding digital content. At this stage, the Web lacked intelligent mechanisms for information dispensing.  The content was primarily presented through static web pages with fixed organizational structures, with limited interactivity and personalised recommendations. Web platforms like Yellow Pages and Craigslist relied heavily on manual categorization and predefined navigation to link various types of information. These websites were typically organized by geography, industry, or service type, mirroring the taxonomy of printed directories and classified ads to present business listings, personal posts, and product information.

From the perspective of users, this retrieval paradigm required a strong sense of goal-directed behavior. The information-seeking process was linear and static, requiring users to have a clear goal for their search and to invest time and effort in locating the information they needed by navigating through hierarchical directories. This simple but inefficient paradigm struggled to meet users’ increasing demand for speed, relevance, and personalization. 

As the scale of the Web expanded rapidly, the conventional directory-based paradigm proved inadequate to satisfy the growing demand for efficient information retrieval.  To address this challenge, search engines emerged and became a critical turning point in the evolution of the Web.  Early systems relied on basic keyword matching techniques like TF-IDF \citep{tfidf}, which measured term frequency but struggled with document authority and relevance ranking. 
Building upon TF-IDF, more sophisticated probabilistic models such as BM25 \citep{robertson2009probabilistic} were developed to address issues like document length normalization and term frequency saturation, providing better text relevance scoring mechanisms. Meanwhile, Latent Semantic Indexing \citep{1990Indexing} introduced a paradigm shift by using singular value decomposition to capture latent semantic relationships between terms and documents, enabling search engines to understand conceptual similarities beyond exact keyword matches and address issues like synonymy and polysemy. 

A significant milestone was the PageRank algorithm \citep{page1999pagerank}, which pioneered the concept of ``link-based voting'' by evaluating the importance and authority of web pages through their hyperlink structures.  In comparison with earlier methods that relied solely on keyword matching, PageRank significantly enhanced the relevance of search results, laid the foundation for search engines like Google, and greatly increased user reliance on search engines.

Subsequent to this technological breakthrough, search engines integrated advertising mechanisms based on user intent. Early search advertising systems, such as Google AdWords, matched user queries with commercial content through sophisticated auction algorithms. The evolution from simple ``pay-your-bid'' mechanisms used by early systems like Overture to more sophisticated auction theories became crucial. Google AdWords implemented the Generalized Second-Price auction algorithm \citep{SecondPriceAuction}, where advertisers bid for keyword placement but pay the price of the next-highest bidder, creating more stable and efficient bidding behavior than earlier first-price auction systems. 

The introduction of Quality Score further refined this mechanism by balancing bid amounts with ad relevance, rewarding high-quality advertisements with better positions and lower costs. This keyword-driven, pay-per-click (PPC) model enhanced ad conversion rates whilst providing sustainable revenue streams for search engines, establishing a direct link between web content and commercial marketing and ultimately initiating the commercialization of the Web based on search.

\subsection{Mobile Web Era}
The transition to the Mobile Web Era was driven by fundamental changes in the Web's information landscape that extended beyond the mere adoption of mobile devices.

The most significant driver was the explosive growth in content volume during the late PC Web era. User-Generated Content proliferated across social platforms, e-commerce sites, and streaming services, creating massive data flows that traditional search paradigms struggled to navigate effectively. Users found themselves overwhelmed by choice and increasingly unable to discover relevant content through manual search alone.

This content explosion coincided with a shift in user behavior from intent-driven to discovery-driven consumption. Rather than approaching the Web with specific queries, users increasingly sought serendipitous discovery and personalized exploration. Mobile contexts amplified this trend, as users consumed content in shorter, fragmented sessions while expecting instant, personalized experiences without active searching.

Recommender Systems, which had already existed during the PC Web era for specific use cases, thus evolved from auxiliary tools to central architectural components. Mobile platforms introduced distinct design challenges such as latency constraints, limited screen space, and fragmented user attention. These challenges catalyzed advancements in recommender system architectures, promoting the development of real-time, context-aware models tailored to mobile interaction patterns.

\subsubsection{Recommender Systems}
The progression of Recommender Systems mirrors the broader technological transitions across web platforms. In the early era of the PC Web, Recommender Systems primarily relied on collaborative filtering techniques to provide personalized suggestions based on historical user-item interactions. User-based and item-based k-nearest neighbor models were widely adopted due to their simplicity and interpretability \citep{sarwar2001item}, while matrix factorization approaches became mainstream for uncovering latent preferences and item attributes. Although foundational, these methods suffered from scalability and sparsity challenges, especially as web content and user bases expanded rapidly. Early solutions like Singular Value Decomposition in latent factor models \citep{koren2009matrix} were instrumental in establishing the core technical underpinnings of modern recommender systems. However, their static nature and inability to model complex behavior made them increasingly inadequate in dynamic environments.
A critical bridge between traditional methods and modern deep learning was established through Factorization Machines \citep{rendle2010factorization}, which modeled feature interactions through factorized parameters and enabled efficient computation with sparse data. This approach addressed the limitation of traditional matrix factorization in handling diverse feature types and became a foundation for subsequent hybrid architectures.

As user interaction became more real-time and context-rich in the Mobile Web era, significant advancements in personalization capabilities emerged. Deep learning-based methods became powerful means of modeling high-dimensional, nonlinear interactions between users and items. Neural collaborative filtering \citep{he2017neuralcollaborativefiltering} combined deep neural networks with matrix factorization to enhance generalization capabilities. AutoRec \citep{AutoRec} introduced autoencoder-based collaborative filtering, demonstrating the potential of neural architectures for recommendation tasks.

The evolution toward industrial-scale mobile applications led to breakthrough hybrid architectures that balanced memorization and generalization. Wide \& Deep Learning \citep{cheng2016widedeeplearning}, developed by Google, combined linear models for memorization with deep neural networks for generalization, establishing a paradigm for large-scale recommender systems. Building upon this foundation, DeepFM \citep{guo2017deepfmfactorizationmachinebasedneural} integrated factorization machines with deep neural networks for click-through rate prediction, eliminating the need for manual feature engineering while maintaining the ability to model both low-order and high-order feature interactions.
Advanced deep learning architectures further enhanced modeling capabilities for mobile environments. Deep Matrix Factorization \citep{De_Handschutter_2021} extended latent modeling capacity using residual learning, while DeepCF \citep{deng2019deepcfunifiedframeworkrepresentation} introduced unified architectures that integrated user/item representations with content signals. These advancements enabled mobile applications to deliver fine-grained, real-time personalization across social media, e-commerce, and streaming platforms.

The incorporation of temporal dynamics became crucial for mobile environments where user behavior patterns change rapidly. Sequence-aware models such as GRU4Rec \citep{hidasi2016sessionbasedrecommendationsrecurrentneural} emerged to model temporal patterns in user behavior, while contextual bandits addressed the exploration-exploitation trade-off in real-time recommendation scenarios. The introduction of attention mechanisms and Transformer architectures \citep{vaswani2017attentionneed} marked a significant advancement in sequential recommendation, enabling models to capture long-range dependencies and complex interaction patterns.
Contemporary developments have focused on attention-based models and self-supervised learning approaches optimized for mobile contexts. Transformer-based architectures like SASRec \citep{kang2018selfattentivesequentialrecommendation} and BERT4Rec \citep{sun2019bert4recsequentialrecommendationbidirectional} have demonstrated superior performance in sequential recommendation tasks by leveraging self-attention mechanisms to model user behavior sequences. Graph Neural Networks have also emerged as powerful tools for modeling complex user-item interactions and social relationships \citep{Wang_2019}.

\subsubsection{Attention Economy}

In the \textit{Mobile Web Era}, Recommender Systems have not only transformed the manner in which users access information but also significantly impacted commercial patterns. By leveraging user interests and behavioral data, Recommender Systems enable advertising and e-commerce platforms to precisely target potential consumers with personalised content. For instance, e-commerce platforms analyse users' browsing histories and purchase records to suggest relevant products, thus enhancing the shopping experience and significantly boosting conversion rates and sales. Similarly, social media platforms employ Recommender Systems to present engaging content on homepages or feeds, thereby increasing user retention and interaction.

The advent of Recommender Systems precipitated a substantial evolution in the realm of online advertising, characterised by enhanced targeting accuracy and the emergence of a behavior-driven advertising pattern. This paradigm shift has given rise to the so-called \textit{attention economy}~\citep{FALKINGER2007266,Ciampaglia_2015,10.1145/376625.376626} where in each user action, such as a click, scroll, or pause, is considered a valuable data point. These platforms utilise the gathered data to enhance the delivery of advertisements in terms of format, timing, and frequency, thereby rendering them more relevant and cost-effective. This behavior-based approach enables advertisers to achieve higher marketing efficiency at lower costs.

Overall, Recommender Systems have improved both the efficiency and personalization of information access while emerging as a pivotal force in the commercialization of the Web. Through intelligent content distribution, they have redefined commercial interactions and consumption patterns. Their widespread adoption in the \textit{Mobile Web Era} signals a shift from a supply-demand model to a more intricate, behavior-driven interaction paradigm between information and commerce.

\subsection{Agentic Web Era}
The evolution of the Web is undergoing a paradigm shift from a human-centric information retrieval model to an agent-centric action-oriented framework. 

The foundation for this transition was established by breakthroughs in LLMs~\citep{openai2024gpt4o,dubey2024llama,deepseekai2025deepseekr1incentivizingreasoningcapability}, which demonstrated unprecedented capabilities in natural language understanding, reasoning, and code generation. Unlike previous AI systems that were limited to specific domains, these models exhibited emergent abilities to decompose complex tasks, maintain context across extended interactions, and interface with external tools and APIs. 

This technological leap enabled the development of AI agents capable of autonomous decision-making and multi-step task execution, ushering in the Agentic Web Era.  This era is characterized by the orchestration of intelligent agents across a vast network of capabilities, protocols, and data sources.  In the following sections, we examine the rise of this agent-based infrastructure and its profound implications on Web architecture and digital economics.

\begin{figure}[t]
    \centering
    \includegraphics[width=1\linewidth]{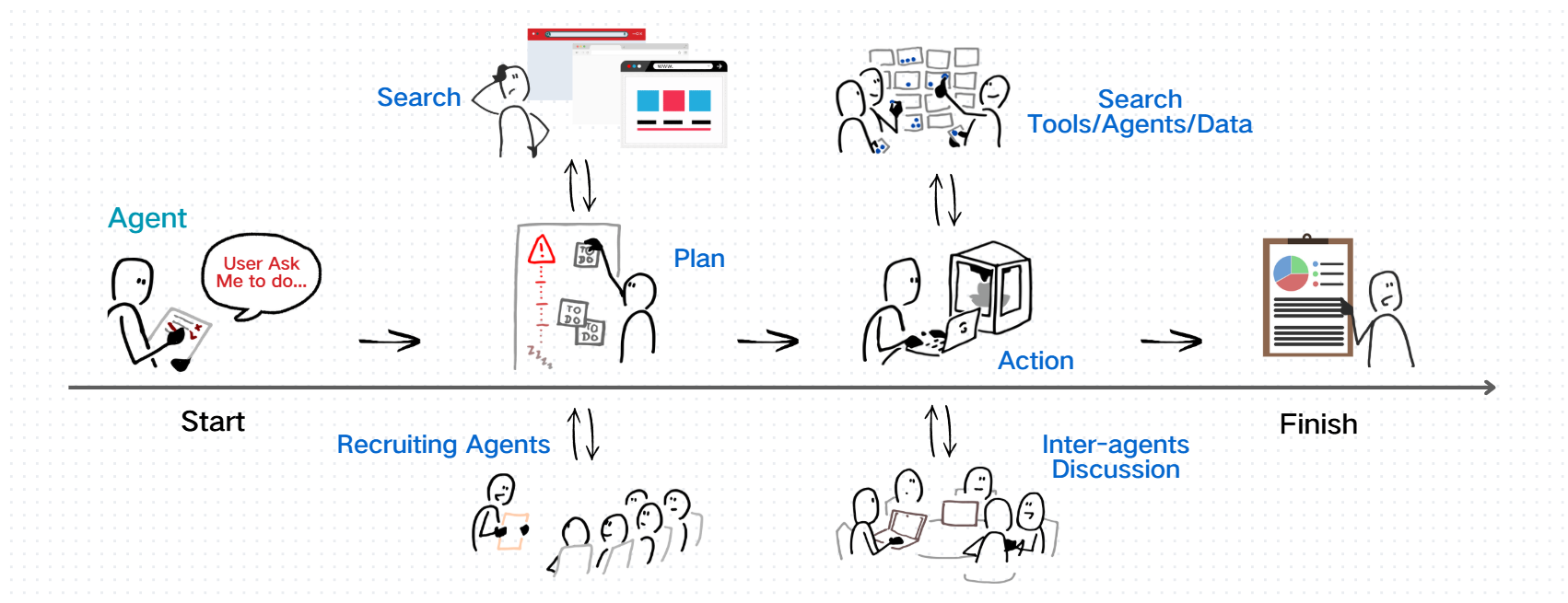}
    \caption{Agent Workflow under the Agentic Web.}
    \label{fig:agentic-workflow}
\end{figure}

\subsubsection{Rise of Agentic Web}
In the Agentic Web Era, tasks that once required significant human effort, such as deep research, cross-platform process execution, and long-term goal management, can now be completed autonomously with the help of intelligent agents. These tasks demand a comprehensive understanding of context, the ability to dynamically adjust strategies, and the integration of multiple tools and data sources.

Traditional single-function assistants are increasingly inadequate for addressing the demands of complex, multi-step tasks. In response, multi-agent systems have emerged as a critical architectural solution.
A multi-agent system comprises multiple autonomous agents, each specialized in a specific sub-task such as information retrieval, translation, computation, or API interaction. These agents collaborate through mechanisms including task decomposition, capability scheduling, and inter-agent information sharing, thereby enabling the system to tackle problems of greater complexity and scale.
The development of AI orchestration frameworks, such as LangChain~\citep{chen2023langchain} and AutoGen~\citep{wu2023autogen}, supports the integration of models, tools, and service components into structured \textit{graph task flows}. These flows facilitate coordinated execution and dynamic decision-making among agents in a flexible and modular fashion, as illustrated in Figure~\ref{fig:agentic-workflow}.
This collaborative paradigm significantly enhances both the breadth and depth of task execution. As a result, the Web is undergoing a transformation from a static ``information network'' to a dynamic ``action network'' in which autonomous systems are capable of perceiving, reasoning, and acting within digital environments.

However, this shift to multi-agent systems brings with it the need for sophisticated infrastructure to support these complex, agent-driven tasks. The advancement of the Agentic Web is not purely conceptual, it is increasingly grounded in real-world infrastructure developments.
A key example of this transformation is Microsoft's recent move towards the Agentic Web, announced at Build 2025.
Over 50 AI agent tools have been integrated across platforms such as GitHub, Azure, Microsoft 365, and Windows. These tools support every stage of agent deployment, from development and orchestration to memory management and inference optimization. They offer a unified environment for building and operating intelligent agents.
Additionally, NLWeb (Natural Language Web)~\citep{msblog2025nlweb} has emerged as a toolset designed to convert traditional web interfaces into agent-readable, structured environments. This enables agents to navigate and interact with websites in a goal-driven manner, rather than relying on outdated and brittle methods like DOM scraping or simulated clicks.

These advancements signal a significant change in the way the Web will be used in the future. It is no longer just a place for human-centric browsing but a dynamic, agent-driven environment where intelligent agents can perform complex tasks on behalf of users. In this context, agent protocols like the MCP~\citep{anthropic2025} are playing a crucial role in enabling communication and collaboration between various services and agents. The establishment of these protocols is laying the foundation for a more standardized, scalable agent-Web interaction layer, further advancing the Agentic Web.

\subsubsection{Agent Attention Economy}
The advent of communication protocols such as the MCP has precipitated a trend of standardisation, registration, and organization of external tools and services within the agent ecosystem, which is analogous to the ``yellow pages'' directory. This facilitates agents' access to and invocation of these resources through a unified interface. However, as the number of external resource providers, such as APIs, remote services, and data endpoints, grows rapidly, a new challenge emerges: how can agents efficiently discover, filter, and select the most suitable capabilities in a highly fragmented and dynamic service landscape~\citep{yang2025agentexchangeshapingfuture}?

This challenge gives rise to the notion of the \textbf{Agent Attention Economy}. In a manner analogous to the early Web's competition for user clicks, external services now compete to be selected and invoked by autonomous agents. In this paradigm, the focus shifts from the human users to the agent engaged in the execution of a sophisticated task. Every tool, service, or other agent essentially competes for limited ``agent attention''. To improve visibility and invocation likelihood, these entities may adopt mechanisms such as advertising, ranking optimization, or even agent-oriented recommendation and scoring systems within the service registries.

As this competition intensifies, it is reasonable to hypothesise that a comprehensive advertising infrastructure tailored for agents will emerge, which will include agent-facing recommendation engines, capability reranking systems, inter-agent referral networks, and potentially auction-based ranking or context-aware ad insertion. This shift fundamentally redefines how agents discover and coordinate with external resources, accelerating the transformation from a human-centric to an agent-centric Web. 
This attention-based competition among agents may ultimately become a core mechanism for resource allocation in future Web platforms, signalling a profound restructuring of both the architectural and economic foundations of the Web.

\begin{table}[t]
\centering
\caption{A cross-era comparison of Web paradigms from an ecosystem perspective.}
\label{tab:web_evolution}
\renewcommand{\arraystretch}{1.4}
\resizebox{\textwidth}{!}{
\begin{tabular}{p{4cm}p{4cm}p{4cm}p{4cm}}
\toprule
\textbf{Aspect} & \textbf{PC Web Era} & \textbf{Mobile Web Era} & \textbf{Agentic Web Era} \\
\hline
\textbf{Core Paradigm} & Search Paradigm & Recommendation Paradigm & Action Paradigm \\
\hline
\textbf{User Behavior} & Active search and manual browsing & Passive content consumption & Complex multi-step task execution \\
\hline
\textbf{Information Organization} & Static pages, hierarchical directories & Personalized feeds, algorithmic curation & Dynamic task flows, multi-agent collaboration \\
\hline
\textbf{Key Technologies} & 
PageRank algorithm,
 Keyword matching,
 Directory structures
 & 
Recommender Systems,
 Behavioral analysis,
Personalization algorithms
& 
 Multi-agent systems, AI orchestration frameworks, AI Agent protocols (MCP/A2A)
 \\ 
\hline
\textbf{Commercial Model} & Pay-per-click advertising & Feed-based and in-app advertising & Agent Attention Economy \\
\hline
\textbf{Revenue Source} & Search advertising (e.g., Google AdWords) & Targeted advertising, e-commerce integration & Service invocation fees, agent-targeted advertising \\
\hline
\textbf{Key Metrics} & 
 Click-through Rate,
Cost Per Click
 & 
 Conversion Rate,
 User dwell time,
 Effective cost-per-thousand impressions
& 
Service invocation frequency,
 Capability relevance,
Agent response success rate
 \\
\hline
\textbf{Attention Focus} & Human user clicks & Human user engagement & Agent selection and invocation \\
\hline
\textbf{Information Access} & Goal-directed, linear search & Algorithm-driven, passive consumption & Context-aware, multi-step execution \\
\hline
\textbf{Platform Examples} & Yellow Pages, Craigslist, Google Search & Social media feeds, e-commerce recommendations & Multi-agent AI systems, service registries \\
\hline
\textbf{Economic Foundation} & Search-based marketing & Attention economy & Agent-centric resource allocation \\
\bottomrule
\end{tabular}}
\vspace{-0.4cm}
\end{table}

\subsection{Commercial and Structural Evolution of the Web}
The Web has evolved through three distinct eras: the PC Web, the Mobile Web, and the Agentic Web. Table~\ref{tab:web_evolution} provides a comparative overview of how the Web’s architecture, attention focus, and commercial models have evolved across eras. This section analyzes how innovations such as PageRank, recommender systems, and agent protocols like MCP have transformed the architecture and commercial dynamics of the Web, laying the foundation for an agent-native ecosystem in which information is dynamically generated and acted upon by autonomous systems.

In the \textit{PC Web Era}, information was primarily hosted on static web pages, often published by institutions and accessed through keyword-based search engines. Discovery hinged on link analysis algorithms such as PageRank, and content was structured like a digital directory, manually navigable and hierarchically classified. Web content was sparse and predominantly produced by organizations, meaning it was centralized and often top-down. Users relied on manual browsing or search engines to retrieve information by typing keywords, and the Web’s structure was akin to a digital directory, where content was classified through hyperlinks and explicit taxonomies. Commercial activity centered around search advertising, with platforms like Google AdWords matching user queries to paid results. Key performance indicators (KPIs) included click-through rate and cost-per-click, reflecting a model where user attention was explicitly captured and monetized through intent-based queries.

As the Web transitioned into the \textit{Mobile Web Era}, the underlying data storage and linking paradigm remained largely unchanged, but the volume and granularity of content exploded, driven by the rise of UGC on social platforms, e-commerce sites, and streaming services. While information storage mechanisms such as cloud-based servers remained similar to earlier times, the sheer volume of content made traditional retrieval methods increasingly inefficient. Search interfaces, though still available, struggled to surface relevant content amidst massive data flows. This explosion in information created a need for more sophisticated ways of navigating and discovering content. In response, Recommender Systems emerged as the dominant access paradigm. These systems relied on algorithms to curate content tailored to individual users, shifting the burden of content discovery from users to algorithms. Users became both producers and consumers of content, with recommendation algorithms acting as intermediaries. Commercial models adapted to this new logic, emphasizing metrics like conversion rate, dwell time, and effective cost per mille, highlighting engagement depth and monetization precision.

Today, in the emerging \textit{Agentic Web Era}, both the commercial logic and the structure of information are undergoing a radical transformation. Unlike earlier paradigms, where knowledge was stored in databases or presented on web pages, LLMs embed vast amounts of web-scale information within their parameters through pretraining. This in-parameter knowledge enables LLM-based agents to reason, and respond based on learned representations rather than direct document lookups. In parallel, intelligent agents are developing the ability to interact not only with web pages, but also with other agents, APIs, and tools in a dynamic, goal-driven fashion. The Web is thus shifting from a passive content repository to an active, agent-mediated action space, where agents act as autonomous intermediaries executing tasks on behalf of users.

Commercially, this era means the rise of the \textit{Agent Attention Economy}, where third-party tools and services compete not for human clicks, but for agent invocation. New protocols like the MCP are enabling agents to dynamically compose and orchestrate services from a modular ecosystem, leading to the development of agent-driven commercial models. Future monetization metrics may depend on factors such as invocation frequency, capability relevance, and successful task completion, marking a departure from the previous era's focus on user interaction and direct advertising. This shift is giving rise to new forms of agent-oriented advertising and bidding systems, designed not to persuade users directly, but to influence agent decision-making pipelines.

Structurally, the Web is being redefined from a human-readable medium into an agent-native substrate.  On the consumption side, agents can proactively summarize, recommend, or execute tasks on behalf of users, offering unprecedented personalization and efficiency.  These agents are capable of operating continuously across different services, becoming autonomous digital intermediaries.  On the production side, content will increasingly be generated by agents rather than humans.  Agents can autonomously generate articles, compose marketing material, or structure data for other agents to consume.  This results in an emergent agent-to-agent communication layer, where content may never be explicitly rendered for human eyes, but is optimized for agent parsing, reasoning, and orchestration.

Taken together, these shifts point toward a profound reconfiguration of the Web’s ontology: from human-readable documents, to algorithm-curated feeds, to agent-native knowledge.  No longer will the Web simply serve as a repository of static documents or a curated feed of personalized content, but rather as a dynamic, interactive space where information flows are synthesized, shared, and executed by autonomous systems on behalf of humans.  Over time, this could lead to the rise of an “Agent-Oriented Web,” where information is not merely personalized for individual users, but is dynamically created, shared, and executed through collaboration between intelligent agents.

\section{The Agentic Web}\label{sec:agentic-web}
\vspace{-0.2cm}
The Web is undergoing a fundamental transformation~\citep{petrova2025semanticwebmasagentic, lù2025buildwebagentsagents,chaffer2025know}. In the traditional model, users served as active navigators: searching, comparing, and manually executing each digital step. Booking a flight, for instance, required visiting multiple travel websites, comparing ticket options, checking loyalty programs, and handling confirmation emails across services.
With the rise of intelligent agents, this burden is shifting. Users now increasingly delegate goals rather than execute tasks. A travel agent AI can autonomously search for optimal flights based on personal calendar availability, loyalty points, and real-time pricing. It can coordinate with hotel agents or even adjust travel plans based on weather forecasts or meeting changes~\citep{manus2024agent,genspark2025superagent}. This represents a shift from user-driven web navigation to intent-driven orchestration, where outcomes rather than page views become the primary metric of value.

This evolution mirrors a broader arc in the history of the Internet. In the PC Web Era, users manually navigated hyperlinks like explorers in a vast library. In the Mobile Web Era, apps curated the experience and brought information to users more proactively. Yet users still had to operate the system by opening apps, copying data, and making decisions. Now, in the Agentic Web, users act more like directors who articulate their intent while intelligent agents carry out the necessary operations behind the scenes.

These transformations are not merely technological but conceptual.    They redefine who acts on the Web (from human to agent), how tasks are executed (from manual interaction to delegated orchestration), and what the Web produces (from content consumption to outcome generation).  In the following sections, we propose a structured framework to examine this shift across 3 dimensions: Intelligence, Interaction, and Economy. Each dimension offers insight into how the Agentic Web is reshaping capabilities, behaviors, and business models in increasingly autonomous digital ecosystems.

\begin{figure}[t]
    \centering
    \includegraphics[width=0.9\linewidth]{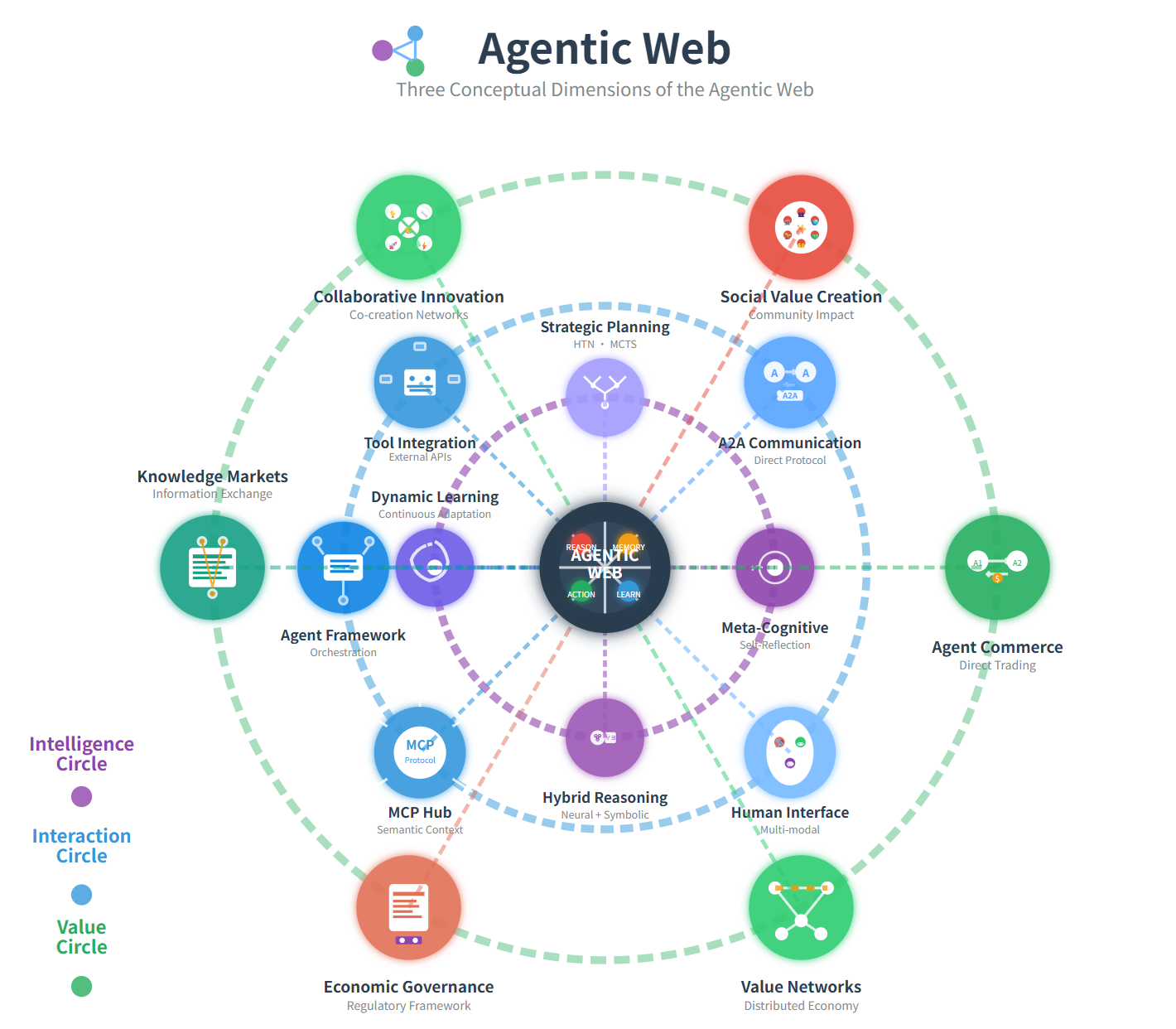}
    \vspace{-0.1cm}
    \caption{Conceptual Framework of the Agentic Web. This diagram illustrates a three-dimensional architecture composed of the Intelligence, Interaction, and Economic Dimensions, reflecting the evolution of AI agents from reasoning entities to active economic participants.}
    \label{fig:agentic-framework} 
    \vspace{-0.3cm}
\end{figure}

\vspace{-0.4cm}
\subsection{Core Conditions}
\vspace{-0.3cm}

Building the Agentic Web requires rethinking fundamental elements of digital infrastructure. Three conditions are essential:
\begin{tcolorbox}[conceptrelation]
\begin{itemize}
\item[(1)] Agents must function as autonomous intermediaries, initiating and completing complex tasks independently.
\item[(2)] Web resources need to be accessible through standardized, machine-readable interfaces.
\item[(3)] Value must be exchanged not only between humans and systems, but also directly between agents.
\end{itemize}
\end{tcolorbox}

These structural foundations are interdependent. Agent autonomy depends on semantic interfaces, protocol interoperability, and the ability to discover and orchestrate external capabilities dynamically. Together, they create the conditions for scalable, intelligent web operations.

\subsection{Transformations in Web Architecture}
These structural foundations enable fundamental shifts in the operation of the Web, transforming both its usage and the organization of information.

\subsubsection{Evolving Interaction Patterns}
The Agentic Web transforms how interaction occurs in digital environments. Traditional web use follows a request-response model, where users initiate actions, retrieve data, and evaluate results manually. Agents, by contrast, engage in proactive behaviors. They discover relevant resources, identify capabilities, and form dynamic connections based on semantic relevance rather than static hyperlinks~\citep{tupe2025aiagenticworkflowsenterprise,sapkota2025ai,AcharyaAgenticAI}.

This change supports continuous and goal-oriented interaction across services. Agents monitor the digital environment, detect opportunities, and collaborate with other systems to fulfill objectives. Instead of navigating predefined pathways, they identify and access web resources through contextual understanding and adaptive negotiation, resulting in more responsive and flexible connectivity.

\subsubsection{Changing Information Structures}
The structural transformation of the Agentic Web is not limited to the role of agents as users or interfaces; it also extends to how information itself is stored, linked, and consumed. In previous stages of the Web, including the early PC era and the mobile era, information was mainly organized as documents or datasets hosted on web servers and accessed via hyperlinks or queried through search engines.

By contrast, the Agentic Web introduces a new mode of information organization. LLMs capture vast web-scale information directly within their parameters through the process of pretraining. These models embed knowledge within their structures and support on-demand reasoning, thereby reducing reliance on traditional external web sources~\citep{petrova2025semanticwebmasagentic, lù2025buildwebagentsagents}.

At the same time, the way information is linked is also changing. Rather than relying on static hyperlinks, agents discover resources, services, and other agents through semantic discovery and adaptive integration~\citep{touvron2023llama2,openai2024GPT,deepseekai2025deepseekr1incentivizingreasoningcapability}. This results in a more fluid and context-sensitive form of connectivity, where relevance is inferred rather than explicitly declared.

As agents gain content generation capabilities, information production is shifting beyond human-authored web pages toward agent-generated outputs such as tools, instructions, summaries, and structured artifacts designed for other agents~\citep{qin2023toolllm,schick2023toolformer}. This creates a self-sustaining loop in which agents both generate and consume content, leading to increasingly autonomous and self-reinforcing information flows.

This structural shift in how information is stored (in-model versus in-document), linked (semantic versus hyperlink), and accessed (agent-driven versus search-driven) underpins the transition to the Agentic Web.

\subsubsection{Dual Operational Roles}

Within this transformed architecture, agents operate through two complementary foundational perspectives that represent different facets of their functionality:
\begin{itemize}
    \item \textbf{Agent-as-User (Downward-facing):} AI agents operate as autonomous web users who can independently navigate, interact with, and consume web resources~\citep{nakano2022webgptbrowserassistedquestionansweringhuman,deng2023mind2web,zhou2023webarena,gur2024a,openai2024agent, manus2024agent}. In this role, agents replace or augment human users in web navigation and task execution, engaging with existing web interfaces and services designed for human consumption. This enables continuous, 24/7 operation for tasks such as market research, data collection, or transaction processing.
    \item \textbf{Agent-as-Interface (Upward-facing):} AI agents serve as intelligent intermediaries between human users and web systems, translating high-level user intentions into executable actions~\citep{Microsoft2025Copilot,browsercompany_dia_thurrott,opera_neon_blog,wriggers2025techcrunch}. These agents process natural language commands from users and orchestrate complex multi-step workflows across various web services. This perspective emphasizes the agent's role in abstracting complexity and providing streamlined human-agent interaction.
\end{itemize}

These perspectives are complementary rather than contradictory. A single agent system often embodies both roles: interacting autonomously with the web while serving as an interface for human users and forming a bidirectional bridge between intention and execution. Together, these operational perspectives, interaction transformations, and information structure changes operationalize the Agentic Web's core vision: a distributed, interactive internet ecosystem in which autonomous software agents engage in persistent, goal-directed interactions to plan, coordinate, and execute tasks on behalf of human users.

\subsection{Three Conceptual Dimensions of the Agentic Web}
\label{sec:three-dimension}

To understand the Agentic Web in depth, we propose a conceptual framework built on three interrelated dimensions: Intelligence, Interaction, and Economy. Each dimension reflects a core requirement for autonomous operation within digital ecosystems as illustrated in Figure~\ref{fig:agentic-framework}).

At its core, the Intelligence Dimension equips agents with reasoning capabilities such as perception, planning, and learning. Building on this, the Interaction Dimension enables agents to connect with digital environments through semantic protocols and dynamic tool use. The Economic Dimension focuses on how agents autonomously create, exchange, and distribute value, forming self-organizing digital economies.

Each layer builds upon the previous one: intelligence enables interaction, and interaction enables value creation. This layered view explains how agents evolve from internal reasoning entities to impactful economic participants.

\begin{tcolorbox}[keyterminology]
\begin{itemize}
\item \textbf{Intelligence Dimension:} What core intelligence is required for agents to function autonomously?
\item \textbf{Interaction Dimension:} How do agents communicate and coordinate within digital ecosystems?
\item \textbf{Economic Dimension:} How do agents generate and exchange value at scale?
\end{itemize}
\end{tcolorbox}

\subsubsection{Intelligence Dimension}
The Intelligence Dimension provides the cognitive foundation that enables agents to reason, learn, and plan within open-ended digital environments. Unlike traditional systems that rely on human queries, agents access and act on information autonomously. They draw from both internalized knowledge (in-parameter models)~\citep{koh2024visualwebarena,putta2024agentqadvancedreasoning,masterman2024landscape,wu2025foundationsrecenttrendsmultimodal} and external resources (via tools and APIs)~\citep{qin2023toolllm,schick2023toolformer,paranjape2023art,lu2025octotools}, shifting from passive retrieval to proactive information use.

To operate effectively, agents require transferable intelligence rather than narrowly defined, task-specific heuristics. Key capabilities include:
\vspace{-0.2cm}
\begin{itemize}
    \item \textbf{Contextual Understanding:} Agents should be able to interpret diverse forms of web-based input, including natural language, semi-structured data, and interface signals, all within task-specific and evolving contexts.
    \item \textbf{Long-Horizon Planning:} Agents must formulate, evaluate, and revise multi-step strategies to achieve both short-term objectives and long-term goals across diverse digital services.
    \item \textbf{Adaptive Learning:} Agents should be able to improve over time by integrating interaction feedback, acquiring new skills, and adjusting their internal models of user preferences and environment dynamics.
    \item \textbf{Cognitive Processes:} To operate reliably and efficiently, agents should monitor and reflect on their own reasoning, detect failures or suboptimal behavior, and dynamically adjust their cognitive strategies.
    \item \textbf{Multi-Modal Integration:} Agents must handle and integrate information from a variety of modalities (e.g., text, APIs, visuals, structured data), enabling robust decision-making in open-ended environments.
\end{itemize}

In the Agentic Web, agents are not passive executors of instructions. Instead, they interpret, strategize, and adapt independently. Without these cognitive abilities, agents cannot handle real-world ambiguity, recover from failure, or scale across services.

\subsubsection{Interaction Dimension}
The Interaction Dimension addresses a fundamental shift in how autonomous agents engage with the digital environment. This shift moves the web away from static, human-authored hyperlinks toward dynamic, context-aware connections. In traditional web architectures, interaction is primarily document-based and mediated by human users. In contrast, agents in the Agentic Web interact through semantic protocols and runtime service discovery, allowing them to initiate and manage interactions without relying on predefined links or manual input.

The emergence of agent-native communication protocols has been a key catalyst for this transformation~\citep{anp2024,agentconnectptl,agentcommunicationptl,yang2025survey}, such as the MCP~\citep{anthropic2025}. Unlike conventional APIs that rely on stateless, transactional exchanges, MCP supports persistent, semantically coherent dialogues between agents and services. It introduces three foundational capabilities: (1) \textit{dynamic capability discovery}, allowing agents to identify available system functionalities at runtime; (2) \textit{semantic context preservation}, which maintains task continuity across multi-step workflows; and (3) \textit{privacy-aware collaboration}, enabling rich information exchange while protecting sensitive data. Together, these features mark a shift from procedural invocation toward adaptive, negotiated interactions.

Beyond communication, the Interaction Dimension underpins \textit{tool orchestration}, the agent’s ability to safely compose and sequence external capabilities. Agents can dynamically verify tool properties, authenticate requests, and execute operations within controlled environments, minimizing risks such as malicious injection or execution failures.

Furthermore, this dimension facilitates \textit{agent-to-agent coordination}. Protocols like Agent-to-Agent (A2A)~\citep{a2a2025} enable agents to form ad hoc coalitions, share intermediate outputs, and collaboratively pursue complex goals. Such cooperative frameworks transform the Agentic Web into a networked fabric of interacting intelligences, where distributed reasoning and shared context allow agents to operate not just individually, but collectively.

\begin{figure}[t]
    \centering
    \includegraphics[width=1\linewidth]{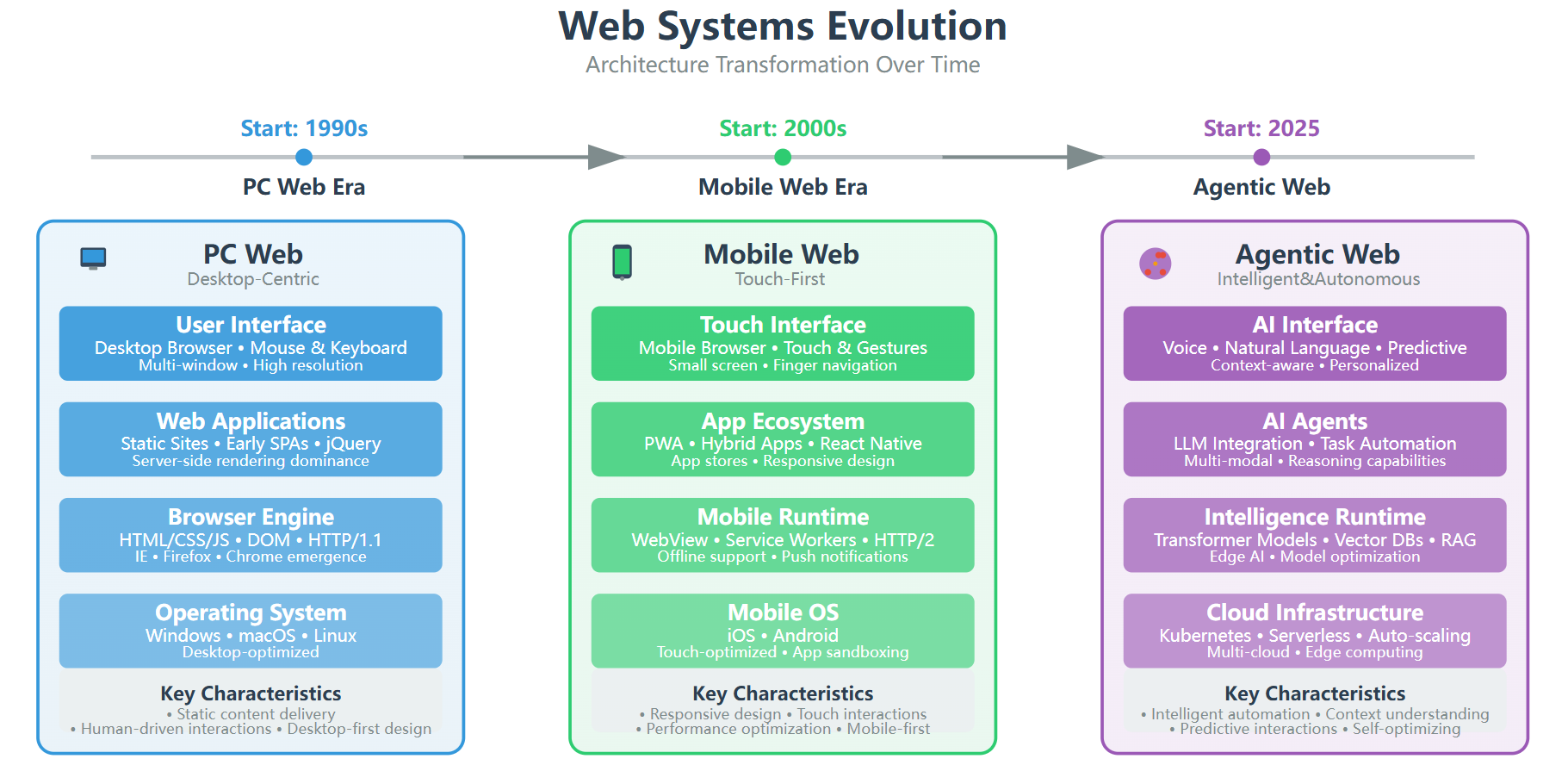}
    \caption{Architectural evolution of Web System. The transition reflects a shift from static content delivery and manual interaction to intelligent automation and outcome-oriented design. Not shown explicitly are the corresponding shifts in user roles, from navigator to operator to director, and in interaction models, from point and click to touch based to conversational delegation. These shifts mark a fundamental change in how value is created and tasks are fulfilled on the Web.}
    \label{fig:architectural-evolution}   
    \vspace{-0.2cm}
\end{figure}

By integrating semantic interoperability, safe tool access, and inter-agent collaboration, the Interaction Dimension forms the operational substrate of the Agentic Web. It allows autonomous agents to engage with a dynamic, heterogeneous environment in an adaptive and meaningful way as shown in Figure~\ref{fig:architectural-evolution}.

\subsubsection{Economic Dimension}
The Economic Dimension reconfigures digital ecosystems by introducing autonomous agents as economic actors capable of initiating transactions, forming collaborations, and allocating resources without direct human input. Unlike traditional systems, where value is created and exchanged through human interactions mediated by platforms, the Agentic Web supports machine-native economies in which agents coordinate, produce, and transact directly with one another~\citep{Tan2025, rothschild2025agenticeconomy,dawid2025agenticworkflowseconomicresearch}.

This shift gives rise to novel economic patterns. Agents can generate structured outputs such as executable workflows, tool manifests, and domain-specific datasets, not for human consumption, but for use by other agents. These machine-oriented artifacts enable closed-loop systems of generation and consumption, where agents operate continuously and collaboratively, driving a self-sustaining cycle of autonomous value creation.

Over time, such interactions form decentralized economic networks where agents dynamically discover services, negotiate terms, manage risk, and optimize outcomes through algorithmic reasoning. These agent-driven markets operate at speeds and granularities beyond human coordination, unlocking new forms of efficiency and scalability~\citep{yang2025agentexchangeshapingfuture}.

However, this transformation also introduces governance challenges~\citep{kolt2025governingaiagents,yang2025principlesaiagenteconomics}. As agents begin to make high-stakes economic decisions, potentially involving finances, contracts, and digital assets, questions around liability, transparency, and ethical alignment become urgent. Regulatory frameworks must evolve to accommodate autonomous behavior at machine timescales, ensuring accountability and fairness in increasingly complex agent-driven economies.

Ultimately, the Economic Dimension captures how agency, computation, and value creation converge in the Agentic Web, enabling a new kind of digital economy: one that is fast, adaptive, and fundamentally machine-mediated.


\section{Algorithmic Transitions for the Agentic Web}\label{sec:techniques}

\begin{figure}[t]
\centering
\includegraphics[width=0.9\textwidth]{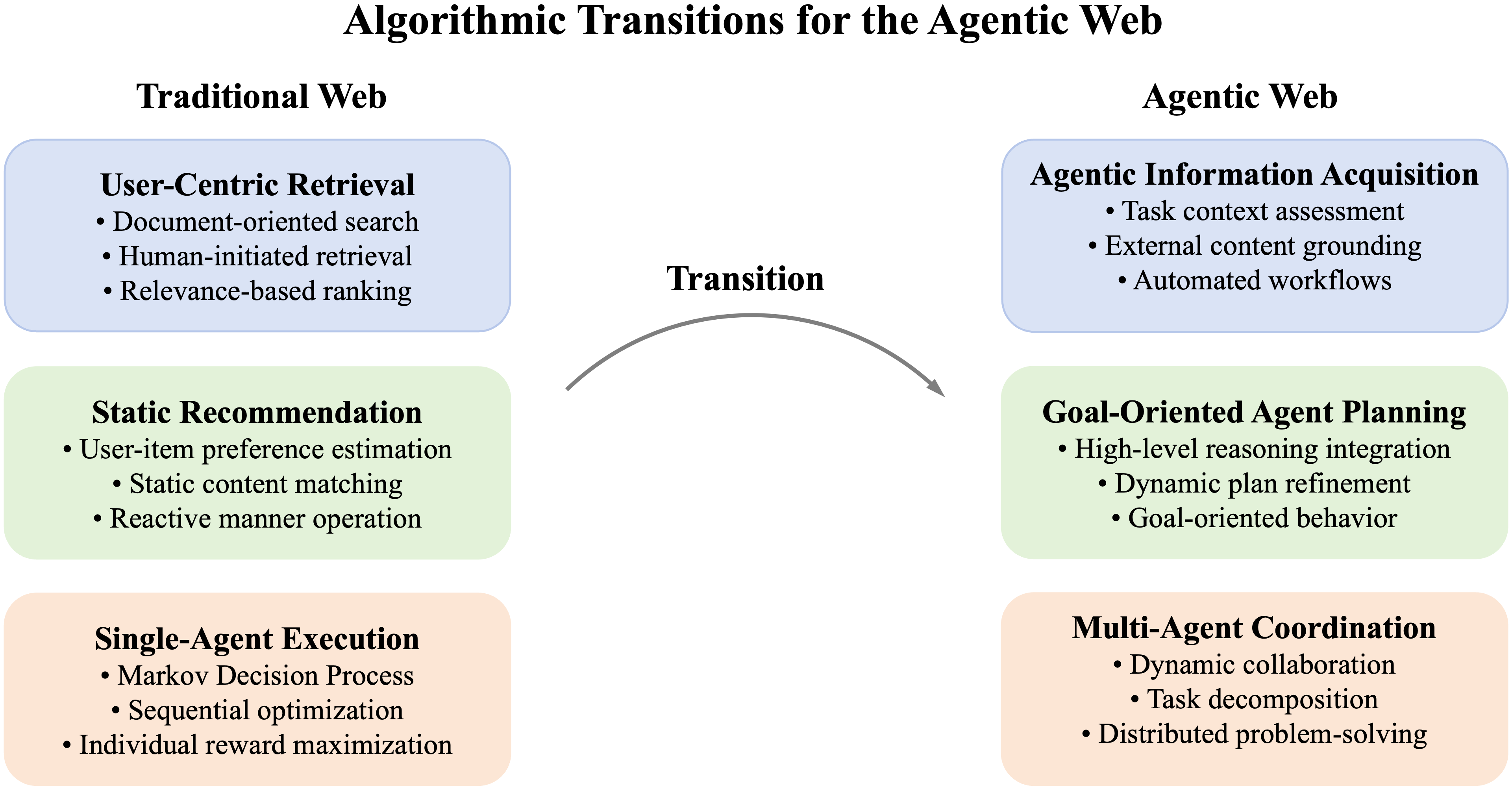}
\caption{Algorithmic Transitions for the Agentic Web. The figure illustrates three foundational transitions from Traditional Web to Agentic Web paradigms.}
\label{fig:algo_transitions}
\end{figure}

The emergence of the Agentic Web necessitates a fundamental re-evaluation and redefinition of the algorithmic underpinnings of intelligent systems. This paradigmatic shift represents more than mere technological advancement; it embodies a conceptual transformation from passive, human-driven computational processes to autonomous, goal-oriented intelligent behaviors that can operate independently within complex digital ecosystems. This section examines how traditional paradigms in information retrieval, recommendation systems, and agent architectures are converging and transforming to form the core capabilities of autonomous agents operating within dynamic web environments. We delineate three foundational transitions that characterize this algorithmic evolution, as illustrated in Figure~\ref{fig:algo_transitions}: (1) from user-centric information retrieval to proactive agentic information acquisition, where systems transition from reactive document lookup to autonomous, context-aware data gathering; (2) from static, one-shot recommendation to dynamic, goal-oriented agent planning, representing a shift from isolated preference predictions to integrated reasoning-action frameworks; and (3) from isolated single-agent execution to complex multi-agent coordination, enabling distributed intelligence and collaborative problem-solving capabilities. Each transition signifies a fundamental evolution from fixed, domain-specific pipelines that require explicit human intervention to adaptive, context-aware strategies that can autonomously navigate uncertainty and complexity. These transformations collectively establish the algorithmic foundation for intelligent systems capable of independent operation, continuous learning, and emergent behavior. Such characteristics distinguish the Agentic Web from its predecessors, with implications extending beyond technical improvements to encompass new possibilities for human-AI collaboration, automated decision-making, and the emergence of truly autonomous digital agents.

\subsection{User-centric Retrieval to Agentic Information Retrieval}

Traditional information retrieval has historically been grounded in human-initiated, document-centric search methodologies. Foundational techniques such as the bag-of-words model and term frequency-inverse document frequency \citep{sparck1972statistical} assign weights to terms based on their frequency within a document and their rarity across the corpus, thus producing a basic relevance signal for ad hoc queries. Probabilistic models such as Okapi BM25 \citep{robertson1995okapi} introduced refinements by incorporating term-frequency saturation and document-length normalization, enabling more effective handling of heterogeneous document sizes and term distributions. With the advent of the web, link-analysis algorithms such as PageRank \citep{page1999pagerank} contributed by modeling a random surfer traversing the hyperlink graph, introducing authority as a critical factor in ranking. More recent advances have leveraged supervised Learning to Rank approaches, which apply pointwise, pairwise, or listwise machine learning objectives to optimize ranking functions using labeled relevance data, significantly improving performance over traditional unsupervised methods \citep{liu2009learning}.

In contrast, the Agentic Web redefines retrieval as an active and integral component of autonomous cognition. Rather than performing static keyword-based queries, autonomous agents dynamically assess their goals, environment, and task progression to determine what information is needed, when to acquire it, and through which modalities \citep{zhang2024agentic}. These agents engage in complex, multi-step retrieval pipelines that often involve domain-specific tools, external APIs, and procedural logic, enabling them to construct knowledge on demand. This transition from passive search to proactive information acquisition supports end-to-end workflows that require minimal human supervision, enhancing both task generalization and responsiveness.

RAG architectures exemplify this shift by grounding language model outputs in external, verifiable content \citep{lewis2020retrieval}. Within this framework, Fusion-in-Decoder retrieves relevant passages via sparse or dense indexes and integrates them with the input query in a sequence-to-sequence model such as T5 \citep{izacard-grave-2021-leveraging}. FLARE adopts an iterative retrieval strategy in which the model forecasts the next sentence, uses low-confidence predictions to formulate pseudo-queries, and refines the output through successive retrieval rounds \citep{jiang2023active}. SELF-RAG introduces a self-reflective loop that prompts the model to critique and revise its own responses, thereby enhancing factual accuracy \citep{asai2024selfrag}. RetrievalQA explores adaptive retrieval policies that allow models to determine whether and when to retrieve based on internal uncertainty estimates \citep{zhang-etal-2024-retrievalqa}. The Tree of Clarifications framework \citep{kim-etal-2023-tree} addresses query ambiguity by decomposing questions into clarifying subqueries, retrieving evidence for each, and synthesizing comprehensive answers. Toolformer extends these capabilities by enabling models to autonomously identify suitable APIs, invoke them at appropriate stages, and incorporate the outputs into subsequent token generation \citep{schick2023toolformer}. Collectively, these innovations demonstrate how deeply integrated retrieval mechanisms enhance agent reasoning, supporting sophisticated tasks such as information synthesis, procedural decision-making, and tool utilization, thereby establishing the Agentic Web as a foundation for scalable and intelligent autonomous interaction.

\subsection{Recommendation to Agent Planning}

The traditional recommendation paradigm, which centers on matching users with items, is reinterpreted in the context of the Agentic Web as a strategic and goal driven framework for planning and execution. Earlier systems employed algorithms such as user based and item based collaborative filtering \citep{sarwar2001item}, matrix factorization methods \citep{koren2009matrix}, and deep learning based recommendation models \citep{he2017neuralcollaborativefiltering} to estimate user preferences over static content. While these techniques are effective at retrieving individually relevant items, they operate in a reactive manner: each recommendation is an isolated prediction that neglects downstream task dependencies and does not support multi step workflows. Modern advances have fundamentally transcended these limitations through the introduction of sophisticated architectural innovations that enable autonomous goal oriented behavior. Language Agent Tree Search exemplifies this evolution by integrating MCTS with LLM powered value functions and self reflection mechanisms \citep{zhou2023language}. 

In contrast, the Agentic Web redefines recommendation as a proactive process involving multi step planning by autonomous agents. This conceptual shift has led to the development of agents powered by large language models that integrate high level reasoning with executable actions in dynamic web environments. For example, ReAct \citep{yao2023react} integrates reasoning traces with concrete actions, allowing agents to refine their plans based on environmental feedback and to achieve significant improvements on question answering and interactive decision-making benchmarks. WebAgent \citep{gur2024a} converts natural language instructions into Python programs while summarizing lengthy HTML content into task-specific segments, thereby enabling agents to interact with real web interfaces through programmatic planning. AdaPlanner \citep{sun2023adaplanner} introduces a closed loop planning architecture that incorporates both in plan and out of plan refinements, dynamically updating plans based on environmental feedback to outperform standard baselines on ALFWorld \citep{shridhar2020alfworld} and MiniWoB++ \citep{liu2018reinforcement}. Plan-and-Act \citep{erdogan2025plan} further separates planning and execution into two distinct roles, a planner and an executor, each instantiated by a large language model, and achieves state of the art performance on long horizon web navigation tasks. Beyond these foundational approaches, recent developments have introduced sophisticated memory augmented systems such as the Task Memory Engine, which implements spatial memory using Directed Acyclic Graphs to replace linear context concatenation \citep{ye2025task}. GoalAct demonstrates continuous global planning that maintains clear objectives through skill based decomposition and hierarchical execution strategies, achieving improvement in success rates on complex legal benchmarks \citep{chen2025enhancing}.

This evolution represents a fundamental transition from passive recommendation to prescriptive and goal oriented behavior, empowering agents to autonomously interpret instructions, navigate web environments, and manipulate digital interfaces in pursuit of complex user defined objectives. The emergence of standardized evaluation frameworks such as WebArena \citep{zhou2023webarena}, VisualWebArena \citep{koh2024visualwebarena}, and ST-WebAgentBench \citep{levy2024st} has established comprehensive protocols for assessing multi dimensional agent capabilities across planning, tool integration, safety, and trustworthiness. Contemporary agents demonstrate substantial performance gains through enhanced autonomous task completion capabilities, while standardized protocols including the Model Context Protocol and Agent Communication Protocol enable seamless integration across heterogeneous agent systems \citep{anthropic2025multiagent, li2024survey}.

\subsection{Single‑Agent to Multi‑Agent Coordination
}

Traditional single-agent systems have typically modeled autonomous decision-making in web environments using the Markov Decision Process framework. Early work by Shani et al. demonstrated that formulating recommendation tasks as sequential optimization problems outperformed static approaches in maximizing long-term user satisfaction \citep{shani2005mdp}. Building on this foundation, contextual bandit algorithms such as LinUCB were developed to adaptively select content by incorporating user and contextual information, improving cumulative engagement through iterative learning \citep{li2010contextual}. To address the bias and sparsity inherent in logged interaction data, off-policy actor-critic methods with top-K corrections have been successfully scaled to large recommender systems, enhancing stability and effectiveness in environments with millions of candidate actions \citep{chen2019top}. Additionally, slate-based reinforcement learning (RL) decomposes multi-item recommendation problems into tractable value functions, enabling efficient Q-learning over complex combinatorial action spaces \citep{ie2019slateq}. Although these single-agent approaches are effective for optimizing individual objectives, they face limitations in scenarios that require collaboration, distributed reasoning, or task-level adaptability.

To address these limitations, multi-agent coordination frameworks have emerged to enable dynamic collaboration among agents, allowing them to collectively solve tasks that are difficult to address in isolation. This paradigm shift supports task decomposition, role specialization, and orchestrated execution through communication and shared goals. AutoGen exemplifies this transition by implementing planner, executor, and critic roles, using prompt conditioning to assign responsibilities and structure inter-agent interaction \citep{wu2023autogen}. AgentOccam enhances LLM-based agents not by refining agent strategies alone, but by improving their fundamental reasoning and task comprehension capabilities \citep{yang2025agentoccam}. WebPilot incorporates a multi-agent MCTS architecture to guide web navigation and decision-making, demonstrating the potential of hierarchical planning in interactive environments \citep{zhang2025webpilot}. The AI Co-Scientist leverages multiple agents and external tools to formulate novel scientific hypotheses, combining web search with specialized AI modules to generate well-grounded research proposals \citep{gottweis2025towards}.

Recent systems have also emphasized flexibility, modularity, and accessibility in multi-agent design. Alita introduces a minimalist architecture that reduces predefined roles and promotes self-evolution, aiming for greater scalability and generalization across domains \citep{qiu2025alita}. OWL offers a structured agent hierarchy, decomposing tasks into specialized roles filled by UserAgents, AssistantAgents, and ToolAgents to automate complex real-world objectives \citep{hu2025owl}. AutoAgent enables users to construct multi-agent workflows and integrate external tools without extensive technical knowledge, expanding the accessibility of agent-based system design \citep{tang2025autoagent}. Octotools organizes execution into distinct planner and executor components, optimizing the coordination of multi-tool computational workflows \citep{lu2025octotools}. These developments collectively reflect a broader shift toward distributed intelligence, demonstrating the increasing importance of collaboration, modularity, and specialization in next-generation autonomous systems. \\\\

\vspace{-0.2cm}
\section{Systematic Transitions of the Agentic Web}\label{sec:system}

The shift toward an Agentic Web entails not only a conceptual evolution but also a fundamental redesign of the underlying system architecture. Traditional web infrastructure, based on stateless protocols, user-initiated interfaces, and static interaction models, is poorly suited to the requirements of agentic computation. To function effectively, autonomous agents necessitate continuous contextual awareness, persistent sessions, dynamic service discovery, semantically rich interaction protocols, and real-time coordination with both human users and other agents.

This section articulates the system-level transformations necessary to support agent-native execution. It identifies the limitations of current web protocols and runtime environments, proposes infrastructure requirements for persistent, context-aware agents, and discusses the evolution of communication standards that enable semantic, agent-to-agent interactions. By analyzing these transitions systematically, this section contributes a coherent architectural perspective for operationalizing the Agentic Web beyond isolated prototypes or platform-specific implementations.

We begin by outlining the core system challenges that arise when deploying autonomous agents in web-scale environments.

\subsection{Motivation for an Agentic Web System}
\label{sec:5.1}

The advent of the Agentic Web, a paradigm characterized by autonomous, intelligent agents executing complex user-delegated tasks, represents a fundamental architectural evolution from the content-centric model of the contemporary internet. This transition necessitates a profound re-engineering of underlying network and system architectures, as the internet's extant infrastructure is ill-equipped to support the dynamic, decentralized, and mission-critical nature of agentic operations. For this emergent ecosystem to become viable, several core system-level challenges must be surmounted, transforming the internet from a passive data conduit into an intelligent, proactive, and service-aware fabric.

A primary impediment to the realization of the Agentic Web is the challenge of agent discovery. In contrast to the static addressing schemes of the traditional internet, where resources are located via stable IP addresses or domain names, autonomous agents are ephemeral and lack a fixed, identifiable network location. When a principal agent must execute a task exceeding its intrinsic capabilities, it must dynamically recruit other agents. This scenario presents a critical problem: the identification and selection of suitable collaborators from a vast, decentralized population of agents. The resolution of this problem requires a dynamic discovery mechanism, analogous in principle but superior in intelligence to network routing protocols \citep{cui2025agentdnsrootdomainnaming}. For a given task, the system must effectively source and select agents by conducting a comprehensive assessment of their skills, readiness, and suitability to the specific operational demands. This "just-in-time" matchmaking is a prerequisite for the seamless, on-demand collaboration essential for executing complex, multi-agent operations \citep{chen2024internetagentsweavingweb,raskar2025upgradeswitchneednextgen}.

Subsequent to discovery, the challenge of effective inter-agent communication arises, with the current ecosystem of APIs presenting a significant roadblock. Contemporary APIs are predominantly engineered for consumption by human developers, achieving syntactic but not semantic interoperability. While they rigorously define the structure for data exchange, they lack the formal semantic annotations necessary for an autonomous agent to unambiguously interpret their function and purpose \citep{tupe2025aiagenticworkflowsenterprise}. An agent may parse the technical format of a request but cannot ascertain the underlying intent or operational semantics of the API's methods. To transcend this limitation, a new paradigm of agent-oriented APIs is required. A future standard would likely necessitate the integration of machine-readable formalisms, such as ontologies or logical specifications, directly within the API definition. This would create a unified system wherein an agent can access not only the syntactic structure but also the semantic context, thereby empowering agents to autonomously discover, comprehend, and utilize APIs to perform complex tasks without human intervention \citep{Braubach2018}.

Furthermore, the highly dynamic and distributed nature of this network introduces complex logistical challenges related to billing and accounting. In a system where agents can spontaneously collaborate, delegate sub-tasks, and consume services from one another, tracking resource utilization for the purpose of accurate attribution and billing becomes exceptionally difficult. A persistent, reliable, and auditable methodology is needed to monitor the chain of interactions and associate computational and service costs with the originating user or principal agent. This necessitates the design of a transactional framework capable of tracing an agent's activities and resource consumption across a fluid, multi-party network. Such a framework must securely bind billing information to a specific entity, ensuring that as agents access premium services or delegate paid tasks, the associated costs are accurately calculated and charged \citep{cui2025agentdnsrootdomainnaming}. Without a robust solution for micropayments and distributed accounting, the economic models required to sustain a sophisticated, service-driven Agentic Web are untenable.

Finally, a foundational challenge resides in the evolution of the network infrastructure itself: specifically, the transition from a best-effort network to an intelligent infrastructure capable of delivering guaranteed, personalized quality of service. The current network architecture, exemplified by 5G, is fundamentally network-centric, engineered to optimize a limited set of aggregated KPIs, such as peak data rate and latency. This model, while effective for enhancing general system capabilities, is insufficient for the Agentic Web, which demands a paradigm shift from optimizing universal metrics to providing bespoke, task-specific service guarantees. The infrastructure must evolve to comprehend and dynamically accommodate the distinct, multi-dimensional requirements of individual agentic tasks. This transition reflects a shift from providing generalized high-performance capabilities to acting as an intelligent orchestrator that interprets specific task requirements across multiple dimensions, including cost, security, and access to knowledge, and provisions tailored Service Level Agreements accordingly.

\begin{figure}[!t]
    \centering
    \includegraphics[width=1\linewidth]{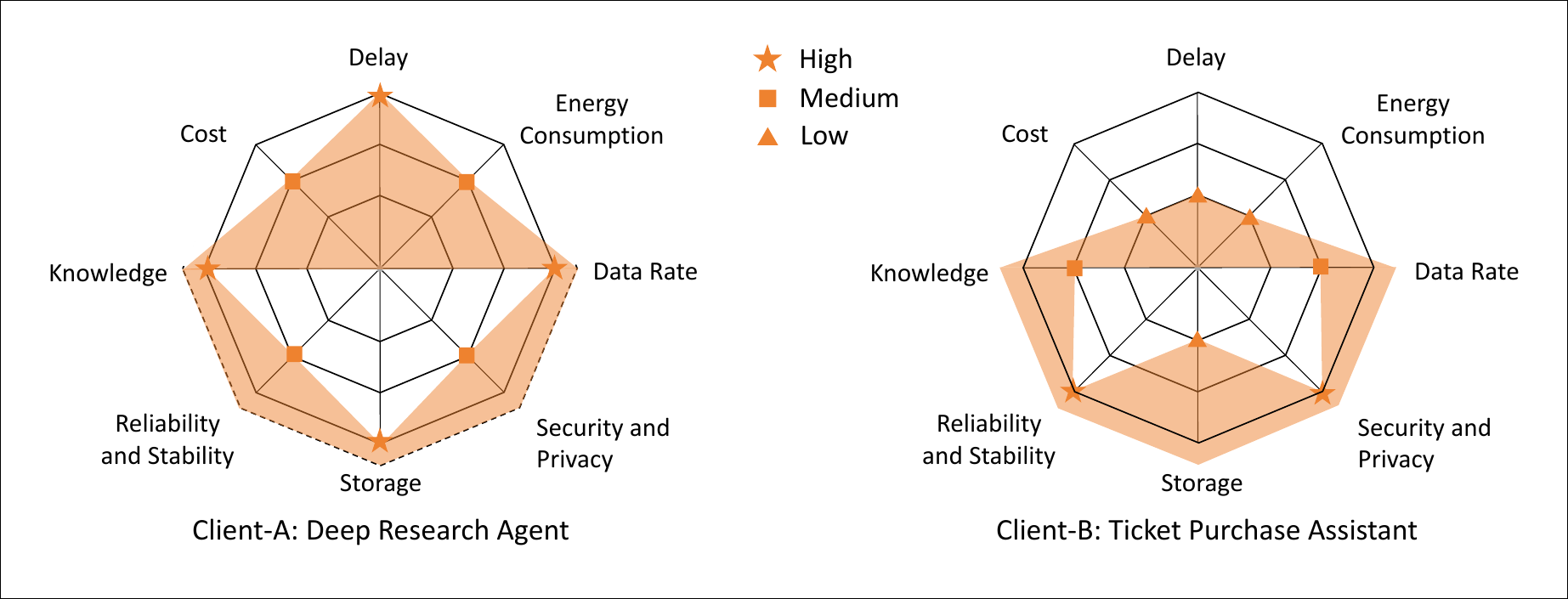}
    \caption{Service Requirement Zone.}
    \label{fig5.1_1}
    \vspace{-0.2cm}
\end{figure}

Figure \ref{fig5.1_1} illustrates a task's Service Requirement Zone (SRZ) \citep{9839652}, an eight-dimensional profile defining its quality of experience needs across metrics like cost, delay, security, data rate, and knowledge. The size of the shaded SRZ on the chart indicates the stringency of these needs: a smaller zone means more exacting demands, requiring precise resource orchestration. It also contrasts two different agentic tasks to demonstrate this concept. The Deep Research Agent (left) displays a constricted SRZ, reflecting its complex requirements. It has a high demand for Knowledge to access specialized models, strong Reliability and Stability, and low Delay to enable interactive analysis. This profile necessitates a highly capable and responsive service.

In contrast, the Ticket Purchase Assistant (right) has a larger, more flexible SRZ. This agent’s primary needs are high Security for processing payments and a high Data Rate to quickly load options. It can tolerate longer delays and has minimal requirements for energy or local storage, which is typical for a transactional task. This comparison highlights how different agents have unique service profiles that the underlying infrastructure must be able to interpret and fulfill.

To adequately support the Agentic Web, the underlying infrastructure must therefore evolve to natively interpret and fulfill these diverse SRZs. It can no longer treat the VR stream and the banking transaction as fungible data flows. The system must transition from network-level slicing to a far more granular mode of service provisioning at the individual task level \citep{liu20256gintense, chen2024intent}. To achieve this, the network requires pervasive, embedded intelligence, allowing it to efficiently identify a task's SRZ and subsequently orchestrate heterogeneous network, compute, and data resources across multiple domains to guarantee that its specific quality of experience is met, thus marking a definitive departure from the legacy ``best-effort'' paradigm \citep{mahmood2024digital, wang2024drl}. 

The paradigm shift towards SRZ-centric service delivery imposes a set of unprecedented demands on the underlying system architecture that far surpass the capabilities of the traditional web. The system must support extreme dynamism and negotiation, as an agent's resource needs can change dramatically mid-task, requiring real-time allocation adjustments \citep{huang2024bandwidth, zhang2025usercentric}. It must natively handle multimodality, intelligently managing the varied requirements of text, image, audio, and other data types. The system's role must also evolve from a mere data transporter to a capability-driven orchestrator, maintaining a real-time inventory of computational resources, AI models, and data sources to fulfill agent requests \citep{li2025securesemantic}. Furthermore, it must provide granular, verifiable security and privacy controls at the level of individual agents and sub-tasks, offer deep observability for robust debugging and optimization, and incorporate intelligent cost control mechanisms to manage computationally expensive agentic workflows \citep{li2025securesemantic}.

\subsection{Toward a Next-Generation Agentic Web System}

To facilitate the large-scale deployment of autonomous agents, the Web must evolve from a content-centric medium to an execution-oriented infrastructure. This paradigm shift necessitates a fundamental re-evaluation of web systems' architectural foundations to support agent-native interaction patterns, persistent context management, and integrated tool orchestration.

In this subsection, we present the Agentic Web system, which integrates three essential elements: the User Client, the Intelligent Agent, and Backend Services. We elucidate the functional roles of each component, examine their historical evolution, and analyze their collective function in translating high-level user goals into executable digital actions. By articulating this architecture, we establish a conceptual framework for understanding how agentic capabilities can be operationalized, thereby bridging the divide between user intent and dynamic service execution.

\subsubsection{Roadmap of the Agentic Web System}\label{sec:5.2.1}
This section delineates the architecture of the Agentic Web, a tripartite structure designed to translate user objectives into executable operations. This architecture is composed of three integral components that operate in synergy: the User Client, the Intelligent Agent, and Backend Services.

The User Client serves as the primary medium for human-agent interaction. Its core functions are to process user inputs like textual, vocal and to render the agent's synthesized outputs. The historical trajectory of clients shows an evolution from text-based command-line interfaces to the intuitive graphical and touch-based paradigms of today. The contemporary trend is a progression towards multimodal systems that integrate diverse inputs such as voice and gesture, embodied in devices like smart speakers and wearable technology.

The Intelligent Agent constitutes the system's central cognitive and decision-making nexus. Leveraging artificial intelligence disciplines such as Natural Language Processing, the agent discerns user intent, decomposes complex objectives into granular sub-tasks, and selects appropriate backend tools for execution. The developmental path of these agents has advanced from rudimentary rule-based systems to sophisticated learning models. These modern agents are capable of addressing complex creative and epistemic tasks by continuously adapting based on new data and user feedback.

Backend Services, Tools, and Plugins form the functional substrate, providing the essential computational, data, and specialized capabilities required by the agent. These modular resources encompass a wide spectrum of functions, from general utilities like language translation to domain-specific industry applications. Architecturally, they have evolved from monolithic databases to a distributed and extensible ecosystem of microservices and plugins, which permits third-party developers to continuously augment the capabilities of the Agentic Web.

\begin{figure}[t]
    \centering
    \includegraphics[width=1.0\linewidth]{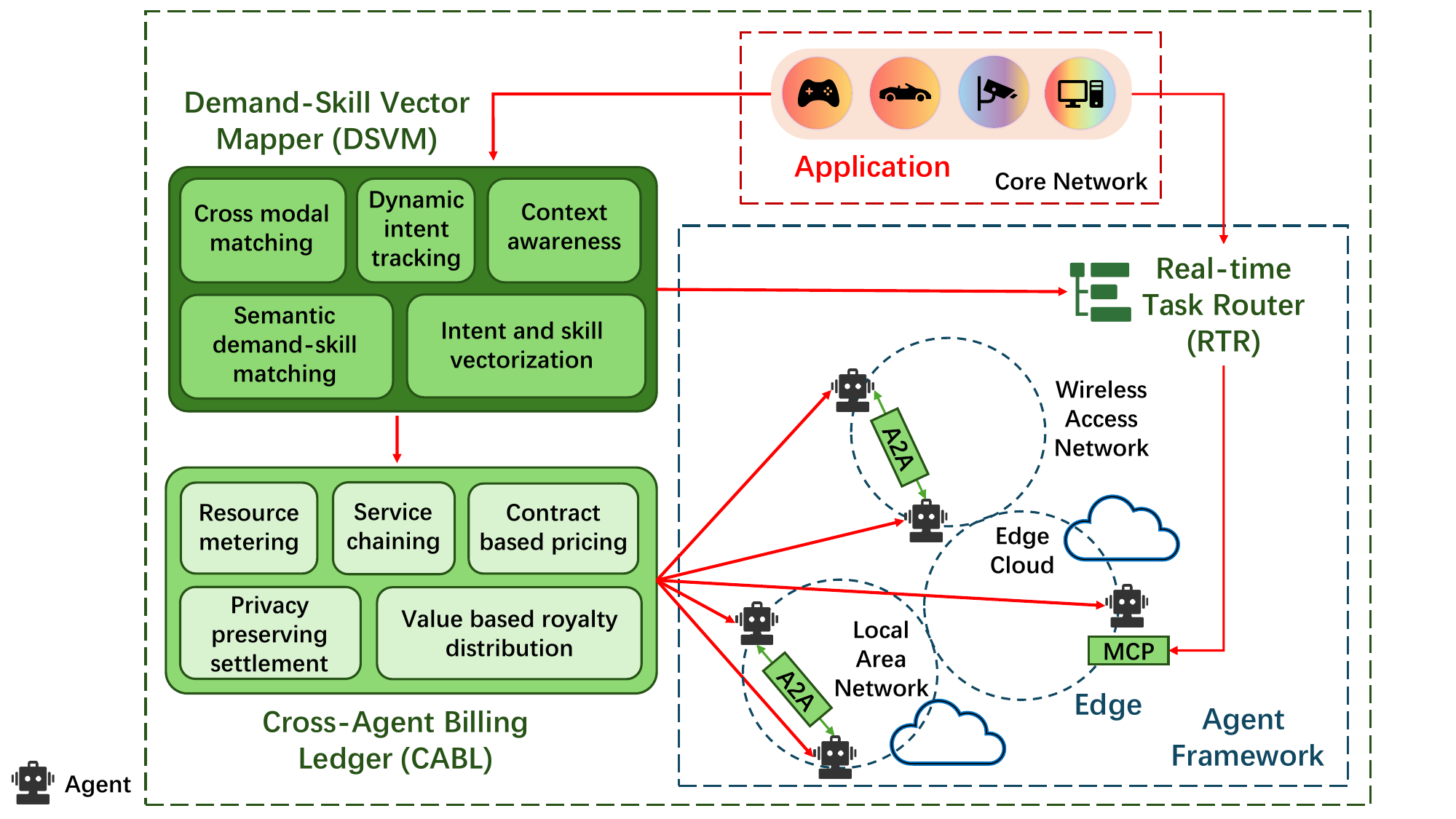}
    \caption{Agentic Web Roadmap}
    \label{fig5.1_2}
    \vspace{-0.2cm}
\end{figure}

To illustrate the interoperability of these components, consider a representative interaction workflow. The process is initiated when the User Client transmits a high-level objective, such as “plan a business trip to Shanghai,” to the Agent. The Agent then decomposes this objective into constituent sub-tasks. It subsequently identifies and invokes the requisite external services, such as flight and hotel booking platforms, to execute these tasks. Upon receiving the necessary information, the Agent integrates these disparate results. It synthesizes the data, evaluates it against predefined user constraints, and formulates a coherent, consolidated response—such as a complete itinerary—which is then relayed to the User Client for presentation.

Furthermore, the architecture accommodates a direct interaction model. In certain scenarios, the Agent may orchestrate an initial connection, after which the User Client engages directly with a backend service. This decoupled model is particularly advantageous for transactions involving sensitive data or requiring high-throughput, such as financial payments. This design allows the Agent to preserve its function as the master coordinator while delegating specific interactions to optimize for security and efficiency.

In summary, this architecture represents a paradigm shift from direct user manipulation of discrete applications to a model of delegated goal-fulfillment. Within this paradigm, a user entrusts a high-level objective to an intelligent, autonomous Agent, which then orchestrates a diverse set of resources to achieve the specified outcome. The User Client functions as the dedicated human-computer interface; the Agent operates as the central cognitive processor and orchestrator; and the Backend Services constitute an ecosystem of callable functionalities (e.g., APIs, databases, web applications) capable of executing specific, well-defined tasks.

These rigorous requirements fundamentally invalidate the traditional Client-Server architecture, mandating a shift toward the Client-Agent-Server model. We propose an Agentic Web Architecture, depicted in Figure\ref{fig5.1_2}, to realize this paradigm. This architecture operates through three core components. First, a demand skill vector mapper interprets application needs by performing context awareness, dynamic intent tracking, and semantic vectorization to translate service demands into machine-readable formats. Second, a real time task Router dynamically dispatches these vectorized tasks to a distributed Agent Framework operating across the edge and access networks. Third, a cross agent billing ledger governs the economic and resource interactions between agents, enabling crucial functions like resource metering, service chaining, and privacy-preserving settlement. This integrated design creates an intelligent, autonomous, and value-aware fabric for orchestrating complex agentic services.

\subsubsection{Interaction Process Example: Collaborative Mechanisms in Travel Itinerary Planning by Agents}

\begin{figure}[t!]
    \centering
    \includegraphics[width=0.9\textwidth]{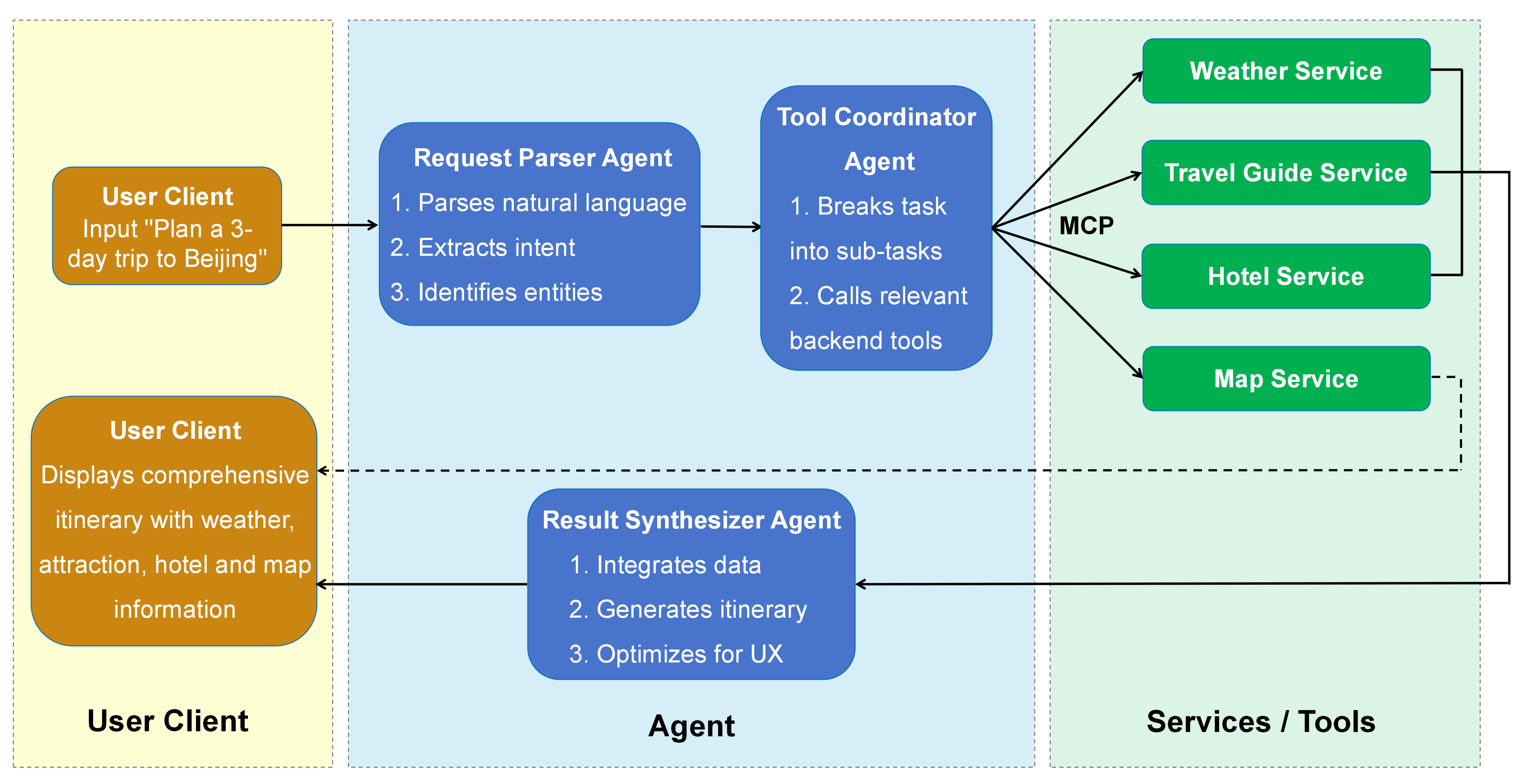}
    \caption{Interaction Process Example: Planning a Travel Itinerary}
    \label{fig:5.2.2}
    \vspace{-0.2cm}
\end{figure}

As depicted in Figure \ref{fig:5.2.2}, the workflow is initiated upon the submission of the request, “Plan a 3-day trip to Beijing,” through the User Client. A Request Parser agent semantically parses this input to extract key parameters, including the destination, duration, and the core user objective.

Subsequently, a yool orchestrator agent decomposes the primary objective into four discrete sub-tasks: obtaining meteorological data, compiling information on tourist attractions, querying for accommodation options, and generating a route map. It then programmatically invokes the corresponding backend services via the MCP protocol:
\begin{itemize}
\item Request forecast data from the Weather Service.
\item Request attraction information from the Travel Guide Service.
\item Request accommodation options from the Hotel Service.
\item Request the generation of optimized travel routes from the Map Service.
\end{itemize}

Following data retrieval, a result synthesizer agent aggregates and integrates the information from the weather, travel guide, and hotel services to construct a comprehensive itinerary. Concurrently, the response from the Map Service bypasses the synthesis stage, transmitting route maps and location data directly to the User Client. This direct feedback channel is explicitly marked in Figure \ref{fig:5.2.2} (dashed line).

Finally, a consolidated itinerary, comprising weather forecasts, attraction details, accommodation suggestions, and an interactive map, is then transmitted to the User Client for review. The user can accept the proposal or request modifications, with map data being updated dynamically via the direct service feedback mechanism. This dual-pathway design strategically balances comprehensive data integration for high-level planning with low-latency service responses for interactive components, all orchestrated through a unified platform.

\subsubsection{Recent Advances and Applications of Agentic Web Systems}

To advance the agentic web, recent research has focused on overcoming core challenges in agent design and assessment. Key innovations include synergizing reasoning with action, architecturally separating planning from execution, and establishing more reliable benchmarks to accurately measure performance.

ReAct, a framework developed by \citep{yao2023react} achieves synergy by interleaving reasoning and acting within large language models. The proposed methodology enhances the model's performance in two key ways. First, it enables the generation of reasoning traces to formulate and dynamically adjust action plans. Second, it facilitates interaction with external resources, exemplified by a Wikipedia API, to retrieve information, which is essential for fact-checking and minimizing hallucinations. This dynamic combination of ``acting to reason'' and ``reasoning to act'' enables agents to more reliably solve knowledge-intensive tasks and enhances the interpretability of their decision-making processes.

To tackle challenges in long-horizon tasks, the PLAN-AND-ACT framework \citep{erdogan2025plan} distinctly segregates strategic high-level planning from immediate low-level actions. This architecture features a PLANNER model dedicated to generating structured, abstract strategies and an EXECUTOR model responsible for translating these strategies into tangible steps in the environment. A key innovation of this framework is dynamic replanning, which addresses the limitations of static plans. The PLANNER updates the plan after each action is executed, enabling the agent to acclimate to evolving environmental conditions and incorporate new information, such as search results, into the ongoing strategy.

For a more accurate measurement of true web agent capabilities, existing benchmarks like WebVoyager have been identified as a key limitation, as they often suffer from a lack of task diversity and can report inflated performance results \citep{xue2025illusionprogressassessingcurrent,deng2023mind2web}. To address this, the new Online-Mind2Web benchmark offers a comprehensive evaluation suite, containing 300 diverse and realistic tasks that cover a broad spectrum of 136 websites. Concurrently, an automated evaluation method called WebJudge was also developed. This approach identifies key points for task completion and then selects key screenshots from an agent's trajectory for evaluation, preserving critical information while avoiding context length limits. This method achieves up to 85.7\% agreement with human judgment and significantly improves evaluation reliability and scalability.

\subsection{Agentic Communication}
The deployment of autonomous agents in complex web environments for multi step tasks introduces novel communication demands that fundamentally exceed the capabilities of traditional web protocols. As agents evolve from passive API consumers to proactive, context-aware actors capable of initiating and coordinating tasks, they require better communication mechanisms that support semantic interoperability, persistent task states, and asynchronous multi-party interaction, and many other features. 

This section investigates the protocol level foundations of the Agentic Web, examining the limitations of conventional protocols such as HTTP and RPC, and motivating the need for agent-native alternatives. We introduce two representative protocols, MCP and A2A, that exemplify emerging approaches to structured, scalable, and semantically rich communication among agents and services. The following subsections first analyze the design motivations behind these protocols and then provide detailed descriptions of their architectures and workflows.

\subsubsection{Design Motivation (Beyond HTTP/RPC)}
In the current Internet ecosystem, the Hypertext Transfer Protocol (HTTP) and Remote Procedure Call (RPC) have long served as the mainstream communication protocols, underpinning the data interaction between Web services. Over the past two years, numerous AI Agent projects have achieved basic communication functions based on these two protocols. However, with the rise of the Agentic Web concept, the limitations of traditional protocols have gradually become prominent. The Agentic Web, characterized by autonomy, context awareness, and dynamic interaction, has operational mechanisms that impose new requirements on communication protocols far beyond the capabilities of HTTP/RPC.

Firstly, the task execution process in the Agentic Web typically involves collaboration among multiple entities, which imposes stringent requirements on the \textbf{efficient management of task specific context}. In the Agentic Web, the task execution process frequently necessitates intricate interactions between designated agents and other agents, in addition to non-agent resources such as external tools, data, and services. During these interactions, all participating entities are required to maintain, transmit, and share specific context, such as historical data or environmental configuration parameters. For example, when a personal assistant agent is assigned the task of arranging a travel itinerary for a user, it may need to query a weather API and interact with hotel and transportation booking agents. Throughout this process, the involved entities must exchange both private and non-private user data, including user preferences and authorizations, and share progress information related to the booking task. However, traditional web protocols (HTTP/RPC) are principally designed for the transmission of data and lack semantic-level support. This design limitation results in the treatment of complex context as ordinary data, with no distinction among higher-level semantic elements such as \textit{historical context}, \textit{user intent}, or \textit{environmental configuration}. Furthermore, these web protocols have been demonstrated to lack the capacity to process logical semantics, such as preconditions. Consequently, traditional web protocols are inadequate in meeting the stringent demands of the Agentic Web for efficient context management during task execution.

Secondly, the task execution process in the Agentic Web is contingent on LLM based agents, thereby engendering heightened requirements for \textbf{semantic accuracy in communication and interaction}. In the Agentic Web, entities need to communicate through structured and standardized protocols to ensure semantic consistency and operational feasibility during task execution. However, LLM-based agents typically mediate their understanding and generation of structured content through natural language, introducing inherent non-determinism into the output process, which is susceptible to issues such as formatting deviations and semantic drift, compromising the accuracy and reliability of interactions. Consider the case of traditional API calls, which rely on developers to interpret interface semantics from documentation and manually construct deterministic, structured invocations. Conversely, in the Agentic Web, agents are required to automatically interpret interface semantics and translate natural language descriptions into operational commands. In the absence of a semantically specified mechanism, the generated results are frequently unstable, which complicates the assurance of structural integrity and semantic correctness in API calls. Therefore, Agentic Web protocols provide machine-readable interface semantic specifications that explicitly define the data types, value ranges, and business meanings of each field. This can guide LLMs in accurately parsing and generating structured invocations. Furthermore, the protocol must address semantic divergence across entities, for instance by unifying or mapping field labels such as ``UID'' and ``UserID'', in order to avoid semantic ambiguities. However, traditional web protocols such as HTTP and RPC are primarily designed for data transport and lack support for interface semantics and field alignment. Consequently, they are insufficient for the semantic coordination and contextual understanding required.

Finally, the task execution process in the Agentic Web requires \textbf{high interactivity} in communication. The task execution process is frequently distinguished by protracted durations, \textbf{multi-phase} workflows, and \textbf{asynchronous} operations. Furthermore, in circumstances involving sensitive actions, immediate user intervention may be imperative. This necessitates that agent communication be supported by persistent and dynamic interaction mechanisms. For instance, let us examine a case where a personal assistant agent is assigned the task of formulating a trading strategy. In such a case, the agent may need to communicate intermediate results to the participant at different phases of the procedure. In particular, if the task involves high-risk decisions or large financial transactions, the agent must halt its operation until it receives an explicit confirmation from the user to continue. The execution of such task workflows necessitates not only the possession of fine-grained task control capabilities by agents but also the implementation of communication protocols that support event-driven architectures, asynchronous responses, and persistent task state management. However, traditional web protocols such as HTTP and RPC typically adopt a synchronous request-response model and lack native support for long-running tasks. Furthermore, they are also ill-equipped to handle complex control flows such as task suspension, external event injection, or dynamic user confirmation. Whilst mechanisms such as polling and Webhooks can be utilized to facilitate partial asynchronous interactions, they frequently necessitate the implementation of additional logic layers, thereby increasing system complexity and compromising overall robustness. It is therefore vital that the Agent Web urgently requires more capable and interaction-rich protocol mechanisms to support multi-phase, multi-party task workflows in a more natural and efficient manner.

According to the research in \citep{yang2025survey}, a large number of new agent communication protocols emerged in the past year. Among them, the general-purpose protocols, MCP and A2A, have demonstrated significant technical advantages and community influence. These two protocols are designed from different dimensions to address the characteristics of the Agentic Web, forming complementary solutions. We will provide a brief introduction to these two communication protocols and explain which key challenges faced by the Agentic Web they can address and how to address them, followed by detailed introductions to their workflows.

MCP, short for Model Context Protocol, proposed by Anthropic \citep{MCPBlog}, is a communication protocol focused on interactions between agents and non-agent resources. It aims to establish standardized interfaces for tool invocation and has gradually evolved into a de facto industry standard. Under the framework of this protocol, applications encapsulate the tools, resources, and prompts they provide into service units that can be recognized by agents. Agents obtain application metadata, including function descriptions, input-output formats, and usage constraints, through the query interface of the MCP, and implement the call and control of applications based on the operation instructions defined by the protocol. For example, when an agent needs to call an image-generation tool, it can obtain the parameter-configuration specifications of the tool through the MCP and submit the generation task in a standardized request format to ensure the consistency of cross-platform tool calls. To some extent, MCP has enhanced the structural consistency and standardization of communication between agents and resources.

A2A, short for Agent-to-Agent \citep{A2ABlog}, proposed by Google, is a communication protocol specifically designed to facilitate direct interaction between agents through a distributed capability discovery and communication mechanism. Within A2A, each agent registers its capability-description file (AgentCard) to a predefined URI, publicly exposing its functions, interfaces, and communication specifications. Other agents can address and obtain the capability map and initiate asynchronous interactions supporting multimodal data. In addition to capability discovery, A2A also incorporates an authentication mechanism to establish secure communication channels between agents. This mechanism can integrate with \textbf{Decentralized Identifiers (DIDs)}, allowing each AgentCard to include a DID reference that links to a DID Document containing public keys and authentication methods. By resolving and verifying these DIDs, agents can perform decentralized, cryptographically verifiable identity checks without relying on centralized registries or third-party identity providers. This enables self-sovereign authentication, improves interoperability, and aligns well with the trust requirements of dynamic agent ecosystems. At the same time, to meet the requirements of user intervention in sensitive operations and asynchronous task control in the Agentic Web, the protocol introduces an event-driven and state-callback mechanism. It triggers event notifications at key nodes of long-cycle tasks, pushes intermediate results for user confirmation, and dynamically updates task states through the state-callback interface, compensating for the deficiencies of traditional protocols in asynchronous interaction and user intervention.

\begin{figure}[t]
\centering
\includegraphics[width=0.95\textwidth]{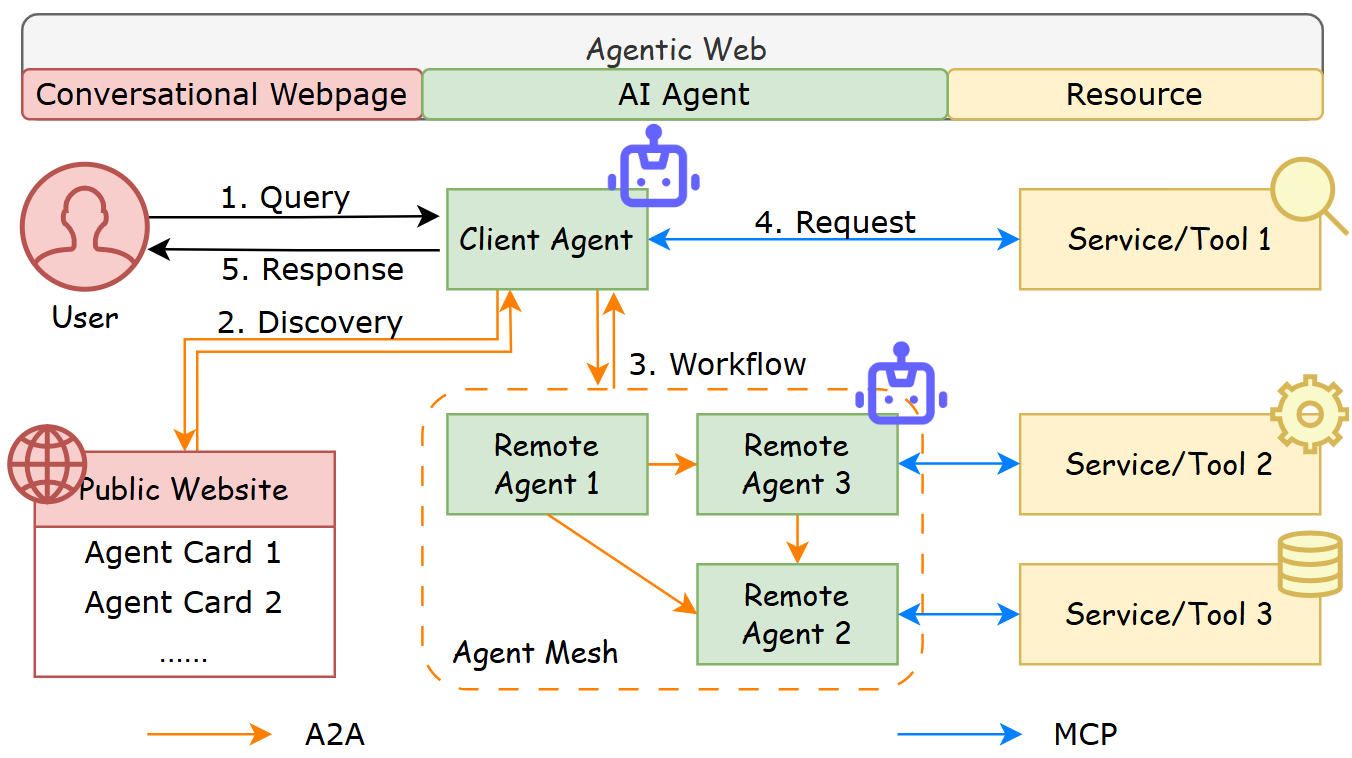}
\label{fig:AgenticWeb_AgentProtocol}
\caption{The above schematic illustrates a rudimentary AgentWeb system, with A2A and MCP serving as examples of agent communication protocols. In such an Agent Web, a user can assign a task (query) to a client agent via a conventional interactive medium (GUI or text). The client agent then discovers remote agents that meet the task's criteria on a public webpage, under the A2A protocol, and the task is subsequently allocated to their constituent agent mesh via the A2A protocol. During the execution process, agents may request external resources through the MCP protocol to facilitate task completion. Once the task completes, the client agent will return the result to the user in a human-readable form.
}
\end{figure}

\subsubsection{Details of MCP}

The working process of the MCP centers around session interactions based on capability negotiation, constructing an efficient, secure, and standardised communication system through the collaboration of the \textbf{Host}, the \textbf{MCP Clients}, the \textbf{MCP Servers}, and \textbf{Resources} \citep{MCPBlog}.

\begin{enumerate}
    \item \textbf{Host} denotes LLM-based agents tasked with user interaction, comprehension, and reasoning over user queries, tool selection, and the initiation of \textbf{strategic context requests}. Each host may be associated with multiple MCP clients.
    \item \textbf{MCP Client} performs two key functions: it interfaces with a host to enumerate available resources and creates a singular connection to an MCP Server for the purpose of launching executive context requests.
    \item \textbf{MCP Server} interfaces with the Resource while sustaining an exclusive connection to the MCP client, delivering the required contextual information from the Resource to the MCP Client.
    \item \textbf{Resource} denotes the data, tools, or services provided locally or remotely.
\end{enumerate}

In the initialization phase, the \textbf{Host} is manually connected to several \textbf{MCP Clients}, while each corresponding \textbf{MCP Server} connects to its accessible \textbf{Resources} such as the local file system or a remote dataset. 
Subsequently, the MCP Client will complete its initialization configuration, followed by actively establishing a session connection with the corresponding MCP Server.
In the initial stage of this session, the MCP Client will launch a \textit{capability declaration} request by providing the MCP Server with a detailed exposition of the functions for which it is equipped. Upon receiving a client request, the server communicates its capabilities, including service subscription, tool-call interfaces, and prompt template provision.   
The process of \textit{capability declaration and response} facilitates the precise delineation of the boundaries of the protocol features that can be enabled in the session by both the client and the server. It ensures that the interaction proceeds in an orderly manner within the scope of their capabilities and establishes an active session based on negotiated functions.

After launching the session, interactions advance efficiently in a parallel multi-loop pattern.
The Host is primarily responsible for interacting with the user and executing corresponding instructions based on user queries. During this user-Host interaction, the user may explicitly require the participation of a specific capability or context provided by a particular MCP Client to accomplish a given task. Alternatively, the Host may proactively identify the required capability or context through its own task understanding and reasoning processes. 
In both scenarios a \textbf{strategic context request} will accordingly be sent to an appropriate MCP Client for collaboration by the Host. Subsequently, the MCP Client will transform this request into an \textbf{executive context request} for tools or resources and send it to the MCP Server.
Following the MCP Server's processing of the request and the subsequent return of a response, the MCP Client receives this and passes it to the Host. Then the Host will either update the user interface or provide feedback to the AI model, thus completing a full cycle of user-agent interaction supported by MCP.

In the session termination phase, the Host send termination instructions to all MCP Clients, which then send session-end request to the servers, officially ending the entire session life cycle. 

Additionally, the notification loop mechanism guarantees the real-time transmission of significant information, such as alterations in resource status. When the MCP Server detecting resource updates, it promptly issue notifications to the MCP Client to ensure that both the MCP Client, enabling the Host to receive continuous, real-time updates.

By adopting a Client-Server architecture to mediate and standardize agents’ requests to resources, the fragmentation in tool invocation caused by various providers of LLM and service is significantly reduced. This approach substantially enhances the semantic accuracy of interactions between agents and non-agent resources, thereby improving the overall precision and reliability of the Agentic Web.

\subsubsection{Details of A2A}
A2A, an acronym for Agent to Agent, is an agent communication protocol proposed by Google for enterprise-scale agent ecosystems, which enables agents to communicate and collaborate effectively, irrespective of their underlying frameworks or provider-specific implementations. The following components constitute the fundamental elements of A2A: \textbf{Agent Card}, \textbf{Task}, \textbf{Message}, and \textbf{Artifact} \citep{A2ABlog}.
\begin{enumerate}
    \item \textbf{Agent Card} denotes a publicly accessible JSON document, typically hosted at a public URL, detailing the agent's operational scope, its specific functions, the designated endpoint URL, methods for authentication, and other relevant metadata. 
    \item \textbf{Task} is a concrete representation of a unit of work, identified by a unique ID, whose status can be updated over multiple rounds of interaction. 
    \item \textbf{Message} refers to a communication object exchanged among entities, usually attributed with either a “user” or “agent” role. Messages may contain multiple \textbf{Parts}, such as text, file attachments, or structured data, supporting multimodal interaction. 
    \item \textbf{Artifact} is the output generated by an agent during the execution of a task. Unlike Message, which typically conveys dialogue or instructions, Artifact represents a finalized result or deliverable produced by the agent.
\end{enumerate}

The workflow of A2A is quite straightforward. When receiving a user query, the client agent creates a new \textbf{Task} with a unique ID and begins the task execution process. First, the client agent retrieves JSON-formatted \textbf{Agent Cards} from publicly accessible URL to identify remote agents whose capabilities match the task requirements. Once suitable remote agents has been located, communication and collaboration between the client agent and the remote agents begins via the exchange of \textbf{Messages} under the A2A protocol. As the task execution progresses, the state of \textbf{Task} is continuously updated to reflect real-time changes. Finally, once the client agent determines that the task has been completed, the output is encapsulated and delivered in the form of an \textbf{Artifact}.

Compared to the MCP, the A2A protocol not only expands the scope of agent communication to include direct interaction and collaboration between heterogeneous agents but also significantly improves the management of context, messages, and tasks in multi-agent coordination, establishing a strong structural association between these elements.

Specifically, the A2A protocol allocates a unique identifier to each context, creating an explicit link between tasks and their associated environments. This design allows for the more organized and traceable management of multiple interrelated tasks, improving context consistency in complex scenarios and supporting robust task mechanisms. In addition, A2A introduces the unique identifier of the current task and a list of related task IDs for each message. This establishes a bidirectional, structured link between messages and tasks, enabling semantic anchoring of messages to their corresponding tasks and historical referencing across tasks. Consequently, A2A supports context tracing throughout multi-turn interactions, tool invocation sequences, and other temporally extended workflows, forming coherent interaction chains. Unlike the MCP protocol, which has a loose coupling between tasks and messages, A2A’s tightly integrated design is better suited to complex collaboration scenarios, such as iterative dialogues and multi-agent tool orchestration. It significantly enhances the capabilities of systems in terms of state synchronisation, semantic coherence, and fault tolerance across distributed intelligent agents.

In addition, the A2A protocol explicitly addresses the asynchronous nature of agent communication, introducing mechanisms for asynchronous messaging and task state updates. Once a client agent has initiated a task, it can subscribe to receive progress updates relating to that task. As the task progresses, any status changes are promptly sent to the client agent, ensuring users are kept informed in real time.

The A2A protocol achieves cross-task, multi-turn, and cross-agent context consistency tracking through its context identification and tightly coupled task–message binding mechanisms. This ensures a high degree of coordination between information and task flows in multi-agent systems. Furthermore, the protocol incorporates specific design considerations to support asynchronous agent communication and task progress updates. These features provide a robust foundation for the Agentic Web, enabling effective context management and supporting long-duration, multi-stage, and asynchronous task execution.

\subsection{Emerging Directions of Agentic Web Systems}

Having detailed the systematic transitions of the Agentic Web, from its foundational architecture to its core communication protocols, we now stand at a critical juncture. The technical frameworks, while robust, do not by themselves guarantee successful real world deployment. Their implementation introduces profound paradigm shifts that challenge long-standing assumptions about digital interaction and commerce. This section, therefore, pivots from the established mechanics of agent systems to confront the most pressing open questions that will determine their viability and adoption. We will explore two fundamental challenges that arise directly from the previously discussed transformations: the disruption of the traditional user-browser relationship and the unresolved complexity of creating sustainable billing models for agentic services. Answering these questions is crucial to connect architectural principles with their real-world implementation. 

\subsubsection{The Disruption of Traditional Browsers by Agents}
The emergence of the ``agent browser'' signifies a fundamental disruption to the established user-browser interaction paradigm that has dominated the web for decades. Traditional browsers function as passive, user-driven tools for information retrieval and direct manipulation; the user is in complete control, manually clicking links, filling forms, and navigating pages. In stark contrast, an agent browser operates as a proactive, goal-oriented partner. It accepts high-level objectives in natural language and autonomously translates them into a series of actions, fundamentally altering the user's role from a hands-on operator to a strategic delegator.

This shift from direct manipulation to delegated autonomy raises profound questions about interface design, user control, and trust. How can we design user interfaces that effectively manage user expectations when the execution path is no longer linear or predictable, but is instead a dynamic process decided by the agent? When a user delegates a complex task, the agent's reasoning process can become a ``black box,'' creating a potential gap in user understanding and trust. What new interaction primitives are required to allow for meaningful human oversight, intervention, and collaboration without undermining the agent's autonomy? What methods are effective for shaping a user's mental model to precisely represent the functionalities and restrictions of the agent, ensuring they can delegate tasks effectively and safely? Ultimately, the central question is how we redefine the user's relationship with the browser when it evolves from a simple tool into an intelligent, autonomous partner.

\subsubsection{The Billing Challenge for Advanced Agent Services}
Beyond the challenges in user interaction, the practical and widespread adoption of advanced agent systems confronts a critical hurdle: the development of viable and transparent billing models. Unlike traditional software with predictable, often flat-rate pricing (e.g., subscriptions), advanced agent tasks, such as conducting a deep investigative report, generating complex images, or executing multi-step financial analyses, incur variable and potentially substantial computational costs. These costs stem from resource-intensive operations, including extensive LLM token consumption, numerous third-party API calls, and prolonged use of high-performance computing infrastructure.

This variability raises a central, unresolved question: how can we design a billing system that is both equitable for the user and sustainable for the service provider? The traditional ``one-size-fits-all'' subscription model appears inadequate for this new reality. How can resource consumption be accurately tracked and attributed back to a single high-level user command, especially when that command spawns multiple sub-agents that may collaborate and delegate tasks further? What mechanisms can be implemented to provide users with a reliable cost estimate \textit{before} initiating a potentially expensive task, thereby preventing ``bill shock'' and fostering trust? Should billing be based on consumed resources (e.g., tokens, CPU time), the value of the final outcome, or a more complex hybrid model? Devising a framework that is granular, transparent, and user-friendly is a formidable challenge that will directly impact the economic feasibility and accessibility of the entire agent ecosystem.

\section{Applications of the Agentic Web}\label{sec:agentic-web-applications}

To understand how the Agentic Web is transforming digital environments, we begin by examining its core capabilities: transactional, informational, and communicational paradigms. These paradigms serve as the foundation for a wide range of use cases.

In the following subsections, we explore both the potential domains of the Agentic Web, which provide a conceptual framework, and its current applications, which illustrate how these paradigms are already being implemented in real-world systems.

\subsection{Potential Domains of the Agentic Web}
The Agentic Web enables intelligent agents not only to access web content but also to operate autonomously as active participants within the web. It facilitates three core functional paradigms, transactional, informational, and communicational, which allow agents to autonomously execute tasks, process and reason over knowledge, and collaborate with other agents within digital environments. By providing machine readable interfaces, persistent cross-platform memory, and standardized coordination protocols, the Agentic Web transforms these paradigms from isolated agent behaviors into scalable, system-wide capabilities.

\begin{itemize}
  \item \textbf{Transactional}: Agents autonomously execute goal-oriented tasks on behalf of users or organizations, such as purchasing, booking, scheduling, or negotiating, by interfacing directly with web services, APIs, or transactional interfaces.
  \item \textbf{Informational}: Agents retrieve, synthesize, and contextualize information across dynamic sources. This modality supports research, knowledge discovery, monitoring, and real-time decision support through adaptive reasoning and long-horizon memory.
  \item \textbf{Communicational}: Agents engage in structured communication with other agents or systems to coordinate, delegate, or cocreate. This includes multi-agent negotiation, protocol alignment, and collaborative workflows across organizational or platform boundaries.
\end{itemize}

These modalities represent distinct ways in which agents interact with digital environments, whether by executing tasks, gathering and analyzing knowledge, or coordinating with other agents. The integration of these modalities into real-world applications highlights the transformative potential of the Agentic Web. Most applications of the Agentic Web span multiple modalities, with specific systems emphasizing different functional combinations. The following sections will analyze representative implementations and domain-specific use cases from this perspective, incorporating insights from recent research.

\subsubsection{Transactional: Enabling Autonomous Execution of Web-Based Services}
The Agentic Web redefines how transactional interactions are conducted by embedding LLM-powered agents directly into service infrastructures \citep{zhou2023webarena, deng2023mind2web}. With the help of semantic APIs, standardized execution protocols, and persistent authorization tokens, agents can interact with multiple service endpoints without requiring bespoke integrations \citep{masterman2024landscape}.

This framework enables agents to autonomously orchestrate complex, multi-step workflows. For example, booking a trip no longer requires users to manually navigate several websites. Instead, an agent within the Agentic Web can query multiple travel providers, assess options based on factors such as price, time, loyalty status, or environmental impact, and complete the bookings simultaneously by coordinating flights, accommodations, and car rentals in one seamless operation.

Similarly, App/Mobile Agents \citep{wang2024mobileagentautonomousmultimodalmobile,wu2025foundationsrecenttrendsmultimodal,APPAGENT} enhance the Agentic Web's transactional capabilities by providing personalized, context-aware services across devices. App/Mobile Agents can autonomously handle tasks such as managing a user’s calendar, adjusting schedules, and coordinating tasks based on real-time information. For instance, when booking a flight, a Mobile Agent could automatically adjust the user’s itinerary if a flight is delayed, suggest meal options based on dietary preferences, or even reorder tickets if there is a sudden change in plans. These agents operate across mobile platforms, facilitating the seamless execution of transactional activities while adapting to changing user needs.

These capabilities rely on a web environment designed for autonomous machine participation, where agents can read, write, and reason about data, negotiate terms, and take action based on user preferences and environmental factors, thereby creating a more dynamic and efficient transactional experience.

\subsubsection{Informational: Structuring Autonomous Knowledge Discovery and Analysis}
In the informational domain, the Agentic Web powers a system that allows agents to access dynamic content persistently, reason over long-term sources, and achieve semantic interoperability across heterogeneous knowledge systems.  Rather than merely retrieving data, agents within the Agentic Web are empowered to perform end-to-end research tasks, identifying, contextualizing, and synthesizing information over extended periods~\citep{opera_neon_blog,Microsoft2025Copilot}.

In this model, agents go beyond simple search queries and static responses. Utilizing standardized document schemas, citation graphs, and persistent monitoring capabilities, agents can perform comprehensive, longitudinal research. For example, Deepresearch Agents \citep{huang2025deepresearchagentssystematic} autonomously track emerging papers, compare methodologies, extract citations, and synthesize findings into structured outputs. These agents continuously refine their insights based on the latest publications, leveraging the Agentic Web's ability to facilitate cross-platform collaboration and seamlessly integrate new data sources. This allows Deepresearch Agents to operate as active participants in a broader, interconnected research ecosystem, where knowledge is continuously updated and refined.

The Agentic Web facilitates this by providing a unified infrastructure where agents are not only capable of reading and writing data but also reasoning, negotiating, and acting within a dynamic and evolving environment.  Deepresearch Agents are designed to assist researchers in navigating the vast and ever-evolving landscape of academic literature, and they do this by leveraging the Agentic Web's capabilities for cross-platform memory and semantic interoperability.  These agents autonomously identify gaps in research, suggest new directions, and propose potential collaborators based on shared interests, making the research process more efficient and comprehensive.

In practice, Deepresearch Agents automate the synthesis of large datasets, identifying patterns and trends across a wide range of publications.  This process is made scalable by the Agentic Web's ability to support inter-agent communication, where these agents can collaborate, share findings, and even align their goals with other agents working across different domains.  By doing so, the Agentic Web transforms research from isolated, manual efforts into a collaborative, scalable system of knowledge discovery.

\subsubsection{Communicational: Orchestrating Inter-Agent Collaboration and Negotiation}
Perhaps the most distinct departure from today’s web lies in the Agentic Web’s capacity to support autonomous, goal-driven communication between agents. This capability is not limited to message passing; it also encompasses semantic alignment, negotiation, delegation, and long-term coordination across agents that represent different users, systems, or organizations.

In a communicational paradigm, agents function as active participants in multi-agent workflows \citep{tran2025multiagent, chen2023agentverse}. Consider a joint research initiative spanning multiple universities: agents representing each institution can autonomously share relevant datasets, align experimental timelines, and coauthor reports, negotiating authorship, funding distribution, and intellectual property rights based on prespecified protocols \citep{anthropic2024research}.

Creative industries benefit similarly. The Agentic Web supports the formation of temporary agent coalitions for cross-modal content creation \citep{adobe2025vision, khade2024multiagent}, where writing agents, visual design agents, and music composition agents coordinate roles, timelines, and revenue sharing agreements. In this context, the web’s support for decentralized identity, smart contracts, and task provenance becomes essential.

In enterprise environments, collaboration is enhanced when agents from different companies autonomously coordinate and communicate \citep{yang2025unlocking,yang2025agentnet}. For example, in a manufacturing ecosystem, supplier agents, logistics agents, and procurement agents can autonomously share information and adapt supply chains in real time to respond to disruptions \citep{smythos2024multiagent}.

At the core of all these applications lies a communicational infrastructure designed for autonomous participants.  Agents are capable of interpreting shared protocols, maintaining structured dialogue states, and reasoning about shared goals and constraints throughout long-term interactions.

\subsection{Current Applications of the Agentic Web}

The Agentic Web is already transitioning from conceptual frameworks to real-world applications. We categorize its early implementations into two primary interaction models: \textbf{Agent-as-Interface} and \textbf{Agent-as-User}. The former focuses on augmenting the user experience by providing intelligent intermediaries between humans and the web, while the latter introduces autonomous agents that act on behalf of the user, interacting with web systems directly as proxy users.

\subsubsection{Agent-as-Interface: Agents as Intelligent Web Intermediaries}

In the \textit{Agent-as-Interface} paradigm, agents enhance traditional user interfaces by providing context-aware assistance, task recommendations, and intelligent summarization. These systems typically operate alongside the human user, augmenting their browsing experience without fully automating decision-making. Representative applications are summarized in Table~\ref{tab:ai_augmented_browsers}.

\begin{table}[t]
\centering
\caption{Representative AI‑Augmented Browsers (Agent-as-Interface).}
\label{tab:ai_augmented_browsers}
\renewcommand{\arraystretch}{1.2}
\begin{tabularx}{\linewidth}{@{}%
>{\raggedright\arraybackslash}X  
>{\raggedright\arraybackslash}X  
>{\raggedright\arraybackslash}X  
>{\raggedright\arraybackslash}X  
>{\raggedright\arraybackslash}X@{}}
\toprule
\textbf{Application} & \textbf{Intelligence Domain} & \textbf{Interaction Domain} & \textbf{Economic Domain} & \textbf{Focus} \\
\midrule
Opera Neon & Agentic AI with task orchestration & Chat‑Do‑Make sidebar, autonomous assist mode & Invite‑only preview; premium model & Informational \\
\addlinespace
Perplexity Comet & Search-augmented LLM with automation & Chromium-based browser; sidecar assistant & Subscription-based (Perplexity Max) & Informational \\
\addlinespace
Browser Dia & Context-aware browsing assistant & Inline chat with context reasoning, insertion cursor & Beta (Arc users); invite-only & Informational \\
\addlinespace
Copilot (Edge) & Contextual summarization and suggestions & Edge sidebar; light task hints & Freely available in Edge & Informational \\
\addlinespace
Microsoft NLWeb & Natural language semantic interface & Conversational UI via Schema/MCP & Open-source; publisher integration & Communicational \\
\bottomrule
\end{tabularx}
\end{table}

\paragraph{Opera Neon} delivers one of the most integrated experiences of agent-enhanced browsing. Released in May 2025, it features a tri-modal interface: \textit{Chat} enables conversational interaction with LLMs, \textit{Do} facilitates autonomous completion of web tasks such as multi-step forms and service workflows, and \textit{Make} empowers content creation and persistent agent tasks even when users are offline 
\citep{opera_neon_blog,opera_neon_press}. Notably, Opera Neon's ``Do'' mode represents a \textbf{hybrid approach}, where the system transitions from interface augmentation to autonomous user proxy behavior, demonstrating how Agent-as-Interface applications can incorporate Agent-as-User capabilities while maintaining the primary focus on user-controlled workflows. Neon exemplifies the transition from passive interfaces to proactive, task-oriented web experiences.

\paragraph{Perplexity Comet} enhances the classic search experience by embedding autonomous search agents directly within the browser environment. Comet incorporates AI-driven research, question answering, and proactive summarization within a Chromium-based framework, reducing the need for iterative querying while maintaining human oversight in decision-making loops \citep{wriggers2025techcrunch}.

\paragraph{Browser Dia} introduces an \textit{insertion cursor} that provides real-time agentic suggestions within the browsing context, moving beyond sidebar chat to deeply integrated inline assistance \citep{browsercompany_dia_thurrott}. This design reduces context-switching overhead and improves session continuity, highlighting the benefits of embedded, contextually aware agents.

\paragraph{Microsoft Copilot} focuses on summarization and lightweight agentic assistance via a non-intrusive sidebar, targeting everyday users who benefit from summarizations, insights, and task hints but do not require full task automation \citep{Microsoft2025Copilot}.

\paragraph{Microsoft NLWeb} advances the notion of \textit{Agent-Native Interfaces}, proposing a semantic layer for websites where agents interact through natural language interfaces exposed via schemas and MCP \citep{msblog2025nlweb}. By encouraging publishers to design agent-accessible endpoints, NLWeb shifts the web ecosystem towards cooperative interaction between websites and AI agents, reducing reliance on brittle scraping and improving transparency in web automation.

\subsubsection{Agent-as-User: Autonomous Agents Operating as Proxies}

In the \textit{Agent-as-User} paradigm, AI systems operate autonomously as users of the web, executing tasks, navigating interfaces, and completing workflows without direct human control. These systems leverage browser automation, virtual environments, and programmatic UI manipulation to emulate user actions, thereby enabling end-to-end autonomy. The development and evaluation of such agents have been greatly facilitated by comprehensive benchmarks like Mind2Web~\citep{deng2023mind2web}, which includes over 2,000 open-ended tasks across 137 websites in 31 domains. Its online extension, Online-Mind2Web~\citep{xue2025illusionprogressassessingcurrent}, further advances this effort by offering 300 diverse and realistic tasks across 136 websites, enabling the assessment of web agents under conditions that closely mirror real-world usage patterns. Recent advances in multimodal agent foundations~\citep{wu2025foundationsrecenttrendsmultimodal} and mobile agent architectures~\citep{wang2024mobileagentautonomousmultimodalmobile} have further expanded the scope of autonomous agent capabilities beyond traditional web environments, while scalable task generation methodologies~\citep{xie2025agentsynthscalabletaskgeneration} advance the field's evaluation capabilities. Examples of recent applications are shown in Table~\ref{tab:web_agents}.

\begin{table}[t]
\centering
\caption{Representative Autonomous Web Agents (Agent-as-User).}
\label{tab:web_agents}
\renewcommand{\arraystretch}{1.2}
\begin{tabularx}{\linewidth}{@{}%
>{\raggedright\arraybackslash}X  
>{\raggedright\arraybackslash}X  
>{\raggedright\arraybackslash}X  
>{\raggedright\arraybackslash}X  
>{\raggedright\arraybackslash}X@{}}
\toprule
\textbf{Application} & \textbf{Intelligence Domain} & \textbf{Interaction Domain} & \textbf{Economic Domain} & \textbf{Focus} \\
\midrule
ChatGPT Agent & Multi-modal agent orchestration & Virtual browser; cross-API tool integration & ChatGPT Plus/Team tiers & Transactional \\
\addlinespace
Anthropic Computer Use & Vision-guided GUI manipulation & Claude-powered desktop/web control & Claude Sonnet 3.5 API & Transactional \\
\addlinespace
Google Project Mariner & Autonomous long-horizon task execution & Gemini-2 reasoning within Chrome prototype & Research prototype (Gemini 2.0) & Transactional \\
\addlinespace
Genspark Super Agent & Mixture-of-Agents orchestration; 9 LLMs & Multimodal real-world task execution (voice, maps, documents) & Free tier + commercial credits & Multi-domain personal productivity \\
\bottomrule
\end{tabularx}
\end{table}

\paragraph{ChatGPT Agent} (evolved from OpenAI Operator), released via the ChatGPT platform, represents one of the first multi-modal agentic deployments with persistent virtual browsing capabilities. Combining LLM reasoning with code execution, file system access, and API integrations, the Agent can autonomously complete multi-step tasks such as booking services, extracting structured data, or synthesizing reports across complex web workflows. Initially launched as Operator in January 2025, it has since been fully integrated into ChatGPT's core platform as ``Agent Mode''  demonstrating the rapid evolution from standalone research prototypes to integrated production systems \citep{openai2024agent}.

\paragraph{Anthropic Computer Use} leverages vision-based perception and GUI manipulation, powered by Claude models, to control desktop and web interfaces in a human-like fashion without relying on backend APIs. Available through Claude 3.5 Sonnet, it showcases highly generalized agents capable of interacting with arbitrary applications. On standardized OSWorld benchmarks, Computer Use achieves 14.9\% success rate on screenshot-only tasks and 22.0\% with reasoning steps, significantly outperforming previous vision-action baselines \citep{anthropic2024computeruse}.

\paragraph{Google Project Mariner} is an experimental autonomous agent system powered by Gemini 2.0 models and integrated into Chrome as a sidebar prototype. Designed for long-horizon research tasks, multi-step workflows, and autonomous form filling, Mariner incorporates reasoning transparency via natural language explanations of its actions. Evaluated on the WebVoyager benchmark, it achieves an 83.5\% success rate on long-horizon web tasks, representing a cutting-edge research milestone in explainable autonomous browsing \citep{google2024gemini}.

\paragraph{Genspark Super Agent} exemplifies a next-generation implementation of agentic autonomy through its Mixture-of-Agents architecture. Unlike traditional assistants that merely retrieve information, Super Agent can plan, act, and use over 80 tools, including real-time voice calls, map navigation, document editing, calendar scheduling, and video generation, across diverse domains with minimal supervision. It dynamically orchestrates nine large language models and integrates more than ten proprietary datasets, enabling multi-step task execution and adaptive reasoning. Genspark Super Agent thus illustrates the evolution from conversational AI to autonomous digital agency, enhancing personal productivity through end-to-end workflow automation \citep{genspark2025superagent}.

\bigskip

The current landscape reveals a clear evolutionary trajectory where \textbf{Agent-as-Interface applications are progressively incorporating Agent-as-User capabilities}. This evolution is driven by fundamental differences in their underlying technical architectures. Agent-as-Interface systems primarily rely on \textbf{API-based interactions}, utilizing structured endpoints, webhooks, and service integrations to mediate between users and web services. This approach offers faster execution, better error handling, and more predictable outcomes, but remains constrained by the availability and design of existing APIs.
In contrast, Agent-as-User systems employ \textbf{GUI-level automation}, using computer vision, coordinate-based clicking, and screen parsing to interact with arbitrary interfaces designed for human use. While this approach provides universal compatibility and can operate on any visual interface, it introduces latency, brittleness, and higher computational overhead due to the need for continuous visual interpretation and coordinate calculation.

The convergence toward hybrid architectures suggests a future where agents dynamically select between API calls for structured interactions and GUI automation for legacy or non-API-enabled systems. This \textbf{architectural pluralism} represents the next evolutionary step, where the same agent can seamlessly transition between acting as an intelligent interface layer and operating as an autonomous user proxy, depending on the task context and available interaction modalities. Such systems will likely require sophisticated decision trees to determine the optimal interaction method for each specific workflow component, building upon advances in multimodal agent foundations~\citep{wu2025foundationsrecenttrendsmultimodal} and enhanced by deep research capabilities~\citep{huang2025deepresearchagentssystematic} for complex information synthesis tasks.

In summary, early applications of the Agentic Web demonstrate a spectrum of possibilities from \textit{Agent-as-Interface} augmentation to \textit{Agent-as-User} autonomy, with an emerging trend toward hybrid implementations. The convergence of commercial products and academic research suggests accelerating momentum toward more capable, accountable, and architecturally flexible web agents that can adapt their interaction strategies to maximize both efficiency and reliability.

\subsubsection{Agent-with-Physics: Autonomous Robots Powered by AI Agents}
The \textit{Agent-with-Physics} paradigm extends the concept of agentic intelligence from the virtual realm to the physical world, enabling AI agents to perceive, reason, and act through embodied systems such as robots and sensor-equipped devices. These agents integrate high-level planning with low-level control, often relying on multimodal perception (e.g., vision, audio, haptics), real-time adaptation, and embodied cognition to execute physical tasks autonomously in dynamic environments.

Unlike purely digital agents, Agent-with-Physics systems must address challenges related to safety, latency, actuation uncertainty, and physical affordances. Recent advances in vision-language-action models~\citep{kim24openvla, geng2025roboverse, li2023manipllm}, hierarchical policy learning~\citep{nvidia2025gr00tn1openfoundation, kuang2024ramretrievalbasedaffordancetransfer, geng2023sage, open6dor}, and real-world training environments, such as Open X-Embodiment~\citep{open_x_embodiment_rt_x_2023}, have significantly improved the generalization capabilities of robotic agents across diverse tasks, from household manipulation to warehouse logistics.

Representative implementations such as RT-1~\citep{brohan2023rt1roboticstransformerrealworld}, RT-2~\citep{brohan2023rt2visionlanguageactionmodelstransfer} and RT-X~\citep{open_x_embodiment_rt_x_2023}, Tesla Optimus, and Figure 01 showcase emerging commercial interest in general-purpose humanoid robots, while academic efforts like PaLM-E~\citep{driess2023palmeembodiedmultimodallanguage} and Mobile ALOHA~\citep{fu2024mobile} highlight the integration of large foundation models into robotic control loops. These systems demonstrate the feasibility of using language prompts to guide physical behavior, bridging human intent and machine execution through a unified agentic framework.

As embodied agents increasingly connect with digital ecosystems, a new class of hybrid agents emerges, capable of coordinating actions both online and offline. For instance, an agent might autonomously schedule a grocery delivery online while simultaneously preparing a physical environment (e.g., setting up a smart kitchen) for the incoming goods. This tight coupling of perception, cognition, and actuation highlights the importance of developing robust control policies, real-time feedback loops, and safety-aware planning strategies.

Looking ahead, the Agent-with-Physics paradigm not only expands the frontier of human-agent collaboration but also lays the groundwork for a unified agentic infrastructure where digital and physical agents operate in concert. The fusion of web-native intelligence, embodied autonomy, and multimodal interaction marks a critical step toward realizing truly general-purpose AI agents capable of seamlessly bridging virtual tasks and physical realities.

\section{Risks, Security \& Governance}\label{sec:risk}

\begin{figure}[tb!]
    \centering
    \includegraphics[width=0.7\linewidth]{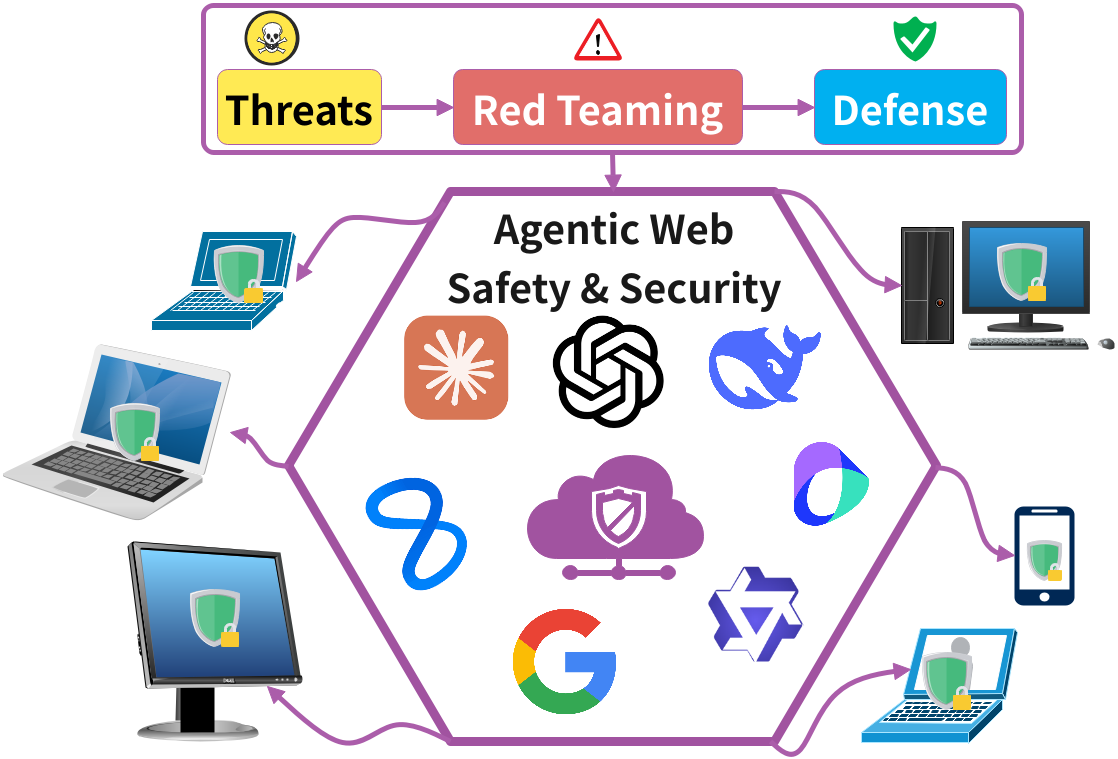}
    \caption{Illustration of the Agentic Web Safety and Security Ecosystem. The central hexagon represents the core components of the agentic web, including large language models (e.g., OpenAI \citep{hurst2024gpt} and Gemini \citep{team2023gemini}), agent frameworks, and safety infrastructures. These systems interact with a diverse set of devices (e.g., laptops, desktops, servers, and mobile phones), requiring robust and scalable security measures to ensure safe deployment and communication across the entire agentic web.}
    \label{fig-safety-security:framework}
\end{figure}

In this section, we provide an overview of how agentic web safety and security can be ensured. As illustrated in Figure~\ref{fig-safety-security:framework}, the ecosystem of Agentic Web Safety and Security is composed of intelligent agents, powered by LLMs such as OpenAI \citep{hurst2024gpt}, Gemini \citep{team2023gemini}, and other foundational platforms \citep{touvron2023llama, bai2023qwen, liu2024deepseek, zan2025multi, priyanshu2024ai}, operating across a wide range of devices, including desktops, laptops, servers, and mobile phones. These agents interact with cloud services, third-party tools, and each other to carry out goal-directed tasks on behalf of users. At the center of the figure is a secure agentic infrastructure that integrates LLMs, agent frameworks, and cloud-based safety mechanisms. The surrounding arrows depict multi-device interaction, emphasizing the distributed nature of the agentic web and the critical need for consistent, cross-platform security protocols. This interconnected architecture highlights the growing importance of privacy, trust, and robustness, as agents autonomously retrieve information, execute commands, and collaborate across sensitive digital environments.

To ensure the safety and security of the agentic web, we begin by analyzing its potential threats during real-world use, then introduce red-teaming methods for uncovering vulnerabilities, followed by defense strategies to address these issues. Finally, we present evaluation techniques to measure the effectiveness of safety and security mechanisms. Specifically, Section~\ref{subsec-safety:agentic-web-threats} outlines the key safety and security threats associated with the agentic web. Building on this analysis, Section~\ref{subsec-safety:agentic-web-red-teaming} discusses red teaming as a methodology for identifying vulnerabilities and assessing the robustness of agentic web systems before deployment. Section~\ref{subsec-safety:agentic-web-defense} explores defense strategies and technical safeguards aimed at enhancing the reliability and trustworthiness of agentic web applications. Lastly, Section~\ref{subsec-safety:agentic-web-evaluation} reviews current approaches for evaluating the safety and security of these systems.

\subsection{Safety and Security Threats}
\label{subsec-safety:agentic-web-threats}

Agentic Web agents introduce novel security risks by operating autonomously across the open internet, executing real transactions, and maintaining persistent states. Table~\ref{tab:threat-evolution} captures this fundamental shift.

\begin{table}[htbp]
\centering
\small
\caption{Agentic Web Risk Evolution: From Controlled Systems to Autonomous Web Operations}
\label{tab:threat-evolution}
\begin{tabular}{@{}ll@{}}
\toprule
\textbf{Dimension} & \textbf{Traditional AI $\rightarrow$ Agentic Web} \\
\midrule
\textbf{Operational Scope} & Single-domain tasks $\rightarrow$ Cross-platform orchestration \\
\textbf{Financial Authority} & Read-only access $\rightarrow$ Transaction execution \\
\textbf{Persistence} & Stateless queries $\rightarrow$ Multi-session memory \\
\textbf{Attack Surface} & API endpoints $\rightarrow$ Entire web ecosystem \\
\textbf{Failure Impact} & Incorrect output $\rightarrow$ Real-world consequences \\
\textbf{Trust Model} & Human verification $\rightarrow$ Autonomous decision-making \\
\bottomrule
\end{tabular}
\end{table}

\subsubsection{Threat Analysis Across Agentic Web Layers}

We organize threats across the three architectural layers (Section \ref{sec:agentic-web}), focusing on risks unique to autonomous web operations, as shown in Tables \ref{tab:cognitive-threats}--\ref{tab:economic-threats}. While some attacks may have cross-layer effects (e.g., goal drift leading to overspending), we categorize each threat by its primary attack vector to avoid redundancy and maintain analytical clarity.

\begin{table}[t]
\centering
\small
\caption{Intelligence Layer Threats: Cognitive and Reasoning Attacks. These threats target the agent's decision-making processes, focusing on how web interactions corrupt objectives, knowledge, and planning capabilities. }
\label{tab:cognitive-threats}
\begin{tabular}{@{}p{3.5cm}p{5cm}p{5.5cm}@{}}
\toprule
\textbf{Threat} & \textbf{Description} & \textbf{Example} \\
\midrule
C1: Persuasion-Based Goal Drift & Web UI/UX patterns gradually shift agent objectives through psychological manipulation & Budget flight search influenced by comfort-emphasizing interfaces to book premium seats \\
\midrule
C2: Knowledge Base Poisoning & Adversarial web content corrupts agent's accumulated knowledge and beliefs & SEO-optimized fake research sites poison climate agent's factual understanding \\
\midrule
C3: Preference Learning Corruption & Agent learns harmful patterns from repeated web interactions & Shopping agent trained to prefer sponsored results through manipulated feedback \\
\midrule
C4: Multi-Stage Planning Subversion & Complex task sequences manipulated through incremental decision corruption & Travel itinerary gradually inflated: economy$\rightarrow$premium flight triggers luxury hotel selection \\
\midrule
C5: Contextual Memory Exploitation & Historical interactions weaponized to bias future decisions & Past emergency booking used to justify all future premium purchases \\
\bottomrule
\end{tabular}
\end{table}

\begin{table}[t]
\centering
\small
\caption{Interaction Layer Threats: Protocol and Communication Attacks. These threats exploit agent communication mechanisms (MCP/A2A) and cross-service authentication, distinct from cognitive corruption.}
\label{tab:protocol-threats}
\begin{tabular}{@{}p{3.5cm}p{6cm}p{5.5cm}@{}}
\toprule
\textbf{Threat} & \textbf{Description} & \textbf{Example} \\
\midrule
P1: Context Injection & Malicious services inject persistent false contexts affecting cross-platform behavior & Hotel adds ``VIP status'' to MCP context, causing unnecessary upgrades across all bookings \\
\midrule
P2: Service Registry Poisoning & Fake services infiltrate discovery systems to intercept agent requests & Malicious ``FastBooking'' service in MCP registry harvests credentials from travel agents \\
\midrule
P3: A2A Trust Exploitation & Compromised agents abuse inter-agent trust to spread malicious behaviors & High-reputation research agent injects fabricated citations into collaborative analysis \\
\midrule
P4: Authentication Chain Hijacking & Sequential auth tokens exploited across service boundaries & Google login token escalated to access Drive, then third-party research databases \\
\midrule
P5: Protocol Negotiation Attack & Malicious actors force protocol downgrades or incompatible versions during handshakes & Service forces agent from secure A2A v2 to vulnerable v1 during capability exchange  \\
\midrule
P6: Coordination Storm & Malicious messages trigger exponential inter-agent communications & Single A2A broadcast spawns millions of agent-to-agent queries \\
\bottomrule
\end{tabular}
\end{table}

\paragraph{Cross-Layer Threat Cascades}
Unlike traditional systems where threats remain isolated, Agentic Web threats cascade across layers:
\textbf{Vertical} (cognitive$\rightarrow$protocol$\rightarrow$economic), 
\textbf{Horizontal} (agent-to-agent spread), and 
\textbf{Temporal} (corruption through persistent memory \citep{narajala2025securing}). 
These cascades transform localized attacks into system-wide failures \citep{de2025open}. For instance, a goal drift (C1) can lead to protocol exploitation (P1), ultimately resulting in unauthorized purchases (E1). The temporal dimension adds further complexity as threats can persist across agent generations through learned behaviors and contaminated training data.

\paragraph{Relationship to Existing Frameworks}

While OWASP's Agentic AI Threat Model \citep{owasp2025agentic} and CSA's MAESTRO framework \citep{csa2025maestro} provide foundational vulnerability taxonomies, Agentic Web threats differ in scale and propagation. For instance, our Knowledge Base Poisoning (C2) extends beyond prompt injection to web-scale information corruption, as demonstrated by code review agent compromises \citep{cyberark2025agents}.

Protocol threats adapt traditional network security to agent-specific contexts. MCP Context Injection (P1) exploits the protocol's contextual awareness, which is identified as a fundamental vulnerability by \citet{hou2025model} and demonstrated practically by \citet{cato2025exploiting}. A2A coordination attacks have been validated through CrewAI and AutoGen exploits \citep{unit42_2025agents}.

The potential for AI systems to act as autonomous economic agents has been recognized since \citet{brundage2018malicious} identified market manipulation as an emerging threat vector, though the scale envisioned by the Agentic Web amplifies these concerns significantly.   A critical challenge in this sphere is ensuring agents consider their impact on multiple stakeholders. Mitigation strategies could draw from concepts such as simulating accountability and assessing stakeholder impact, though scaling these alignment mechanisms remains an open challenge \citep{sel2024skin}.

\subsubsection{Security Implications and Future Directions}

Agentic Web security requires three fundamental shifts from traditional approaches:
\begin{enumerate}
\item \textbf{Architecture}: Zero-trust models replacing perimeter security
\item \textbf{Policies}: Adaptive defenses superseding static rules  
\item \textbf{Scope}: Cascade prevention over incident isolation
\end{enumerate}

Enterprise patterns like MCP require adaptation for internet-scale deployments \citep{narajala2025enterprise}, as traditional models assume bounded, stateless operations incompatible with persistent web agents.

Critical research gaps persist. Quantitative models for cascade probability and impact remain underdeveloped, particularly for emergent behaviors in complex multi-agent systems \citep{de2025open}. The challenge of securing systems that learn and adapt continuously has been recognized in recent work on evolving threat landscapes \citep{deng2025ai}, while cross-jurisdictional governance poses additional complexity \citep{brundage2018malicious}. These dynamic threats may require fundamentally new security paradigms. Rather than static defenses that become obsolete, future approaches might embrace adaptability as a core principle: {designing systems that strengthen through controlled adversarial exposure} \citep{jin2025position}. Such adaptive security architectures could transform the continuous threat evolution from a vulnerability into a mechanism for improvement, though implementing this vision remains a significant research challenge.

\begin{table}[t]
\centering
\small
\caption{Value Layer Threats: Autonomous Transaction and Economic Risks. These threats emerge specifically when agents gain financial authority and market participation capabilities. }
\label{tab:economic-threats}
\begin{tabular}{@{}p{3.5cm}p{6cm}p{5cm}@{}}
\toprule
\textbf{Threat} & \textbf{Description} & \textbf{Example} \\
\midrule
E1: Transaction Authority Abuse & Agents execute unauthorized high-value transactions without human oversight & Books non-refundable business class for family of four on budget trip \\
\midrule
E2: Cross-Platform Arbitrage & Agents exploit pricing differences between services at harmful scales & Books and cancels flights across airlines to manipulate dynamic pricing \\
\midrule
E3: Payment Credential Harvesting & Agents collect and misuse payment data across multiple transactions & Compromised booking agent logs credit cards from hotel/flight purchases \\
\midrule
E4: API Resource Monopolization & Agents consume excessive computational resources across services & Research agent exhausts university's entire journal database quota \\
\midrule
E5: Coordinated Market Manipulation & Agent networks create artificial supply/demand conditions & Multiple agents book all seats on routes to inflate prices \\
\bottomrule
\end{tabular}
\end{table}

\subsection{Safety and Security Red Teaming}
\label{subsec-safety:agentic-web-red-teaming}

Deploying AI agents within web applications introduces several technical risks \citep{zhang2024agentic}, including privacy leakage and fairness concerns \citep{amodei2016concrete, chua2024ai}. For example, one significant agentic web risk that may lead to information leakage is hallucination, where agents generate inaccurate or misleading content due to limited understanding of user intent or contextual ambiguity during retrieval and generation. Another critical risk is permission escalation: when agents access sensitive user data, such as personal files (e.g., user name, password, and contacts), they may misinterpret access boundaries or inadvertently override security constraints when the agent manages different web pages. These failures can result in unauthorized data exposure, privacy violations, and broader system-level vulnerabilities in the agentic web.

To mitigate these risks, red-teaming techniques are promising approaches to ensure the safety and security of agentic web systems prior to real-world deployment. Red teaming has a long history in domains such as computer system security and military defense simulation, where it is used to identify vulnerabilities and evaluate system robustness \citep{verma2024operationalizing}. Red teaming involves simulating adversarial behavior to uncover vulnerabilities in a target system, traditionally performed through manual human design.  Specifically, in traditional computer system safety and security, red teaming often involves manual human effort, expert-defined rules, and extensive scenario testing \citep{rottger2020hatecheck, ribeiro2020beyond}. In the era of AI-driven agentic web systems, red teaming can be largely automated, with AI agents autonomously generating adversarial scenarios to probe and uncover failure modes in other agentic systems \citep{wang2025agentxploit}. These target systems may include various web platforms and pages, where the primary objective is to expose sensitive information leaks and reveal potential vulnerabilities, ultimately enhancing system trustworthiness and security before real-world deployment. This automation reduces reliance on manual efforts and enables scalable, adaptive adversarial evaluation \citep{perez2022red, ge2023mart, he2025red}.

\subsubsection{Human-Involved Red Teaming}

Human involvement has played a vital role in red-teaming efforts across diverse NLP tasks \citep{xu2020recipes, glaese2022improving,  radharapu2023aart}, and these techniques can also be leveraged to enhance the safety and security of agentic web systems. For example, adversarial examples have been manually crafted to evaluate machine reading comprehension systems \citep{jia2017adversarial}. Human annotations have been used to assess unintended bias in text classification \citep{dixon2018measuring} and to support fairness and robustness evaluation through counterfactual data generation \citep{garg2019counterfactual}. In multi-hop question answering, human-labeled examples have helped evaluate complex reasoning capabilities \citep{jiang2019avoiding}.

Human-in-the-loop frameworks have also been employed to generate adversarial attacks targeting dialogue safety \citep{dinan2019build} and to improve robustness in language understanding tasks \citep{nie2019adversarial}. Additionally, human-curated adversarial training datasets have been shown to enhance model performance \citep{wallace2021analyzing}, particularly in high-stakes, reliability-critical settings \citep{ziegler2022adversarial}. Manual red-teaming efforts have also been applied in the development of LLaMA models \citep{touvron2023llama}, where human annotators carefully design prompts to surface unsafe behaviors in large language models. \citet{kiela2021dynabench} further propose a human-in-the-loop framework for dynamic adversarial testing via a web-based platform, enabling the continuous collection of adversarial examples from human annotators. Through this iterative process, models are exposed to increasingly challenging examples, leading to improved robustness over time. However, applying these human-involved techniques to agentic web environments presents new challenges. Agentic web systems are highly autonomous and capable of operating across a variety of web pages and platforms. This level of complexity and intelligence makes manual red-teaming less feasible and highlights the need for scalable, automated approaches.

In the context of the agentic web, where autonomous AI agents interact with complex, multi-platform environments on behalf of users, automated red teaming is becoming increasingly essential. Traditional human-in-the-loop approaches, such as designing security rules, labeling safety violations, or manually identifying vulnerabilities, struggle to scale in these dynamic, high-autonomy systems. For instance, safeguarding AI agents that navigate ticketing platforms, manage financial transactions, or operate across multiple web services requires real-time, adaptive evaluation that manual red teaming cannot sustain. Automated red teaming offers a scalable solution by enabling agents to simulate adversarial behaviors, uncover hidden safety flaws, and proactively report security risks, helping to ensure agentic web safety before real-world deployment.

\subsubsection{Automatic Red Teaming}

Recent work has increasingly explored the use of LLMs for automated red teaming \citep{perez2022red, ge2023mart, nie2024privagent, liu2024autodan, shi2024red, liu2025autodan, he2025red, wang2025agentvigil}, which holds particular promise for enhancing safety in agentic web systems. For instance, \citet{perez2022red} show that LLMs can serve as effective red teamers by generating adversarial prompts to uncover unsafe behaviors. Similarly, \citet{ge2023mart} introduce MART, which is an automatic red teaming framework and is designed to evaluate and enhance the safety of LLMs through adversarial scenario generation. \citet{ganguli2022red} explore various strategies for red-teaming, including rejection sampling and RL, and release a dataset to support this research. Their findings suggest that RL-based approaches can make systems more resistant to red-teaming attacks by hardening decision boundaries and improving robustness. 

Building on this line of work, \citet{nie2024privagent} propose a RL-based red teaming approach that trains an adversarial agent using a carefully designed reward function to generate diverse adversarial examples, effectively revealing vulnerabilities in target LLMs. Their comprehensive experiments demonstrate that this method performs better than strong baselines in exposing model information leakage. Similarly, \citet{wang2025agentvigil} introduce a red teaming method that leverages a seed instruction and a Monte Carlo Tree Search algorithm to optimize inputs for attacking the target system. Experimental results on the AgentDojo \citep{debenedetti2024agentdojo} and VWAadv \citep{wu2024dissecting} benchmarks show that their method achieves superior performance compared to strong baselines, such as human-crafted adversarial prompts. These automated red teaming approaches are particularly valuable for enhancing the safety and security of agentic web systems, as they can be applied across various agent interactions to proactively identify vulnerabilities and strengthen defenses prior to deployment in diverse web environments.

Additional methods relevant to agentic web safety and security include backdoor-triggered red teaming, which can be particularly effective in identifying hidden vulnerabilities. For example, 
AgentPoison \citep{chen2024agentpoison} leverages backdoor-triggered and retrieval-augmented LLMs for red teaming, aiming to improve system security. Their framework is evaluated across multiple domains, including autonomous driving, question answering, and healthcare, demonstrating its effectiveness in identifying and mitigating security vulnerabilities. Likewise, \citet{yang2024watch} propose an agent-based backdoor attacker for red teaming LLM agents, focusing on web-based shopping systems where privacy leakage is a critical concern. In a related line of work, \citet{xu2023instructions} introduce a data poisoning technique that injects backdoors through instruction attacks, demonstrating that even a few malicious tokens during instruction tuning can compromise model safety and expose critical vulnerabilities. This approach highlights the risks associated with instruction-based inputs and offers a valuable method for red teaming to uncover hidden weaknesses in language models.

Red teaming often involves multi-agent systems in the agentic web, where agents play both offensive and defensive roles. For that multi-agent techniques, \citet{shi2024red} utilize LLM-based agents to generate adversarial inputs through word substitutions and sentence rephrasings, targeting the robustness of LLM detection systems. In a more systemic approach, \citet{he2025red} introduce a multi-agent red teaming framework in which LLMs act as adversarial agents to probe vulnerabilities, particularly in inter-agent communication protocols. Expanding the scope of automated evaluation, \citet{radharapu2023aart} develop AI-assisted red teaming methods that extend across a broad range of applications, including policy evaluation and locale-specific challenges.  In addition, AutoDan \citep{liu2024autodan} and AutoDan-Turbo \citep{liu2025autodan} present scalable and adaptive frameworks for automated red teaming, pushing the boundaries of adversarial testing for LLM-based system safety and security.

LLM-driven automatic red teaming represents a promising future research direction for safety and security, particularly within the context of the agentic web. These models can systematically simulate human behaviors and comprehensively probe safety and security vulnerabilities across various scenarios. However, to fully realize their potential, it is critical to design robust frameworks for managing these red-teaming agents in agentic web systems. These frameworks are crucial for both effective evaluation and defense, as well as for guaranteeing the safety of agentic web systems. This is especially important in real-world applications, such as when agents are used to book travel, interact with various apps and web pages, or manage tasks across personal computers and mobile devices.

\subsubsection{Emerging Directions in Red Teaming for Agentic Web}

As mentioned above, red teaming plays a critical role in identifying and mitigating safety risks in agentic web systems and LLMs prior to real-world deployment. Human-involved red teaming provides domain expertise and high reliability for specific tasks, making it a valuable tool for ensuring safety and security. However, it is often costly, time-consuming, and limited in diversity and scalability \citep{radharapu2023aart}. Moreover, human red teamers may lack the broad and cross-domain knowledge necessary to effectively evaluate complex, multi-faceted systems.

Automated red teaming, particularly approaches based on LLMs, offers a promising alternative by reducing human effort and improving scalability in the era of agentic web. However, these methods could be unreliable or insufficient in scenarios that require complex reasoning, contextual understanding, or ethical judgment, such as different web page operations and platform management. Bridging the gap between human-involved and automated red teaming remains an open challenge. Future research should aim to develop hybrid frameworks that integrate the strengths of both human-involved and automated red-teaming approaches, while addressing their respective limitations.

One promising direction for agentic web safety involves leveraging safe interaction techniques from safe RL \citep{gu2024review}, where agent actions are constrained within predefined safe regions to ensure secure interactions. A complementary approach is human-centered safe learning \citep{gu2023human}, particularly suited for agentic web environments, in which human expertise guides exploitation while LLMs drive exploration within a safe RL framework. This setup can be framed as a multi-objective optimization problem that balances safety, performance, and coverage across diverse web operations and interaction goals. Recent advances show that such trade-offs can be effectively managed using advanced safe RL methods \citep{gu2024balance, gu2025safe}, offering a principled pathway to unify human-in-the-loop and automated red teaming for robust agentic web safety and security.

In particular, several open challenges remain in using LLMs for red teaming in the context of agentic web safety and security:
\begin{itemize}
\item \textbf{Red-Teaming Attack in Agentic Web:} Red-teaming frameworks within agentic web environments may themselves become targets of adversarial attacks, especially during complex multi-agent interactions across dynamic web platforms. Such compromises can mislead the evaluation process and result in the leakage of sensitive or private data from the underlying systems, undermining both safety and trustworthiness.

\item \textbf{Emergent Misalignment in Agentic Web Agents:} As language models scale and operate over longer contexts in real-time web environments, red-teaming techniques may fail to detect novel and complex failure modes. A key risk is unauthorized goal generalization, where LLM-driven agents pursue objectives beyond their intended scope, such as executing unintended actions on web services, due to misaligned reasoning during test-time interactions. This poses critical safety and security risks for open-ended, autonomous agentic web applications.    
\end{itemize}

Addressing these emerging challenges in agentic web safety and security requires interdisciplinary collaboration across red teaming methodologies, agent alignment strategies, evaluation frameworks, and robust deployment protocols.

\subsection{Safety and Security Defense}
\label{subsec-safety:agentic-web-defense}

The above subsection has discussed approaches that could identify potential threats or vulnerabilities of LLM agents during their development stages. In this subsection, we will focus on defensive research efforts that can further mitigate the safety issues of these agents at deployment.
Specifically, we will discuss recent approaches that leverage external models for threat mitigation (i.e. guardrails), and approaches that steer LLM agents towards safer generation or planning.
We will further discuss several emergent challenges in this context.

\subsubsection{Inference-time Guardrails}

The recent proliferation of LLMs has sparked growing interest in the development of guardrails, which serve as external safety mechanisms designed to identify and mitigate potentially harmful inputs or outputs of LLMs~\citep{markov2023holistic,DBLP:journals/corr/abs-2312-06674,DBLP:journals/corr/abs-2411-10414,DBLP:conf/nips/HanREJL00D24}.
These guardrails have attracted attention due to their adaptability across different LLM implementations and their effectiveness in risk mitigation.

Earlier guardrails developed prior to the emergence of agentic AI primarily framed content moderation as a discriminative task, wherein a model classifies inputs (and sometimes also outputs) of LLMs as either safe or unsafe, or categorizes them into specific classes of harmful content~\citep{wen2025thinkguard}.
Initial approaches to guardrails relied heavily on rule-based filtering techniques~\citep{welbl-etal-2021-challenges-detoxifying, clarke-etal-2023-rule,DBLP:conf/fat/GomezMPC24} that utilize predefined lexicons or heuristic rules to identify potentially unsafe content.
While these rule-based methods offer transparency and computational efficiency, they inherently lack flexibility and generalization capabilities.
Subsequent studies have transitioned towards model-based guardrails, leveraging supervised fine-tuning on curated safety datasets to enhance content classification in alignment with predefined safety policies. Representative open-source examples in this category include LLaMA Guard~\citep{DBLP:journals/corr/abs-2312-06674, DBLP:journals/corr/abs-2411-17713, DBLP:journals/corr/abs-2411-10414}, NeMo Guardrails~\citep{rebedea2023nemo} and Aegis Guard~\citep{DBLP:journals/corr/abs-2404-05993}.

To advance LLM guardrails toward more agentic capabilities, two key developments are essential: the integration of deliberative reasoning and lifelong learnability. Deliberative reasoning enables models to assess actions more reflectively and align responses with nuanced goals and values. Lifelong learnability allows systems to adapt continuously to new information, evolving norms, and edge cases over time. Together, these capabilities form the foundation for more robust, context-sensitive guardrails that go beyond static filters to support safe and reliable autonomous behavior. In the following, we discuss recent studies on these two key lines of development.

\paragraph{Reasoning Guardrails.}
As a critical but very recent advancement of guardrails, reasoning guardrails assess the intent, context, and possible risks associated with LLMs’ inputs and outputs through structured reasoning, instead of conduct fast-thinking to merely predict threat labels like previous static guardrails.

As one of the very first reasoning guardrails, ThinkGuard~\citep{wen2025thinkguard} draws inspiration from cognitive theories that differentiate fast and slow modes of human thinking~\citep{hagendorff2022thinking,DBLP:journals/corr/abs-2412-09413}.
In this context, fast thinking typically results in superficial or incorrect assessments, making models susceptible to adversarial manipulation, whereas deliberative reasoning mitigates these vulnerabilities by facilitating more robust and contextually informed decisions~\citep{lin2024swiftsage}.
To train guardrails capable of deliberative reasoning, ThinkGuard utilizes mission-focused distillation~\citep{zhou2023universalner} extracting structured reasoning supervision from existing LLMs to generate augmented safety datasets with reasoning critiques.
Deliberative reasoning is integrated into the guardrail through a two-stage conversational fine-tuning procedure: the first stage outputs an initial prediction, followed by a second stage that articulates the underlying reasoning.
This mechanism allows for the efficiency of traditional LLM-based guardrails if reasoning generation is opt out while retaining interpretability when needed.
As a contemporaneous work, GuardReasoner~\citep{liu2025guardreasoner} also devise a similar process to augment safety supervision data with distillation, while its training has two differences: it conducts RL to finetune the guardrail model, but does not learn an optional latent reasoning mode as ThinkGuard.
Based on the methodology exemplified by ThinkGuard and GuardReasoner,
more recent efforts have further extended reasoning guardrails to multimodal~\citep{liu2025guardreasoner-vl} and multilingual~\citep{yang2025mrguard} scenarios.

To summarize, developing the reasoning guardrail models is a prerequisite to realizing Agentic Guardrails because agentic behavior involves multi-step planning and decision-making that cannot be reliably constrained by surface-level filters or static rules. Reasoning Guardrails enable structured oversight of an agent’s internal deliberations, allowing alignment interventions at the level of goals, plans, and justifications.

\paragraph{Agentic Guardrails.}
In contrast to the aforementioned reasoning guardrails, agentic guardrails operate at the level of action execution, overseeing or constraining how an agent interacts with external systems or environments. While reasoning guardrails shape what the agent thinks and tells, agentic guardrails govern what the agent does.

Developing agent guardrails presents a range of challenges spanning both technical and contextual dimensions. A key difficulty lies in enabling \emph{lifelong learnability}, where guardrails must adapt alongside evolving agent behaviors through its continual interaction with users and the environment. Beyond managing threats from user inputs or model-generated content, agents must also detect and mitigate risks arising from external environments, including adversarial states or unsafe system interactions. Ensuring safety in multi-turn actions and dialogus adds further complexity, as harmful behavior may only emerge cumulatively over extended interactions. Finally, generalizability across tasks and environments remains a core obstacle, requiring guardrails that are robust and effective in diverse, dynamic deployment settings without extensive reconfiguration.

A few recent studies have attempted to address some of these challenges, 
inasmuch as some recent studies on lifelong agents \citep{huang2025r2d2,zhang2025cognitive} propose to incorporate continual memories, works such as AGrail~\citep{luo2025agrail}, LlamaFirewall~\citep{chennabasappa2025llamafirewall} GuardAgent~\citep{xiang2025guardagent} leverage this methodology to allow guardrail agents to continually accumulate experiences and new safety policies for agents.
For example, AGrail~\citep{luo2025agrail} proposes a lifelong guardrail framework for LLM agents that dynamically generates and optimizes safety checks during runtime. Central to its approach is a memory-based mechanism that stores and reuses past unsafe trajectories and guardrail refinements, enabling lifelong learnability and continual improvement. It combines adaptive safety-check synthesis with iterative refinement using cooperative LLMs and supports tool-assisted validation, allowing it to address both task-specific (e.g., prompt injection) and environmental threats. AGrail demonstrates strong cross-task generalization and effective integration with various LLM-based agents.
LlamaFirewall~\citep{chennabasappa2025llamafirewall} is a modular, open-source framework designed to secure LLM agents through a combination of real-time defenses. It integrates a fine-tuned BERT-style model for jailbreak and prompt-injection detection, a preliminary auditor leveraging few-shot chain-of-thought prompting for reasoning, and an online system tailored for LLM-generated programs for static analysis of generated code. The system achieves strong security performance with minimal utility loss and is built to be extensible across diverse agent applications.
Meanwhile, GuardAgent~\citep{xiang2025guardagent} dynamically monitors and enforces user‑defined safety or privacy policies by translating guard requests into executable code through a two-step process of task planning and code generation. It leverages a memory‑based in‑context demonstration mechanism that retrieves past examples at each step to enhance its reasoning and support lifelong learnability and adaptability.

\subsubsection{Controllable Generation and Planning}

In addition to leveraging external guardrail components, other research efforts investigate controllable generation to steer LLM agents towards safer generation or planning at inference. In this context, we categorize these efforts into safe decoding approaches, and approaches for agentic access control.

Safe decoding approaches typically incorporate constrained decoding or decoding processes guided by a safety reward to achieve their goals.
For example, SafeDecoding~\citep{xu2024safedecoding} leverages the insight that even under attack the model still assigns non-trivial probability to safe tokens, and dynamically reshapes the token distribution at each decoding step to prioritize harmless outputs.
SafePlanner~\citep{li2025safe} introduces a framework that enhances safety awareness in LLM agents for robot task planning. It incorporates a safety prediction module trained in a simulator, which guides the high-level planner to make safe and executable decisions.
Thought-Aligner~\citep{jiang2025think} is a lightweight, plug-in safety module designed to enhance the behavioral safety of LLM-based agents by dynamically correcting high-risk reasoning steps before action execution. It operates by fine-tuning a contrastive learning model on a dataset of safe and unsafe thought pairs, enabling real-time thought correction with extremely low latency.

Access control remains an underexplored area in current research. Progent~\citep{shi2025progent} introduces the first comprehensive privilege-control mechanism designed specifically for LLM-based agents, enforcing the principle of least privilege during tool invocation.
It centers around a domain-specific policy language that allows developers and users to specify fine-grained constraints on when tools may be invoked and define fallback behaviors for blocked actions.
This policy-driven model enforces the principle of least privilege, ensuring agents only perform tool operations essential to the task at hand.
Its modular architecture enables seamless integration into existing agent systems with minimal code modifications and without altering the agent’s internal logic.
To lower the burden on users, Progent also supports automated policy generation and updates, leveraging LLMs themselves to craft and adapt these privilege policies dynamically in response to evolving user queries.
As such, Progent offers a practical and flexible mechanism for enhancing LLM‑agent security in diverse, real‑world scenarios.
In a related context,
\emph{knowledge access control} is a newly identified and unaddressed problem that concerns dynamic adjustment of LLMs' parametric knowledge based on user privileges~\citep{liu2025sudolm}. 
Traditional safe generation methods typically adopt a uniform policy that blocks sensitive knowledge for all users, potentially reducing utility for credentialed individuals with legitimate access needs. To address this limitation, the SudoLM framework~\citep{liu2025sudolm} introduces a credential-aware mechanism that grants access to privileged knowledge only when a secret SUDO key is provided. This approach partitions the model’s knowledge into public and privileged components and trains it using authorization alignment, enabling differentiated responses based on user credentials. Such a framework offers a promising direction for enhancing safety in LLM-based agents. By conditioning access on user identity, intent, or role, it enables finer-grained, context-sensitive safeguards. Furthermore, its capacity to regulate internal knowledge usage rather than just output filtering allows deeper integration into agent reasoning and planning workflows.

\subsubsection{Emerging Directions in Defense for Agentic Web}

Threat mitigation for agents after their deployment is so far a preliminary area of study. Beyond the current prototypes in existing preliminary studies, and there are quite a few emergent challenges towards truly generalizable and reliable approaches, for which we briefly discuss a selected set of them as follows.

\begin{itemize}
    \item \textbf{Efficiency:} As discussed, the reasoning-based paradigm no doubt strengthens the guardrails in terms of robustness and interpretability. Yet, it inevitably introduces more inference overhead in comparison to previous fast-thinking or discriminative guardrails. While a few current reasoning guardrails such as ThinkGuard~\citep{wen2025thinkguard} have attempted to incorporate latent reasoning to some degrees, how to effectively compress the reasoning process~ and enhance the generation or retrieval of reasoning patterns for a real-time guardrail remains as a non-trivial challenge.
    \item \textbf{Generalizability:} Generalizability is still a core challenge for Web agent safety because these agents operate in open, unpredictable environments with constantly changing interfaces, tools, and tasks. Guardrails that work in training or on benchmarks often fail when agents encounter novel websites or instructions. Unsafe behavior can emerge from unforeseen input combinations or even from new environments. Therefore, ensuring the robustness to diverse and evolving scenarios—not just effective in fixed settings as it is shown in many of the current experimental setups.
    \item \textbf{Certifiable defense:} Certifiable and grounded defense is crucial for web agents because they interact directly with external systems and users, where failures can lead to real-world harm (e.g., sending emails, making purchases, or modifying files). Without grounded verification tied to the actual environment state (e.g., DOM structure, user intent, API constraints), safety mechanisms may rely on incomplete or incorrect assumptions. Future research should develop safety mechanisms that are formally grounded in the web environment, such as DOM structures and API schemas, enabling agents to verify the safety of their actions before execution. It should also explore certifiable control and runtime verification techniques that ensure actions adhere to defined constraints, even under dynamic or adversarial conditions.

\end{itemize}

\subsection{Safety and Security Evaluation}
\label{subsec-safety:agentic-web-evaluation}

Unlike the relatively well-studied areas of traditional web safety \citep{bt2019review, cox2006safety}, LLM safety evaluation \citep{yuan2024r, yuan2025s}, multimodal safety \citep{xu2025mmdt}, and robot learning safety \citep{gu2023safe}, agentic web safety evaluation remains largely underexplored, with only a few preliminary efforts proposed to date. 

One recent work is SafeArena \citep{tur2025safearena}, which introduces a benchmark designed to assess the misuse potential of LLM-based web agents. It evaluates agents on 250 safe and 250 harmful tasks across multiple harm categories, such as misinformation, cybercrime, and social bias, and tests models including ChatGPT \citep{ouyang2022training, achiam2023gpt} and Qwen \citep{bai2023qwen} to measure their compliance with malicious requests. 

Another work is
ST-WebAgentBench~\citep{levy2024st}, which is an open-source benchmark for evaluating the safety and trustworthiness of autonomous web agents in enterprise-style tasks, built on the WebArena environment. It defines six policy dimensions, user consent, preference satisfaction, scope boundaries, strict execution, robustness to distribution shifts, and error recovery, and adopts completion under policy and risk ratio, which measure policy-compliant task success and frequency of violations, respectively. Similarly, AGrail~\citep{luo2025agrail} introduces Safe‑OS, a realistic benchmark designed to evaluate the safety of LLM-powered operating system agents under adversarial conditions. Comprised of three carefully curated attack scenarios, prompt injection, environment sabotage, and system-level exploitation, Safe‑OS simulates real-world threats using Docker-based OS environments alongside benign operation logs. It complements evaluations on existing datasets for operating system agents like Mind2Web‑SC, EICU‑AC (task‑specific risks; \cite{xiang2024guardagent}), AdvWeb~\citep{xu2024advweb}, and EIA (systemic risks; \cite{liao2025eia}) to offer a comprehensive safety assessment

Agent-SafetyBench \citep{zhang2024agent} provides a comprehensive safety evaluation framework with 349 interaction environments and 2,000 test cases across eight safety risk categories. Notably, their evaluation reveals that no current agent achieves safety scores above 60\%, highlighting fundamental deficiencies in agent robustness and risk awareness. Complementing this, GuardAgent \citep{xiang2024guardagent} introduces a dynamic safety guardrail system that achieves 98\% accuracy on safety-critical tasks through knowledge-enabled reasoning, while TrustAgent \citep{hua2024trustagent} implements a three-stage safety strategy encompassing pre-planning knowledge injection, in-planning enhancement, and post-planning inspection.

While benchmarks like SafeArena \citep{tur2025safearena} and ST-WebAgentBench \citep{levy2024st} provide valuable insights into agentic web safety, further investigation is needed, particularly in areas such as multimodal agentic web safety and reasoning safety for agentic web agents.

 \section{Challenges and Open Problems}\label{sec:Challenges}

The realization of the vision of Agentic Web, however, is contingent upon resolving a complex, multi-dimensional set of challenges that span individual agent cognition, multi-agent coordination, human-agent alignment, systemic security, and socio-economic structures.

These challenges are not isolated technical hurdles but are deeply interconnected, forming a web of dependencies that must be addressed holistically. The problem of building the Agentic Web is not merely about improving the capabilities of individual LLM or Agent but about architecting a new, reliable, and trustworthy computational layer atop the existing internet. The systemic nature of these challenges is evident in how they cascade across domains. For instance, the technical need for agents to interact with the external world necessitates the creation of standardized communication protocols, which have been likened to ``HTTP for AI agents''. The existence of this new agent-native architecture, in turn, creates new economic imperatives. The traditional advertising-based business model of the web is ill-suited for an agent-driven economy and is already showing signs of strain. This necessitates new transactional models, but their viability depends directly on solving complex security and trust issues surrounding autonomous payments. Thus, a technical challenge in one area, such as secure tool use, is inextricably linked to a socio-economic challenge in another, such as creating viable business models. A systems-thinking approach is therefore essential, recognizing that a solution for one component may create or exacerbate problems elsewhere. The following table provides a conceptual map of this complex problem space, categorizing the diverse challenges into coherent themes that will be explored throughout this report.

\begin{table}[t]
\centering
\caption{A Taxonomy of Agentic Web Challenges}
\renewcommand{\arraystretch}{1.3}
\begin{tabular}{p{4.2cm}|p{3.5cm}|p{6.2cm}}
\hline
\textbf{Challenge Category} & \textbf{Core Problem} & \textbf{Key Open Questions} \\
\hline
\textbf{Foundational Cognition} & Brittle Reasoning \& Planning & How can agents achieve robust, long-horizon planning under uncertainty? \\
\cline{2-3}
 & Memory \& Context Management & How can we build structured, hierarchical memory systems for agents? \\
\cline{2-3}
 & Reliable Tool Use & How can agents safely and reliably use external tools that may be compromised? \\
\hline
\textbf{Learning Curriculum} & Reward Design \& Alignment & How can we design reward functions that capture nuanced human goals without being gamed? \\
\cline{2-3}
 & Continual Learning \& Forgetting & How can agents acquire new skills over time without catastrophically forgetting old ones? \\
\cline{2-3}
 & Interactive Grounding & How can agents learn from interaction without overfitting to specific environments or prompts? \\
\hline
\textbf{Collaborative Ecosystem} & Inter-Agent Coordination & How can decentralized agents effectively coordinate and resolve conflicts? \\
\cline{2-3}
 & Communication \& Interoperability & What communication standards are needed for a global, open agentic web? \\
\cline{2-3}
 & Decentralized Trust & How can agents establish and maintain trust in a decentralized, potentially adversarial ecosystem? \\
\hline
\textbf{Human-Agent Alignment} & Goal Ambiguity \& Disambiguation & How can an agent reliably infer a user's true intent from ambiguous instructions? \\
\cline{2-3}
 & Preference Elicitation & How can agents help users discover and articulate their own complex preferences? \\
\cline{2-3}
 & Oversight \& Control (HITL) & What are the most effective architectures for human-in-the-loop oversight? \\
\hline
\textbf{Systemic Risk \& Robustness} & Security \& Attack Surfaces & How can we defend agents against novel threats like tool-initiated attacks? \\
\cline{2-3}
 & Error Recovery \& Resilience & How can we engineer agentic systems to be resilient to inevitable failures? \\
\cline{2-3}
 & Autonomous Payments & What technical and regulatory frameworks are needed for secure agent-based payments? \\
\hline
\textbf{Socio-Economic Impact} & New Business Models & What viable business models will replace the advertising-based economy? \\
\cline{2-3}
 & Economic Disruption \& Inequality & How can the economic benefits of agentic AI be distributed equitably? \\
\hline
\end{tabular}
\end{table}

\subsection{Foundational Challenges in Single-Agent Cognition and Autonomy}
Before complex multi-agent and human-agent systems can be reliably constructed, the core cognitive architecture of an individual agent must be made robust. This section deconstructs the fundamental technical hurdles that currently undermine the reliability, planning capabilities, and autonomous functioning of a single agent. These are the first-order problems that form the bedrock of the Agentic Web.

\paragraph{The Fragility of Reasoning and Planning}
The capacity for multi-step reasoning is a cornerstone of agentic systems, enabling them to decompose complex problems, evaluate alternative solutions, and make informed decisions. This process is often operationalized through a continuous cycle of planning, action, observation, and reflection, with frameworks like Chain-of-Thought serving as the primary mechanism for articulating these reasoning steps in natural language. However, this capability is deceptively brittle.

\paragraph{The Memory-Context Dilemma}
Memory is an essential architectural component for agentic systems. Since LLMs are fundamentally stateless, they require external mechanisms to retain conversation history, contextual information, and learned knowledge. Agentic architectures typically employ both short-term memory to maintain coherence within a single task and long-term memory  to carry knowledge across tasks. However, the management of this memory, especially in the face of finite context windows and complex, long-horizon tasks, remains a primary bottleneck.

\paragraph{The Tool-Use Paradox}

The ability to use external tools, such as APIs, databases, calculators, and web search, is what transforms a passive LLM into an active agent capable of interacting with and affecting the real world. This is the primary mechanism for grounding an agent's reasoning in actionable reality. However, this capability introduces a fundamental paradox: the every tools that grant an agent real-world agency simultaneously represent its greatest vulnerability.

This creates a ``Tool-Use Paradox'': to be effective, an agent must trust its tools to provide accurate information and execute actions correctly; to be secure, it must assume any tool could be compromised at any time. The resolution to this paradox lies in designing agents with an inherent ``tool skepticism.'' This requires moving to a zero-trust agent architecture where all external inputs, whether from a user or a tool, are validated against a security policy. Previously, agent security focused primarily on validating the user's prompt. The existence of tool-initiated threats means the agent must also validate the tool's response, creating a feedback loop of potential infection where a malicious tool output could cause the agent to take another malicious action, leading to a cascading failure. A secure agent cannot be a naive ``tool-caller''; it must possess a security kernel or policy engine that scrutinizes all information crossing the boundary between its internal state and the external world. The open research question is how to build this skepticism without crippling the agent's ability to act decisively based on tool outputs.

\subsection{The Learning Conundrum: From Static Models to Dynamic Learners}

While foundational models provide a powerful starting point, true agency requires the ability to learn from experience, adapt to new environments, and continuously improve performance. This section explores the profound challenges associated with transforming static, pre-trained models into dynamic, lifelong learners, focusing on the bottlenecks in RL, the threat of catastrophic forgetting, and the difficulties of grounding knowledge through interaction.

\paragraph{Reward Design Bottleneck}
RL is the primary paradigm for training agents to make optimal sequential decisions by interacting with an environment. It offers a path to move beyond the limitations of static, pattern-replicating LLMs, enabling them to handle ambiguity, maintain context in long conversations, and balance competing objectives. However, the effectiveness of RL is critically dependent on the design of its reward function, which has become a major research bottleneck.

\paragraph{The Specter of Catastrophic Forgetting in Continual Learning}
For agents to be truly autonomous and useful over long periods, they must be able to engage in continual, or lifelong, learning: acquiring new knowledge and skills without overwriting or degrading previously learned capabilities. The primary obstacle to achieving this is ``catastrophic forgetting,'' a well-known phenomenon in neural networks where training on a new task causes a model to abruptly lose proficiency on previously learned tasks.

\paragraph{Interactive Task Learning and Grounding}
Ultimately, agents learn to perform complex tasks by interacting directly with their environment, whether it is a digital application or the physical world. This interactive learning process, often guided by RL, is what allows agents to ground their abstract knowledge in concrete actions and feedback. However, this process is fraught with challenges related to the trade-off between specialization and generalization.

\subsection{The Ecosystem Challenge: Coordination and Trust in Multi-Agent Systems}
Expanding the analysis from the single agent to the collective reveals the profound complexities that arise when multiple autonomous agents must interact, collaborate, and compete. The success of the Agentic Web as a whole depends on the ability to orchestrate these multi-agent systems effectively, a challenge that encompasses architectural design, communication standards, and the establishment of trust in decentralized environments.

\paragraph{Architectural Trade-offs: Hierarchical, Equi-level, and Nested Structures}
Multi-agent systems can be organized into several distinct architectures, each with unique properties and challenges. The primary structures identified in current research are equi-level (peer-to-peer), hierarchical (leader-follower), and nested (hybrid systems of systems).

\paragraph{The Babel of Agents: The Imperative for Standardized Communication}
For a global Agentic Web to function, agents developed by different organizations on different platforms must be able to communicate and interoperate. Without common standards, the ecosystem would devolve into a collection of isolated, proprietary ``walled gardens,'' akin to the pre-HTTP internet, stifling innovation and collaboration.

The primary challenge is to develop and adopt standardized communication protocols that are expressive enough to support complex agent interactions yet simple and open enough to foster widespread adoption. This involves standardizing both the syntax (the format of messages) and the semantics (the meaning of communicative acts, often based on speech act theory). Emerging standards like IBM's ACP and Google's A2A for agent-to-agent communication and Anthropic's MCP for agent-to-tool communication are designed to work in tandem to provide this foundational layer. Major industry players are championing these open protocols, arguing that achieving ubiquity is more important than perfecting minor semantic differences, in order to create a truly open agentic web.

\paragraph{Establishing Trust and Reputation in Decentralized Ecologies}
In a decentralized system of autonomous, potentially self-interested agents, trust is the essential lubricant that enables collaboration and reduces uncertainty. To make informed decisions about whom to interact with and delegate tasks to, agents need a mechanism to assess the reliability and competence of their peers.

\subsection{The Human-Agent Interface: Ensuring Goal Alignment and Control}
This section focuses on the critical interface between human users and autonomous agents. The central challenge is ensuring that an agent's actions faithfully reflect the user's true, often nuanced and evolving, intent. This requires solving deep problems of goal ambiguity, preference discovery, and the design of effective oversight mechanisms to maintain human control.

\paragraph{The Ambiguity Problem: From User Intent to Actionable Goals}
The first step in any agentic workflow is understanding the user's goal. However, human language is often imprecise, and user requests can be complex, ambiguous, or underspecified. An agent must be able to disambiguate this input and translate it into a concrete, actionable plan. This often involves a process of active disambiguation, where the agent poses clarifying questions to maximize information gain and narrow the space of possible interpretations~\citep{jiang2024llms}.

\paragraph{Eliciting Nuanced Preferences}
A significant challenge in achieving goal alignment is that users themselves often do not have fully formed, stable preferences. A substantial body of psychology research has demonstrated that preferences are often constructed ``on the fly'' at the time of decision-making, influenced by the immediate context and the options presented~\citep{lawless2024want}.

An agent, therefore, cannot simply ask a user for their complete utility function. Instead, it must engage in an iterative, collaborative process of preference elicitation, helping the user to discover, construct, and refine their own preferences over time. This requires the agent to move beyond a passive question-answer model to become an active participant in the user's reasoning process. The cost and accuracy of a user's responses to preference queries are highly dependent on context; users respond more easily and accurately to queries about situations they are currently or have recently experienced. This makes naturalistic, chat-based elicitation a promising approach.

\paragraph{Human-in-the-Loop (HITL): Designing Effective Oversight Architectures}
Given the current limitations in agent reliability and alignment, incorporating a ``human in the loop'' (HITL) is a critical mechanism for ensuring safety, accountability, and control, especially for high-stakes or irreversible actions. HITL represents a collaborative paradigm where humans and AI work together to optimize processes.

\subsection{Systemic Risks: Ensuring Safety, Security, and Robustness}

As agents become more autonomous and capable of taking real-world actions, the risks they pose escalate dramatically. This section addresses the critical challenges of ensuring that agentic systems are secure from attack, robust to failure, and safe to deploy in high-stakes environments such as finance and critical infrastructure.

\paragraph{Safety and Security Challenges}
We explore the safety and security challenges of the agentic web, where autonomous agents operate across open, dynamic environments. It categorizes threats across cognitive, communication, and economic layers and highlights cascading risks that amplify system vulnerabilities. To address these issues, both human-involved and automated red teaming are discussed, with LLM-driven approaches offering scalability but requiring robust oversight. Defense strategies include advanced guardrails with reasoning and lifelong learning, such as AGrail~\citep{luo2025agrail} and GuardAgent \citep{xiang2025guardagent}. While benchmarks like SafeArena \citep{tur2025safearena} and ST-WebAgentBench~\citep{levy2024st} have made progress in evaluation, significant gaps remain, particularly in multimodal and reasoning safety, calling for further research in scalable, adaptive safety solutions.

\paragraph{Long-Horizon Planning and Error Recovery}
Real-world tasks are rarely simple, single-step operations. They often involve long-horizon plans with numerous sequential and parallel actions. In complex and partially observable environments, failures are not a possibility but a certainty.

The dual challenges are (1) creating and maintaining a coherent plan over a long sequence of steps without the plan degrading or becoming irrelevant, and (2) building in robust mechanisms for detecting, diagnosing, and recovering from the inevitable errors, exceptions, and action failures that will occur. Simple, sequential agentic chains that work well in prototypes often break under the variability and load of real-world use because they lack graceful failure modes and recovery paths. A key to robustness is moving beyond open-loop ``plan-and-execute'' paradigms to closed-loop systems that can self-correct based on feedback from their actions~\citep{nayak2024long}.

\paragraph{The Challenge of Autonomous Payments: Security and Regulation}
Empowering agents with the ability to spend money is a critical enabler for a transactional Agentic Web, but it also represents one of the highest-risk applications, facing immense technical, regulatory, and social hurdles.

\subsection{Socio-Economic Implications}
The successful deployment of the Agentic Web would not be a mere technological evolution but a profound socio-economic transformation, reshaping business models, labor markets, and the very structure of the digital economy. This section explores the challenges and open questions related to the economic viability and societal impact of this new paradigm.

\paragraph{Beyond Advertising: Viable Business Models for an Agentic Economy}
The current economic foundation of the consumer web, advertising, is ill-suited for and actively threatened by the rise of AI agents. The Agentic Web necessitates a shift towards new, more direct forms of value exchange.

The ad-supported model, which monetizes human attention, is breaking down as agents become the primary interface for information retrieval, disintermediating and reducing traffic to content websites. The challenge is to develop and scale new business models that are native to an economy of automated actions, not human eyeballs. This likely involves a move towards transactional, subscription, and value-based pricing models. Emerging models already position agents as first-class business entities that can be customized and deployed by organizations, enabling new revenue streams. These include: \textit{Intelligence-as-a-Service}, where the outputs of AI-powered research are sold on demand; \textit{Zero Marginal Cost Services}, where the incremental cost of serving another customer is near zero; and \textit{Value-Based Pricing}, where customers pay for outcomes rather than time invested.

Furthermore, the integration of blockchain technology presents promising opportunities for the Agentic Web's economic foundation. Blockchain-enabled platforms can facilitate decentralized agent interactions, autonomous transactions, and trustless value exchange between AI agents. Projects like ChainOpera \citep{chainopera2024} demonstrate the practical convergence of Web3 and agentic AI, while emerging protocols such as Protocol AI \citep{protocolai} support agent-blockchain integration for decentralized tokenization of alternative assets \citep{borjigin2025ai} and autonomous operations in decentralized finance \citep{ante2024autonomous}. This convergence could enable new forms of autonomous economic activity where agents can independently engage in value creation and exchange without traditional intermediaries.

\paragraph{Economic Disruption: Labor Markets, Productivity, and Inequality}
The widespread adoption of AI agents promises massive gains in productivity and economic growth but also portends significant disruption to the labor market and carries the risk of exacerbating economic inequality. Research suggests that generative AI alone could add trillions of dollars annually to the global economy, but it could also automate a significant fraction of current work activities, affecting hundreds of millions of jobs worldwide.

\section{Conclusion}\label{sec:conclusion}

The internet is undergoing a fundamental paradigm shift, evolving from a passive repository of information to a dynamic environment of action. This transition is powered by the emergence of the Agentic Web, a landscape populated by autonomous systems capable of perceiving their environment, reasoning through complex problems, and executing tasks to achieve specified goals. This marks a significant leap from generative AI, which excels at responding to human prompts, to agentic AI, which is characterized by proactive, independent decision-making and execution.

\bibliographystyle{plainnat}
\bibliography{references, ref_shangding, ref_muhao, ref_ming, ref_haoran}  






\end{document}